\documentclass[10pt,twocolumn,letterpaper]{article}

\usepackage{wacv}
\usepackage{times}
\usepackage{epsfig}
\usepackage{graphicx}
\usepackage{amsmath}
\usepackage{amssymb}
\usepackage{booktabs}
\usepackage{datetime2}
\usepackage{bm}
\usepackage{color,soul}          % color
\usepackage{microtype}      % microtypography
\usepackage{dcolumn}        % Align table columns on decimal point
\usepackage{url}            % simple URL typesetting
\usepackage{booktabs}       % professional-quality tables
\usepackage{amsfonts}       % blackboard math symbols
\usepackage{nicefrac}       % compact symbols for 1/2, etc.
\usepackage{enumitem}
\usepackage{multirow}
\usepackage{multicol}
\usepackage{float}
\usepackage{algorithm}
\usepackage[noend]{algpseudocode}
\usepackage{amsthm}
\usepackage{subfiles}
\usepackage{adjustbox}
\usepackage{subcaption}
\usepackage{makecell}
\usepackage[normalem]{ulem}
\usepackage{caption}
\usepackage{nopageno}
\usepackage{rotating}
\usepackage[flushmargin,para]{footmisc}
\usepackage[table]{xcolor}
\usepackage{diagbox}
\def\wacvPaperID{1531} % Enter the WACV Paper ID here

\wacvalgorithmstrack   % Uncomment this line if you are submitting to the Algorithms Track.
\wacvfinalcopy % *** Uncomment this line for the final submission

\ifwacvfinal
\usepackage[breaklinks=true,bookmarks=false]{hyperref}
\else
\usepackage[pagebackref=true,breaklinks=true,colorlinks,bookmarks=false]{hyperref}
\fi
\usepackage[capitalize]{cleveref}
\crefname{section}{Sec.}{Secs.}
\Crefname{section}{Section}{Sections}
\Crefname{table}{Table}{Tables}
\crefname{table}{Tab.}{Tabs.}
\newcommand{\fig}[1]{Fig.~\ref{#1}}

\newcommand{\realnum}{\mathbb{R}}
\newcommand{\loss}{\mathcal{L}}

\newcommand{\image}{\mathbf{I}}
\newcommand{\imageheight}{H}
\newcommand{\imagewidth}{W}

\newcommand{\featuremapheight}{H'}
\newcommand{\featuremapwidth}{W'}
\newcommand{\featuremapchannels}{D}

\newcommand{\modelindex}{i}

\newcommand{\numclasses}{C}
\newcommand{\nummodels}{N}

\newcommand{\probabilitymap}{\mathbf{P}}

\newcommand{\groundtruthsegmap}{\mathbf{Y}}

\newcommand{\ensemble}{\mathcal{B}}
\newcommand{\ensemblemember}{B}
\newcommand{\ensemblecombination}{\Phi}

\newcommand{\sequentialensemble}{\mathcal{G}}
\newcommand{\deepnetwork}{G}

\newcommand{\convoperation}{\mathbf{F}}
\newcommand{\intermediatefeaturemap}{\mathbf{X}}
\newcommand{\sharedoutput}{\mathbf{e}}
\newcommand{\modulationscaleoutput}{\boldsymbol{\sigma}}
\newcommand{\modulationtranslationoutput}{\boldsymbol{\beta}}

\definecolor{lightgray}{gray}{0.97}
\definecolor{lightblue}{rgb}{0.93,0.95,1.0}
\definecolor{gold}{rgb}{0.85,.66,0}

\begin{document}

%%%%%%%%% TITLE
\title{Sequential Ensembling for Semantic Segmentation}

\author{{Rawal Khirodkar}\textsuperscript{{1\thanks{Work done during an internship at Amazon}}} \hspace{0.4cm} {Brandon Smith}\textsuperscript{{2}} \hspace{0.4cm} {Siddhartha Chandra}\textsuperscript{{2}} \hspace{0.4cm} {Amit Agrawal}\textsuperscript{3} \hspace{0.4cm} {Antonio Criminisi}\textsuperscript{2\thanks{Now at Microsoft}}\\
\textsuperscript{1}Carnegie Mellon University \hspace{1cm} \textsuperscript{2}Amazon Lab 126\hspace{1cm} \textsuperscript{3}Amazon Fashion\\
} 
\maketitle
\thispagestyle{empty}

%%%%%%%%% ABSTRACT
\begin{abstract}
Ensemble approaches for deep-learning-based semantic segmentation remain insufficiently explored despite the proliferation of competitive benchmarks and downstream applications. In this work, we explore and benchmark the popular ensembling approach of combining predictions of multiple, independently-trained, state-of-the-art models at test time on popular datasets. Furthermore, we propose a novel method inspired by boosting to sequentially ensemble networks that significantly outperforms the na\"{i}ve ensemble baseline. Our approach trains a cascade of models conditioned on class probabilities predicted by the previous model as an additional input. A key benefit of this approach is that it allows for dynamic computation offloading, which is useful for deploying models on mobile devices. Our proposed novel ADaptive modulatiON (ADON) block allows spatial feature modulation at various layers using previous stage probabilities. Our approach does not require any sophisticated sample selection strategies during training, and works with multiple neural architectures. We significantly improve over the na\"{i}ve ensemble baseline on challenging datasets such as Cityscapes, ADE-20K, COCO-Stuff and PASCAL-Context and set a new state-of-the-art.
\end{abstract}

%%%------------------------------------------------------------------------
%%%%%%%%% INTRODUCTION
\vspace*{-0.25in}
\section{Introduction}
\label{sec:intro}
% % -----------------------begin intro figure--------------------
\setlength{\belowcaptionskip}{-10pt}
 \begin{figure}[t]
 \begin{center}
  \includegraphics[width=1\linewidth, height=0.9\linewidth]{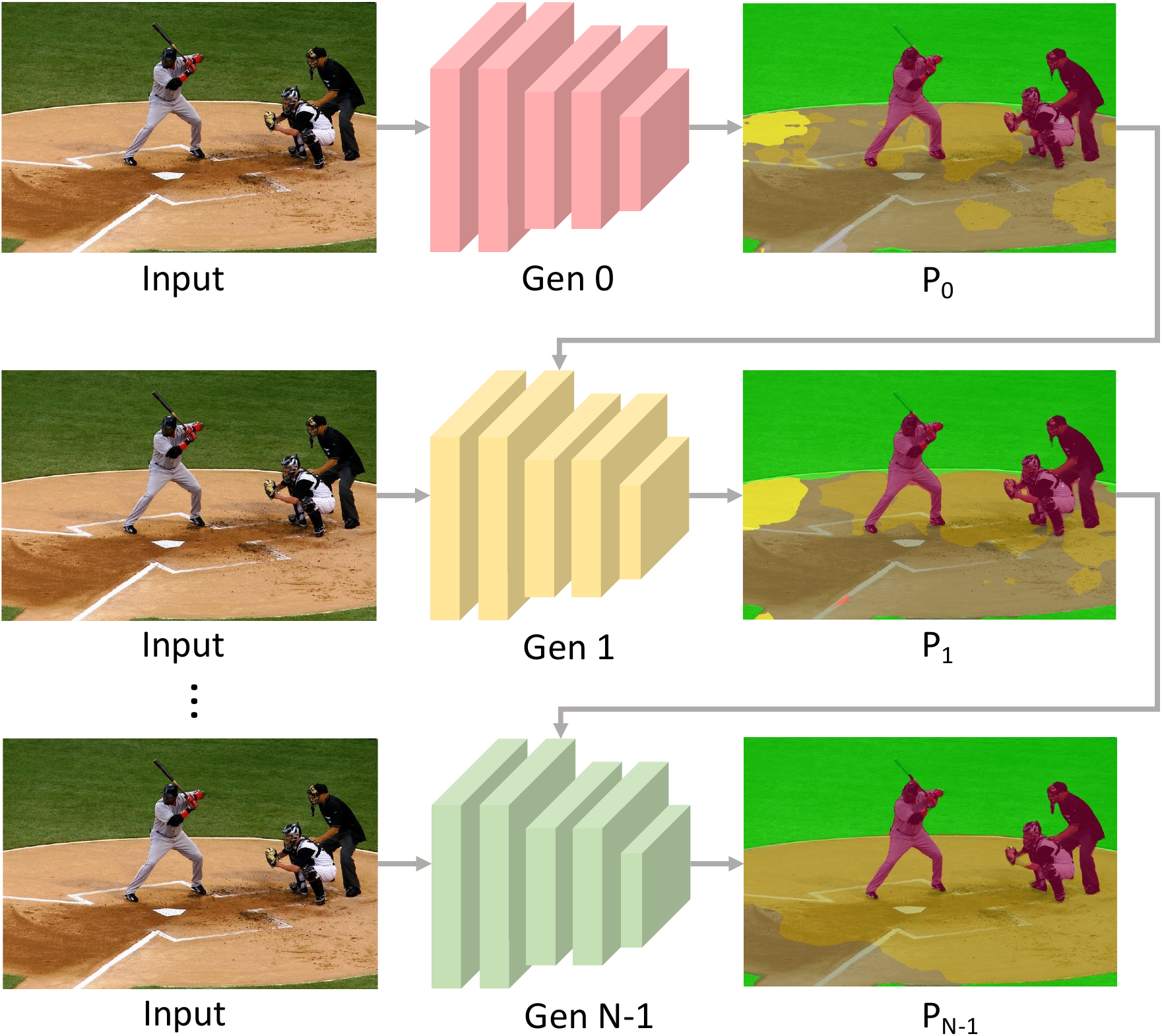}
 \end{center}
  \vspace*{-0.2in}
 \caption{\textbf{Sequential Ensembling:} We sequentially train $N$ generations as a cascade. Each generation predicts segmentation probabilities from the input image conditioned on the segmentation probabilities from the previous generation. 
}
\label{fig:introduction}
\vspace*{-0.1in}
 \end{figure}
 %  % -----------------------end intro figure--------------------

Semantic segmentation, \ie, pixel classification, is a fundamental task in computer vision with applications in a wide variety of domains, such as autonomous driving~\cite{siam2017deep}, robotic navigation~\cite{lin2017learning}, medical analysis~\cite{ronneberger2015u}, and scene understanding~\cite{zhou2019semantic}. Over the years, deep learning based semantic segmentation methods have achieved remarkable performance ~\cite{yu2017dilated,lin2017refinenet,yu2015multi,badrinarayanan2017segnet,long2015fully,yuan2020object,wu2019wider}. Most of these approaches focus on training a \textit{single} network with novel architectures and loss functions to improve performance, overlooking the extensive classical literature on ensembles~\cite{breiman1996bagging, schapire1995decision, viola2001rapid}. Ensembles have been shown to improve accuracy~\cite{breiman2001ml}, uncertainty estimation, and out-of-distribution robustness~\cite{fort_arxiv_2019}. We argue that deep learning approaches can benefit greatly from the pioneering work done in ensemble learning. In this paper, we methodically explore deep ensembles for semantic segmentation with the goal of further improving performance of state-of-the-art semantic segmentation models.

\begin{figure}
\centering
\includegraphics[width=0.9\linewidth]{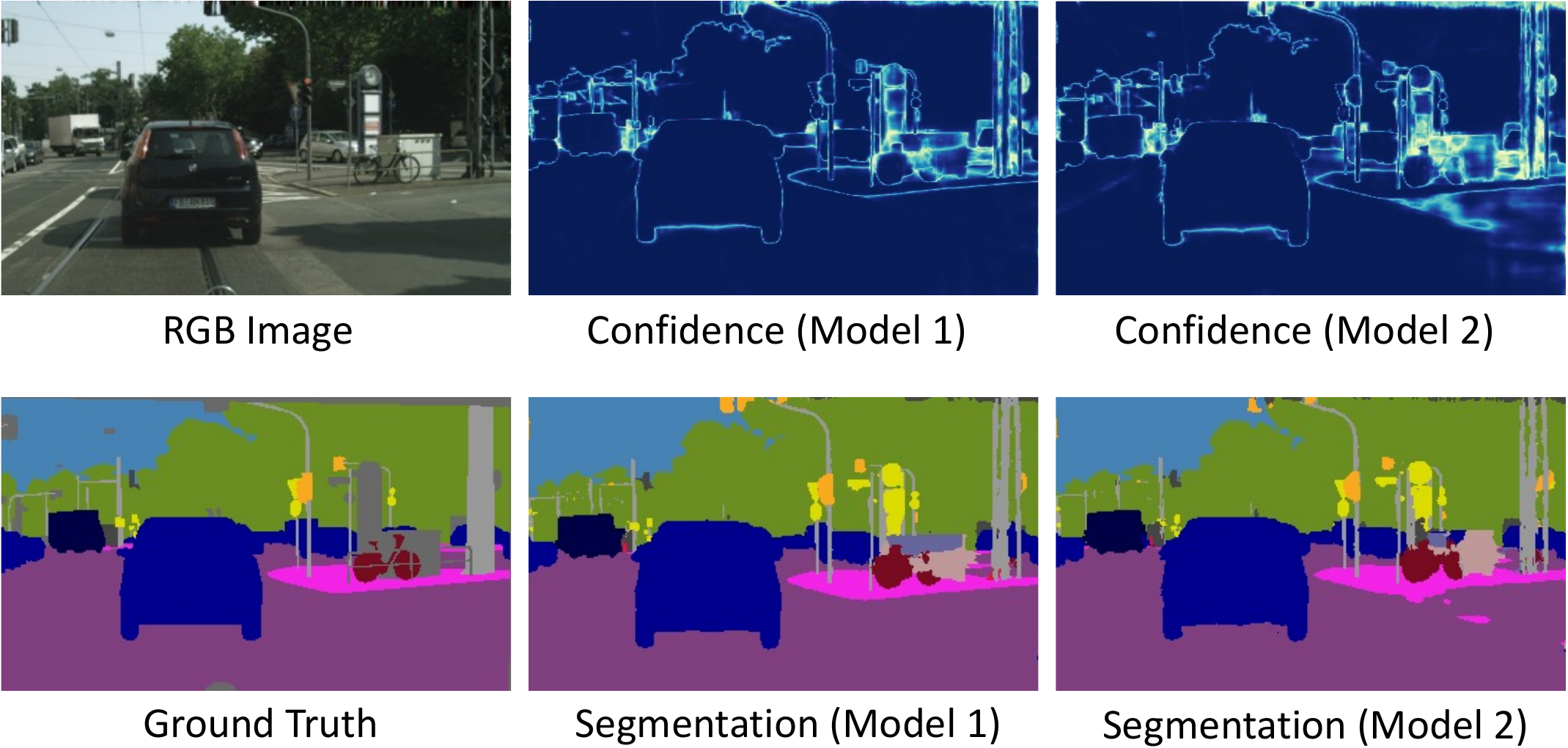}
\caption{Heatmap visualization of per-pixel confidence (max class probability) of independently trained models (\textcolor{blue}{blue} is high confidence). We observe that confidence on a pixel is not necessarily correlated with accuracy of segmentation model. Empirically, our method performs better than confidence weighted ensembling rule. }
\label{fig:pixel_confidence}
\end{figure}

A na\"{i}ve deep ensembling approach is to train multiple networks and fuse their outputs (\eg, average their predicted probabilities). We benchmark this ensembling approach for state-of-the-art deep semantic segmentation models across multiple datasets. Similar to~\cite{krizhevsky2012imagenet, chen2017deeplab, wang2020surprising}, the baseline ensembling rule averages predicted probabilities using equal weight for all the models within the ensemble. We refer to this as \textit{simple ensembling} (SIM-ENS). Consistent with~\cite{chen2017deeplab, wang2020surprising}, and as shown in \cref{table:intro_performance}, we observe diminishing returns as the number of models $\nummodels$ in the ensemble increases. For dense per-pixel predictions, simple ensembles are not effective as distinct models perform well on different regions of the image. A natural variation is to try other ensembling rules such as max-voting and weighted-averaging using model confidence estimates~\cite{devries2018learning}. However, these strategies result in moderate gains similar to simple ensembles as shown in~\cref{table:intro_performance}, owing to challenges in obtaining reliable per pixel confidence weights (see Fig. \ref{fig:pixel_confidence}).

\begin{figure}
\centering
\includegraphics[width=1\linewidth,]{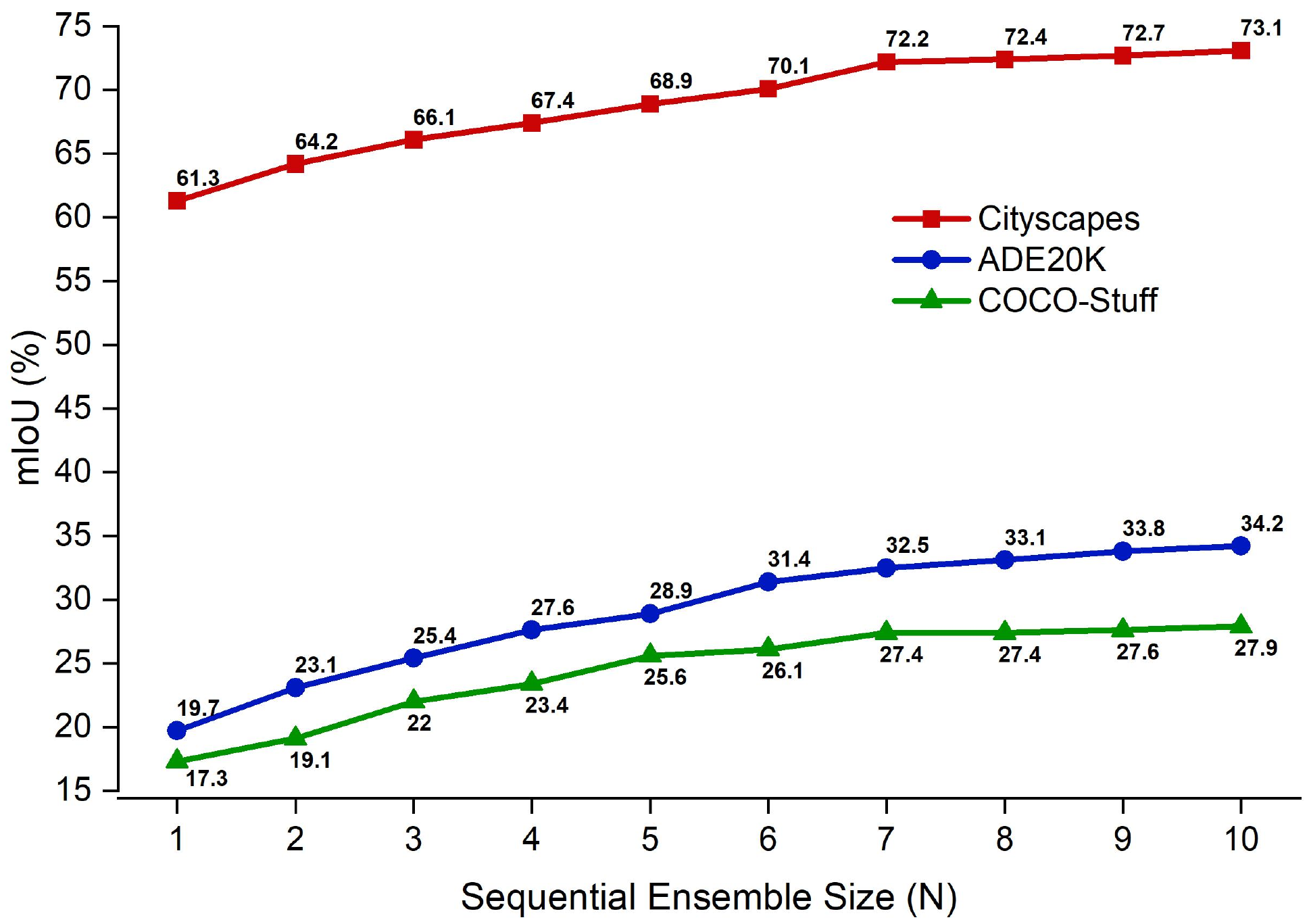}
\caption{Performance of Sequential Ensembling using FCN~\cite{long2015fully} MobileNetv2-D8~\cite{sandler2018mobilenetv2} backbone on the \texttt{val} set of various datasets using single-scale testing. We see increase in performance as more stages are added, allowing run-time trade-off between accuracy and speed for edge devices.}
\label{fig:introduction_plot}
\end{figure}

In this work, we propose \textit{sequential ensembling} as an alternative to simple ensembling. Sequential ensembling (SEQ-ENS) is a data-driven approach to \textit{learn} an optimal ensemble compared to ad-hoc ensembling approaches (\eg, averaging or voting). Inspired by boosting~\cite{schapire1990strength} and cascaded refinement~\cite{moon2019posefix, chen2017photographic}, we train $N$ generations of deep models sequentially, as shown in \fig{fig:introduction}. Generation $G_i$ uses input image $\mathbf{I}$ and the predicted probability map $\probabilitymap_{i-1}$ from generation $G_{i-1}$ to predict $\probabilitymap_i$. 
Our approach only requires probability maps from the previous generation and does not require any sophisticated sample selection strategy during training. 
To condition on probability maps, we adopt the simple yet effective mechanism of feature modulation~\cite{perez2018film, dumoulin2016learned, huang2017arbitrary}. We propose a novel ADaptive modulatiON (ADON) block which injects information from $\probabilitymap_{i-1}$ at multiple depths within the network. Dense prediction is aided by spatial context and the ADON block allows spatial modulation of intermediate features of $G_i$. Each subsequent generation learns to incorporate knowledge from the previous generation using class-specific spatial context.
Unlike simple ensembles, each model can dedicate network capacity to correct mistakes made by prior generations. Our approach allows for increase in accuracy as the ensemble size is increased, as shown for multiple datasets in \fig{fig:introduction_plot}.

% % -----------------------begin intro table--------------------
\begin{table}[b]
% \captionsetup{font=small}
\small

\begin{center}

\resizebox{3.3in}{!}{
    \renewcommand{\arraystretch}{1.4}
    \rowcolors{1}{}{lightgray}
    \begin{tabular}{@{}l|l|l|l @{}}
    \Xhline{3\arrayrulewidth}
         \small{\textbf{Method}} &\small{\textbf{Cityscapes}}   & \small{\textbf{ADE20K}} & \small{\textbf{COCO-Stuff}}\\
    \hline
    \small{Single Model} \scriptsize{($N=1$)} & 76.2  &  33.1 & 32.4 \\
    \small{SIM-ENS} \scriptsize{($N=2$)} &  77.1 (+0.9) & 34.5 (+1.4) & 33.5 (+1.1)\\
    \small{SIM-ENS} \scriptsize{($N=5$)} &  77.5 (+1.3) & 34.9 (+1.8) & 33.8 (+1.4) \\
    \hline
    \small{SIM-ENS} \scriptsize{($N=15$)} &  77.8 (+1.6) & 35.8 (+2.7) & 34.9 (+2.5) \\
    \small{Voting Ens.} \scriptsize{($N=15$)} &  76.9 (+0.7) & 35.3 (+2.2) & 35.1 (+2.7) \\
    \small{W-Avg. Ens.} \scriptsize{($N=15$)} &  78.0 (+1.8) & 35.0 (+1.9) & 34.6 (+2.2) \\
    \hline
    \small{SEQ-ENS} \scriptsize{($N=2$)} & \textbf{78.9 (+2.7)} & \textbf{37.6 (+4.5)} & \textbf{36.3 (+3.9)} \\
    \Xhline{3\arrayrulewidth}
    \end{tabular}
}
\caption{Sequential ensembling (SEQ-ENS) offers significantly higher improvement (mIOU) compared to traditional ensembling approaches based on voting and simple/weighted averaging. Results are reported on the \texttt{val} sets using HRNetW18s-v2~\cite{sun2019high} backbone with single-scale testing.}
\label{table:intro_performance}
\end{center}
\end{table}
 %  % -----------------------end intro table--------------------

 % -----------------------begin intro plot figure--------------------
 % -----------------------end intro plot figure--------------------

\textit{Sequential ensembles} are constructed by training a new generation with parameters of all previous generations $\{G_{i-1}, G_{i-2}, \dots G_0\}$ frozen. This affords several important benefits. For example, the sequential chain can be interrupted at any depth, and each stage can still output pixel-wise probabilities. This is a valuable property that allows for dynamic computation offloading~\cite{mao2016dynamic} where the ensemble size can be adjusted on the fly to meet resource limitations without any retraining. This strategy is useful for deploying models on diverse mobile platforms to best suit changing on-device constraints, \eg, diminishing battery life, applications that require accuracy vs.~runtime tradeoffs (fast previews vs.~slow high-quality results), responsiveness requirements in the context of changing mission-critical workloads. 

% To propagate information from prior generations to the current generation, we adopt the simple yet effective mechanism of feature modulation~\cite{perez2018film, dumoulin2016learned, huang2017arbitrary}. We propose a novel ADaptive modulatiON (ADON) block which processes the probability maps of the previous generation. The ADON blocks inject cues from $\probabilitymap_{i-1}$ at multiple depths in the backbone of $G_i$. The spatial context is necessary for dense prediction, and the ADON block allows spatial modulation of intermediate features of $G_i$ without changing the backbone architecture. We show that unlike \textit{early fusion} (\eg, channel-wise concatenation) of $\probabilitymap_{i-1}$ with $\mathbf{I}$, ADON can effectively utilize the information in $P_{i-1}$ as well as the input image for training $G_i$. Also, in comparison to \textit{late fusion} of $\probabilitymap_{i-1}$ with $\mathbf{I}$, ADON preserves low-level image details in the segmentation map, \eg, predicting poles, fences, traffic lights. 

Our proposed approach achieves state-of-the-art results on Cityscapes~\cite{cordts2016cityscapes}, ADE20K~\cite{zhou2017scene}, PASCAL-Context~\cite{mottaghi_cvpr14} and COCO-Stuff~\cite{caesar2018cvpr}. For backbones with lower computational complexity, sequential ensembles $(N=2)$ outperforms simple ensembles $(N=15)$ across multiple datasets, as shown in \cref{table:intro_performance}. In Tab.~\ref{table:adon_placement}, we additionally show that a  $2$ model sequential ensemble with single-scale inference outperforms a single model with multi-scale test-time augmentation using $12$ inferences ($6$x lower inferences). Furthermore, we show that sequential models can (a) generalize across different backbone architectures (\cref{table:sim_ens_compare}), (b) exhibit \textit{self-improvement} via the ability to refine their own segmentation maps iteratively yielding non-trivial gains (\cref{table:self_improvement}) and (c) can be used to create general neural network graphs leveraging diversity gains from ensembling (\cref{table:ensemble_of_chains}). 

In summary: 
\begin{itemize}
\itemsep0em 
  \item We explore ensembling approaches for deep neural networks to improve state-of-the-art segmentation networks and provide an ensembling benchmark of various models across multiple semantic segmentation datasets.
  \item Our proposed sequential ensembling outperforms na\"{i}ve ensembling methods and advances the state-of-the-art on challenging benchmarks.
  \item Our proposed feature modulation ADON blocks efficiently incorporates information from previous generations in the sequential ensemble chains.
\end{itemize}

% -----------------------begin intro confidence figure--------------------
 % -----------------------end intro confidence figure--------------------

%%%------------------------------------------------------------------------
%%%%%%%%% RELATED WORK
\section{Related Work}
\label{sec:related_work}
\textbf{Semantic-Segmentation:} Semantic segmentation is an extension of the classification problem from image level to pixel level. Over the past few years, advances in deep learning for image classification~\cite{he2016deep, krizhevsky2012imagenet, simonyan2014very, dosovitskiy2020image, wang2020deep, tan2019efficientnet, shotton2009textonboost} led to significant improvements for semantic segmentation~\cite{sun2019high, long2015fully, liu2021swin, ronneberger2015u, yuan2020object}. These efforts benefited from the introduction of popular datasets such as Cityscapes~\cite{cordts2016cityscapes}, ADE20K~\cite{zhou2017scene}, PASCAL-Context~\cite{mottaghi_cvpr14}, PASCAL-VOC~\cite{everingham2015pascal}, COCO-Stuff~\cite{caesar2018cvpr}, \etc. In this work, we benchmark ensembling for segmentation.

\textbf{Ensembles for Segmentation:} Ensemble learning has been well-studied in machine learning; seminal works include bagging~\cite{breiman1996bagging}, boosting~\cite{schapire1990strength}, and AdaBoost~\cite{freund1997decision}. Ensembles of deep models have been used to boost performance on many tasks such as image classification~\cite{szegedy2015going, huang2017snapshot}, machine translation~\cite{wen2020batchensemble} and uncertainty estimation~\cite{lakshminarayanan_nips_2017, fort_arxiv_2019, wenzel2020hyperparameter, zhao2005survey}. However, ensembling approaches have not been fully explored for semantic segmentation owing to computational challenges in dense pixel prediction. Our work benchmarks ensembles of state-of-the-art models across various segmentation datasets~\cite{cordts2016cityscapes,zhou2017scene,mottaghi_cvpr14,khirodkar2019domain, caesar2018cvpr,everingham2015pascal} and model families~\cite{sun2019high,long2015fully,xie2021segformer}. We hope this will be helpful in inspiring and evaluating new ideas in field. Further, inspired by boosting paradigm~\cite{schapire1990strength}, our proposed sequential ensemble methodology achieves new state-of-the-art segmentation results.

\textbf{Segmentation Refinement:} Many refinement methods~\cite{gidaris2017detect, dias2020probabilistic, kuo2019shapemask, lin2017refinenet, islam2017label,  li2016iterative, yuan2020segfix, iwase2021repose} have been proposed. These refinement approaches depend on the segmentation model~\cite{li2016iterative}, super-pixels~\cite{dias2020probabilistic}, multi-scale input~\cite{lin2017refinenet}, data-generation~\cite{khirodkar2018adversarial}, or object boundary information~\cite{yuan2020segfix}. In contrast to~\cite{yuan2020segfix} which assumes that most errors are at object boundaries, our approach does not make any such assumptions and can be applied to improve segmentation predictions at pixels away from boundaries, especially for lighter models (e.g. MobileNets~\cite{sandler2018mobilenetv2}). Sequential ensembles offer a principled way of conditioning on probability outputs from a given model. The sequential nature allows training more than one refinement models and offers trade-off between accuracy/speed at run-time.

% -----------------------begin method figure--------------------
\setlength{\belowcaptionskip}{-12pt}
\begin{figure*}
    \begin{center}
        \includegraphics[width=1\linewidth, height=0.2\linewidth]{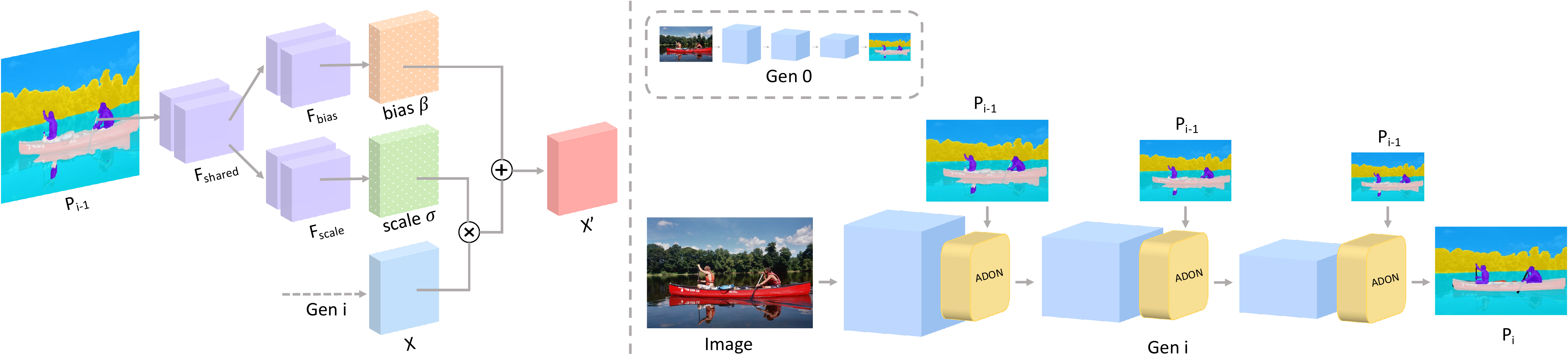}
    \end{center}
    \vspace{-12pt}
    \caption{\textit{Left}: ADaptive ModulatiON (ADON) Block (in \textcolor{gold}{gold}) learns parameters scale~$\boldsymbol{\sigma}$ and bias~$\boldsymbol{\beta}$ for spatial affine transformation of intermediate features $\mathbf{X}$ (in \textcolor{blue}{blue}) from the segmentation probability map $\mathbf{P}_{i-1}$. \textit{Right}: Introduction of ADON blocks at various depths allows conditioning on spatial probability maps at various levels within the network.}
    \label{fig:method_architecture}
\end{figure*}
% -----------------------end method figure--------------------

\textbf{Feature Modulation Methods:} Various forms of feature modulation methods have proven highly effective across a number of domains: Conditional Instance Norm~\cite{dumoulin2016learned, ghiasi2017exploring}, Adaptive Instance Norm~\cite{huang2017arbitrary} for neural style transfer, Dynamic Layer Norm~\cite{kim2017dynamic} for speech recognition, FiLM~\cite{perez2018film} for visual question answering, and MIMB~\cite{khirodkar2021multi, khirodkar2022occluded} for pose estimation. We show that feature-wise affine conditioning is effective in conditioning models in a cascade for high-resolution semantic segmentation.

%%%------------------------------------------------------------------------
%%%%%%%%% METHOD
\section{Method}
\label{sec:method}
Semantic segmentation aims to classify pixels in an input image $\image \in \realnum^{\imageheight \times \imagewidth \times3}$ into $\numclasses$ classes.
Most methods, \eg,~\cite{sun2019high, chen2017rethinking, zhao2017pyramid}, transform this problem to estimating per-pixel class probabilities to create a probability map $\probabilitymap \in \realnum^{\imageheight \times \imagewidth \times \numclasses}$ from $\image$ using a deep network $\deepnetwork$ such that $\probabilitymap = \deepnetwork(\image)$.
The segmentation network $\deepnetwork$ is trained in a supervised way to minimize loss $\loss(\groundtruthsegmap,\probabilitymap)$~\cite{jadon2020survey}, where $\groundtruthsegmap$ is the ground truth segmentation map of image $\image$. At inference, each pixel is assigned the label corresponding to the highest probability.

\textbf{Simple Ensembles:} Consider a pretrained model ensemble $\ensemble$ with $\nummodels$ segmentation models, $\ensemble = \{\ensemblemember_0, \ensemblemember_1, \dots, \ensemblemember_{\nummodels-1} \}$, where each ensemble member $\ensemblemember_\modelindex$ is independently trained.
The prediction $\probabilitymap$ of the ensemble $\ensemble$ on the input image $\image$ can be defined as follows:
\begin{eqnarray}
\probabilitymap_{\modelindex} & = &\ensemblemember_\modelindex(\image). \\
\probabilitymap & = &\ensemble(\image), \\
 & = &\ensemblecombination \big(\probabilitymap_0, \probabilitymap_1, \dots, \probabilitymap_{\nummodels-1} \big),
\end{eqnarray}
where $\ensemblecombination$ is a strategy to combine individual model predictions, \eg, %. For example, if output probabilities are averaged, $\ensemblecombination$ can be defined as
%\begin{eqnarray}
%\ensemblecombination_{\text{avg}} &=& \frac{1}{\nummodels} %\sum_{\modelindex=0}^{\nummodels-1} \probabilitymap_{\modelindex}.
%\end{eqnarray}
%In case of weighted average, the output probabilities can be combined using pre-defined weights $\mathbf{w}_\modelindex$ for each model $i$:
\vspace{-6pt}
\begin{eqnarray}
\ensemblecombination &=& \frac{1}{\nummodels} \sum_{\modelindex=0}^{\nummodels-1} \mathbf{w}_\modelindex \probabilitymap_{\modelindex},
\end{eqnarray}
where $\mathbf{w}_\modelindex$ can be uniform (average) or vary as a function of output probabilities (weighted average). Other strategies such as median averaging, majority voting, stacking~\cite{wolpert_nn_1992}, or selecting best model~\cite{dzeroski_ml_2004} can also be used to define $\ensemblecombination$. Designing an optimal $\ensemblecombination$ for semantic segmentation is crucial, yet challenging since different models in an ensemble may be more accurate on different regions of an image.

\subsection{Sequential Ensembles}
We propose a sequential ensembling strategy that avoids the combination rule $\ensemblecombination$ and \textit{learns} an optimal ensembling from the training data. 
Let $\sequentialensemble = \{\deepnetwork_0, \deepnetwork_1, \dots, \deepnetwork_{\nummodels-1}\}$ be $\nummodels$ segmentation models in the sequential ensemble. 
Each model $\deepnetwork_\modelindex$ forms a Markov chain of conditional dependence on the previous models $\deepnetwork_{\modelindex-1}, \dots \deepnetwork_0$ as follows.
\vspace*{-0.05in}
\begin{eqnarray}
\probabilitymap_0 & = & \deepnetwork_0(\image), \\
\probabilitymap_{\modelindex} & = & \deepnetwork_{\modelindex}(\image, \probabilitymap_{\modelindex-1}). 
 \end{eqnarray}
Each model in the sequential ensemble is trained to predict $\probabilitymap$ given the input image, conditioned on the previous model's prediction as shown in \cref{fig:introduction}. This allows each new model in the ensemble the opportunity to correct errors made by the previous model in the chain.

% --------------the main table!--------
\begin{table*}[t]
\captionsetup{font=small}
% \captionsetup{font=scriptsize}
\centering
\small
% \setlength{\tabcolsep}{2.7pt} %% controls col space, bigger => more space between cols
% \setlength{\tabcolsep}{5pt} %% controls col space, bigger => more space between cols
% \rowcolors{1}{}{lightgray} %%% alternating row grey color, comment to remove
% \setlength{\belowcaptionskip}{-10pt}
\resizebox{6.9in}{!}{
    \renewcommand{\arraystretch}{1.2} % Default value: 1, controls row space
    \begin{tabular}{@{}l|l| c c c c c c c c c c c| l |l| l| l@{}}
    \toprule[1.0pt]

    \multirow{2}{*}{\textbf{Method}} & \multirow{3}{*}{\textbf{Arch}} & \multicolumn{12}{c|}{\textbf{Cityscapes (val)}} &  \multirow{3}{*}{\textbf{ADE-20K}}  & \multirow{3}{*}{\textbf{COCO-Stuff}} & \multirow{3}{*}{\footnotesize{\textbf{PASCAL-Cnxt}}}\\

    & & \textbf{wall} & \textbf{fnce} & \textbf{tlgt} & \textbf{tsgn} &  \textbf{prsn} & \textbf{ridr} &  \textbf{trck} & \textbf{bus} & \textbf{train} & \textbf{mcyl} & \textbf{bcyl} & \textbf{mIoU} & & & \\
    
     \textcolor{gray}{Class (\%) $\rightarrow$} & & \textcolor{gray}{0.6} & \textcolor{gray}{0.7} & \textcolor{gray}{0.2} & \textcolor{gray}{0.6} &  \textcolor{gray}{1.1} & \textcolor{gray}{0.2} &  \textcolor{gray}{0.3} & \textcolor{gray}{0.3} & \textcolor{gray}{0.1} & \textcolor{gray}{0.1} & \textcolor{gray}{0.6} & & (val) & (test) & (test)\\

    \hline
    \footnotesize{MobileNetv3}~\cite{howard2019searching} & D8s & 39.4	& 47.0	& 45.1	& 63.1	& 71.1	& 45.1	& 45.7	& 64.7	& 51.5	& 39.3	& 67.0	& 64.1 & 29.7 & 24.1 & 32.7\\
    
    SIM-ENS & D8s 	&41.1	&49.1	&48.1	&65.4	&73.4	&46.4	&47.9	&68.2	&51.5	&44.1	&68.3	&65.4 (+1.3) & 31.4 (+1.7) & 26.3 (+2.2) & 34.2 (+1.5)\\
    
    SEQ-ENS \footnotesize{(Ours)} & D8s 	&\textbf{42.6}	&\textbf{52.4}	&\textbf{52.7}	&\textbf{68.3}	&\textbf{74.5}	&\textbf{48.8}	&\textbf{54.6}	&\textbf{70.8}	&\textbf{51.8}	&\textbf{47.9}	&\textbf{69.5}	&\textbf{68.7 (+4.6)} & \textbf{35.3 (+5.6)} & \textbf{28.6 (+4.5)}  & \textbf{36.9 (+4.2)}\\
    
    \hline
    \footnotesize{MobileNetv3}~\cite{howard2019searching} & D8 	&48.3	&51.4	&57.7	&69.6	&75.5	&50.1	&58.4	&73.5	&51.6	&50.3	&70.6	&69.5 & 31.8 & 26.8 & 34.4\\
    
    SIM-ENS & D8 	&49.0	&53.5	&60.4	&71.6	&76.4	&53.7	&62.1	&74.5	&53.7	&54.8	&73.1	&71.1 (+1.6) & 33.9 (+2.1) & 28.5 (+1.7) & 36.0 (+1.6)\\
    
    SEQ-ENS \footnotesize{(Ours)} & D8 	&\textbf{50.3}	&\textbf{56.8}	&\textbf{65.0}	&\textbf{78.9}	&\textbf{78.1}	&\textbf{57.8}	&\textbf{68.9}	&\textbf{75.1}	&\textbf{57.4}	&\textbf{59.4}	&\textbf{75.4}	&\textbf{73.6 (+4.1)} & \textbf{35.6 (+3.8)} & \textbf{30.0 (+3.2)} & \textbf{37.3 (+2.9)}\\

    \hline\hline
    HRNet~\cite{sun2019high} & H-18s 	&53.2	&61.5	&70.3	&77.9	&81.1	&59.2	&66.4	&84	 &70.5	&59.7	&76.1	&76.2 & 33.1 & 29.4 & 38.9 \\
    
    SIM-ENS & H-18s 	&56.3	&61.1	&71.8	&79.4	&82.4	&61.6		&69.4	&84.4	&67.5	&61.9	&77.0	&77.1 (+0.9) & 34.5 (+1.4) & 30.0 (+0.6) & 40.4 (+1.5) \\
    
    SEQ-ENS \footnotesize{(Ours)} & H-18s 	&\textbf{59}	&\textbf{63.9}	&\textbf{73.6}	&\textbf{79.6} &\textbf{82.8}	&\textbf{62.2}	&\textbf{75.5}	&\textbf{88}	&\textbf{78}	&\textbf{63.4}	&\textbf{78.2}	&\textbf{78.9 (+2.7)} & \textbf{37.6 (+4.5)} & \textbf{32.7 (+3.3)} & \textbf{42.0 (+3.1)}\\
    
    \hline
    HRNet~\cite{sun2019high} & H-18 	&59.1	&63.7 &73.2	&80.6	&82.8	&61.8	&75.9	&87.7	&75.5	&60.4	&77.3	&78.7 & 36.8 & 33.0 & 42.6\\
    
    SIM-ENS & H-18 	&59.2	&63.6	&74.1	&81.1	&83.1	&62.0	&76.5	&87.8	&74.3	&60.7	&78.1	&79.0 (+0.3) & 37.6 (+0.8) & 34.1 (+1.1) & 43.9 (+1.3)	\\
    
    SEQ-ENS \footnotesize{(Ours)} & H-18 	&\textbf{58.0}	&\textbf{64.6}	&\textbf{75.0}	&\textbf{82.0}	&\textbf{84.0}	&\textbf{63.2}	&\textbf{79.1}	&\textbf{90.1}	&\textbf{78.6}	&\textbf{62.7}	&\textbf{79.1}	&\textbf{79.8(+1.1)} & \textbf{40.9 (+4.1)} & \textbf{35.7 (+2.7)} & \textbf{46.1 (+3.5)}\\
    \hline
    
    HRNet~\cite{sun2019high} & H-48 	&56.4	&65.6	&75.5	&81.9	&84.1	&66.3	&79.7	&89.8	&84.2	&68.8	&80.1	&80.5 & 42.0 & 38.3 & 51.1\\
    
    SIM-ENS & H-48 	&56.6	&65.1	&75.8	&82.1	&84.2	&67.1	&79.8	&89.9	&84.6	&68.3	&79.9	&80.6 (+0.1) & 42.5 (+0.5) & 39.7 (+1.4) & 52.1 (+1.0)\\
    
    SEQ-ENS \footnotesize{(Ours)} & H-48 	&\textbf{58.9}	&\textbf{66.9}	&\textbf{76.8}	&\textbf{83.2}	&\textbf{84.8}	&\textbf{68.3}	&\textbf{80.1}	&\textbf{90.6}	&\textbf{85.6}	&\textbf{68.8}	&\textbf{80.9}	&\textbf{81.3 (+0.8)} & \textbf{45.6 (+3.6)} & \textbf{40.8 (+2.5)} & \textbf{53.8 (+2.7)}\\
    
    \hline\hline
    DeepLabv3+~\cite{chen2018encoder} & R-18 	&51.4	&58.2	&69.9	&77.6 &81.3	&60.8	&76.1	&85.4	&72.5	&63.1	&76.3	&76.8 & 34.1 & 29.7 & 43.5\\
    
    SIM-ENS & R-18	&51.1	&58.3	&71.5	&78.9	&82.1	&61.0	&77.8	&87.3	&74.5	&64.3	&76.9	&77.8 (+1.0) & 35.4 (+1.3) & 31.2 (+1.5) & 45.8 (+2.3)\\
    
    SEQ-ENS \footnotesize{(Ours)} & R-18 	&\textbf{51.2}	&\textbf{59.4}	&\textbf{74.6}	&\textbf{80.8}	&\textbf{83.6}	&\textbf{62.8}	&\textbf{79.8}	&\textbf{88.1}	&\textbf{77.0}	&\textbf{67.2}	&\textbf{78.3}	&\textbf{78.9 (+2.1)} & \textbf{37.6 (+3.5)} & \textbf{35.3 (+4.1)} & \textbf{46.2 (+2.7)}\\
    
    \hline
    DeepLabv3+~\cite{chen2018encoder} & R-50 	&51.2	&62.5	&74.3	&82.2	&84.2	&65.8	&81.1	&88.8	&84.7	&68.6	&79.5	&79.8 & 42.7 & 37.4 & 50.4\\
    
    SIM-ENS & R-50 	&51.8	&62.8	&74.5	&82.8	&84.6	&65.9		&\textbf{81.4}	&89.8	&85.1	&69.2	&79.8	&80.1 (+0.3) & 43.3 (+0.6) & 38.2 (+0.8) & 51.8 (+1.4)\\
    
    SEQ-ENS \footnotesize{(Ours)} & R-50  &\textbf{53.4}	&\textbf{63.2}	&\textbf{75.6}	&\textbf{83.5}	&\textbf{85.0}	&\textbf{66.3}		&79.3	&\textbf{91.9}	&\textbf{85.2}	&\textbf{70.0}	&\textbf{80.2}	&\textbf{80.7 (+0.9)} & \textbf{45.1 (+2.4)} & \textbf{39.2 (+1.8)} &  \textbf{52.4 (+2.0)}\\
    
    \hline
    DeepLabv3+~\cite{chen2018encoder} & R-101 	&54.9	&64.4	&74.6	&81.9	&84.6	&67.7	&85.2	&91.8	&85.2	&71.3	&80.3	&80.9  & 44.6 & 38.8 & 53.2\\
    
    SIM-ENS & R-101 	&55.1	&64.6	&74.7	&82.0	&84.7	&67.8	&85.0	&91.9	&85.8	&71.9	&80.3	&81.1 (+0.2) & 45.0 (+0.4) & 39.0 (+0.2) & 54.0 (+0.8)\\
    
    SEQ-ENS \footnotesize{(Ours)} & R-101 	&\textbf{56.0}	&\textbf{64.6} &\textbf{75.2}	&\textbf{82.6}	&\textbf{84.9}	&\textbf{68.5}	&\textbf{85.4}	&\textbf{92.3}	&\textbf{86.5}	&\textbf{72.5}	&\textbf{80.5}	&\textbf{81.5 (+0.6)} & \textbf{46.8 (+2.2)} & \textbf{40.3 (+1.3)} & \textbf{55.1 (+1.9)}\\

    \bottomrule[1.0pt]
    
    \end{tabular}
}
\vspace*{-0.1in}
\caption{Comparison of SEQ-ENS (Ours) with SIM-ENS using $N=2$ on various datasets. R-@ and H-@ stand for ResNet-@ and HRNet-W@ respectively. D8 is the output stride of DeepLabv3+. \texttt{s} denotes the \textit{small} version of the backbone. We report class-wise mIoU for 11 rare classes in Cityscapes. For rare classes such as \textit{motor-cycle (mcycl)}, SEQ-ENS improves mIOU by $\approx 2X$ compared to SIM-ENS.}
\label{table:sim_ens_compare}
\vspace*{-0.2in}
\end{table*}
% -------------------------

\subsection{Adaptive Modulation Block}
A key challenge is designing the architecture in order to take the previous generation's probabilities as a conditioning input. A na\"{i}ve \textit{early fusion} approach would be to simply concatenate the input image $\mathbf{I}$ with the probability map $\probabilitymap_{i-1}$ corresponding to the output of the previous generation. Similarly, \textit{late fusion} would concatenate feature maps from later layers within the network with appropriately down-sampled probability map $\probabilitymap_{i-1}$. However, both of these approaches fail to improve performance. 

We describe the ADaptive modulatiON (ADON) block that can be easily introduced in any existing feature extraction backbone to overcome this issue (see \cref{fig:method_architecture}). ADON allows spatial modulation of intermediate feature maps using the conditioning input $\probabilitymap_{i-1}$. The $\modelindex^{\text{th}}$ generation model $\deepnetwork_{\modelindex}$ in the sequential ensemble uses ADON blocks to leverage information from the prediction $\probabilitymap_{\modelindex-1}$ of the previous generation model. 
Similar to the Batch Normalization~\cite{ioffe2015batch}, ADON learns to adaptively influence the output of the neural network by applying an affine transformation to the network's intermediate features based on $\probabilitymap_{\modelindex-1}$.

Let $\intermediatefeaturemap \in \realnum^{\featuremapheight \times \featuremapwidth \times \featuremapchannels}$ be an intermediate feature map in the segmentation network $\deepnetwork_\modelindex$. 
The ADON block consists of operations $\convoperation_\text{shared}$, $\convoperation_\text{scale}$ and $\convoperation_\text{bias}$ on the probability map $\probabilitymap_{\modelindex-1}$ from the previous generation $\deepnetwork_{\modelindex-1}$. 
The modulation outputs $\modulationscaleoutput$ and $\modulationtranslationoutput$ are used to element-wise scale and bias the intermediate feature map $\intermediatefeaturemap$ to produce $\intermediatefeaturemap^\prime$. 
We implement the operations $\convoperation_\text{shared}$, $\convoperation_\text{scale}$ and $\convoperation_\text{bias}$ using a simple two-layer convolutional network, whose design is in the supplemental material. First, the probability map $\probabilitymap_{\modelindex-1} \in \realnum^{\imageheight \times \imagewidth \times \numclasses}$ is spatially downsampled to match the 2D resolution of $\intermediatefeaturemap$. The operation $\convoperation_\text{shared}$ then maps $\probabilitymap_{\modelindex-1} \in \realnum^{\featuremapheight \times \featuremapwidth \times \featuremapchannels}$ to a $K$-dimensional latent space,  $\mathbf{e} \in  \realnum^{\featuremapheight \times \featuremapwidth \times K}$. We use this latent space to predict the scale and bias parameters jointly using $\convoperation_\text{scale}$ and $\convoperation_\text{bias}$ where $\modulationscaleoutput, \modulationtranslationoutput \in \realnum^{\featuremapheight \times \featuremapwidth \times \featuremapchannels}$.
\vspace*{-0.05in}
\begin{eqnarray}
\sharedoutput & = &\convoperation_\text{shared}(\probabilitymap_{\modelindex-1}), \\
\modulationscaleoutput & = &\convoperation_\text{scale}(\sharedoutput), \\
\modulationtranslationoutput & = &\convoperation_\text{bias}(\sharedoutput), \\
\intermediatefeaturemap^\prime & = & \modulationscaleoutput * \intermediatefeaturemap + \modulationtranslationoutput.
\end{eqnarray}
Thus, ADON is a spatial generalization of channel-wise feature modulation blocks proposed in~\cite{hu2018squeeze, khirodkar2021multi}. 
As the modulation parameters are adaptive to the spatially variant input probability maps, the proposed ADON block is an effective way of injecting segmentation information at multiple layers within the network in comparison to early/late fusion~\cite{gadzicki2020early}. 
ADON blocks inserted early in the network capture finer details such as object boundaries and those inserted late resolve the class confusion between similar looking classes.

%%%------------------------------------------------------------------------
%%%%%%%%% EXPERIMENTS
\section{Experiments}
\label{sec:experiments}

We evaluate our approach on the following datasets.

% \noindent
\textbf{Cityscapes}~\cite{cordts2016cityscapes} is a real world driving dataset that consists of 2975 \textit{train}, 500 \texttt{val} and 1525 \texttt{test} images with resolution $2048 \times 1024$. The dataset contains $19$ semantic categories for the segmentation task.

% \noindent
\textbf{ADE20K}~\cite{zhou2017scene} is used in ImageNet scene parsing challenge 2016 consisting of around 20k \texttt{train}, 2k \texttt{val}, 3k \texttt{test} images spanning 150 fine-grained semantic categories and diverse scenes.

% \noindent
\textbf{COCO-Stuff}~\cite{caesar2018cvpr} is a challenging scene parsing dataset that contains 171 semantic classes. We use the smaller version with 10k images. The \texttt{train} set and \texttt{test} set consists of 9k and 1k images respectively.

% \noindent
\textbf{PASCAL-Context}~\cite{mottaghi_cvpr14} consists of 59 semantic classes and 1 background label.
The dataset contains 4998 \texttt{train} and 5105 \texttt{test} images. We follow the standard testing procedure~\cite{sun2019high}. The image is resized to $480 \times 480$ and then fed into our network.
The resulting $480 \times 480$ label maps are then resized to the original image size.

% \noindent
\textbf{Implementation details.} We use the \textit{mmsegmentation}\footnote{\url{https://github.com/open-mmlab/mmsegmentation}} codebase for the implementation of various model families. We use the available pretrained models as $\deepnetwork_0$ in the sequential ensemble, and focus on improving their performance by adding additional generations.
For fairness, the models in the later generations are trained with the same hyper-parameters as the $\deepnetwork_0$ model. We use the imagenet pretrained backbone for all our experiments. For SIM-ENS, the segmentation head of the backbone is randomly initialized across multiple runs. During training, we apply data augmentation using: random resize with ratio $0.5-2.0$, random horizontal flipping and random cropping for all datasets. The models are trained using either SGD~\cite{ruder2016overview} or AdamW~\cite{kingma2014adam} optimizer for 160k iterations. The learning rate is set to an initial value and then decayed using a polynomial LR schedule. We report semantic segmentation performance using mean Intersection over Union (mIoU) for all datasets. For the HRNet~\cite{sun2019high} family, we insert $10$ ADON blocks for Cityscapes and Pascal-Context datasets and $5$ ADON blocks for ADE20K and COCO-Stuff datasets. We use $K=128$ for all our experiments. Further, unless mentioned otherwise we report results using inference at single-scale and no flipping. Please refer to the supplemental for more details.

% ------------------------------------------
\vspace*{-0.05in}
\subsection{Comparisons with Simple Ensembles}
 We compare SEQ-ENS and SIM-ENS in Tab.~\ref{table:sim_ens_compare} for $N=2$. We benchmark three model families, MobileNetv3~\cite{howard2019searching}, HRNet~\cite{sun2019high}, and DeepLabv3+~\cite{chen2018encoder}, with different backbones. SEQ-ENS consistently improves over SIM-ENS in all settings; relative improvements over baseline with respect to SIM-ENS range from $2$x to $8$x. 

\begin{figure*}
\centering
\includegraphics[width=1\linewidth, height=0.4\linewidth]{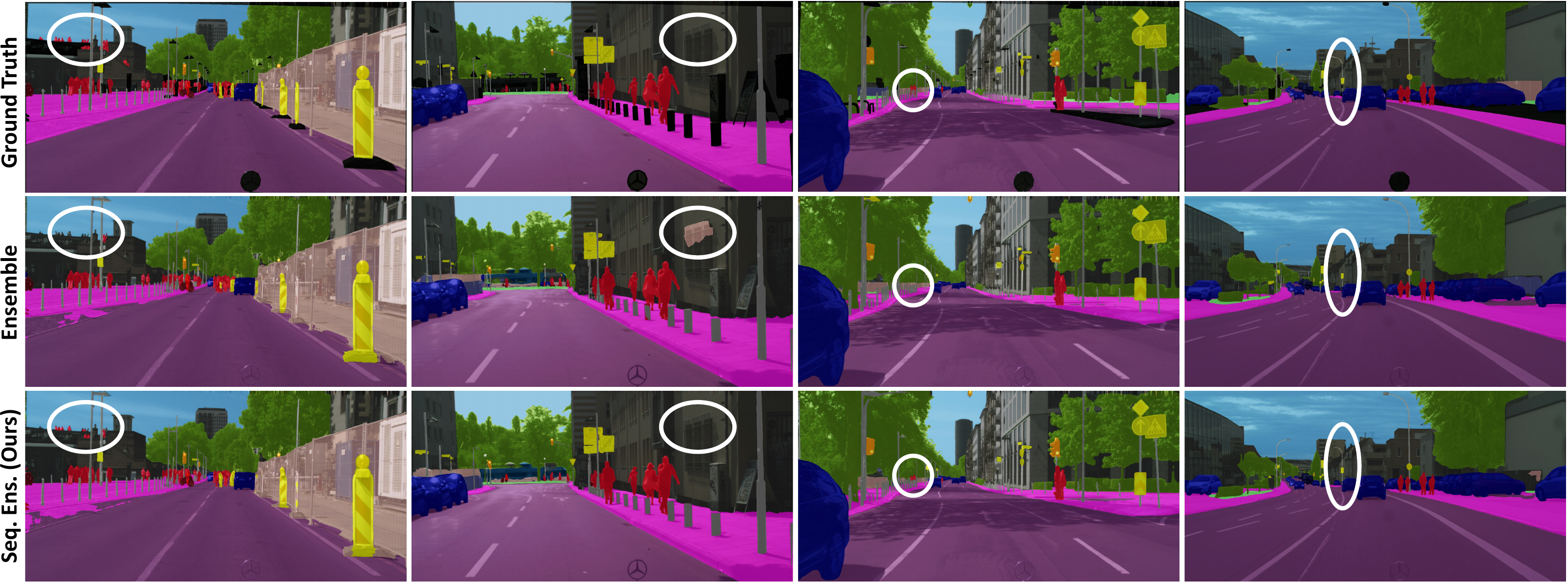}
\includegraphics[width=1\linewidth, height=0.4\linewidth]{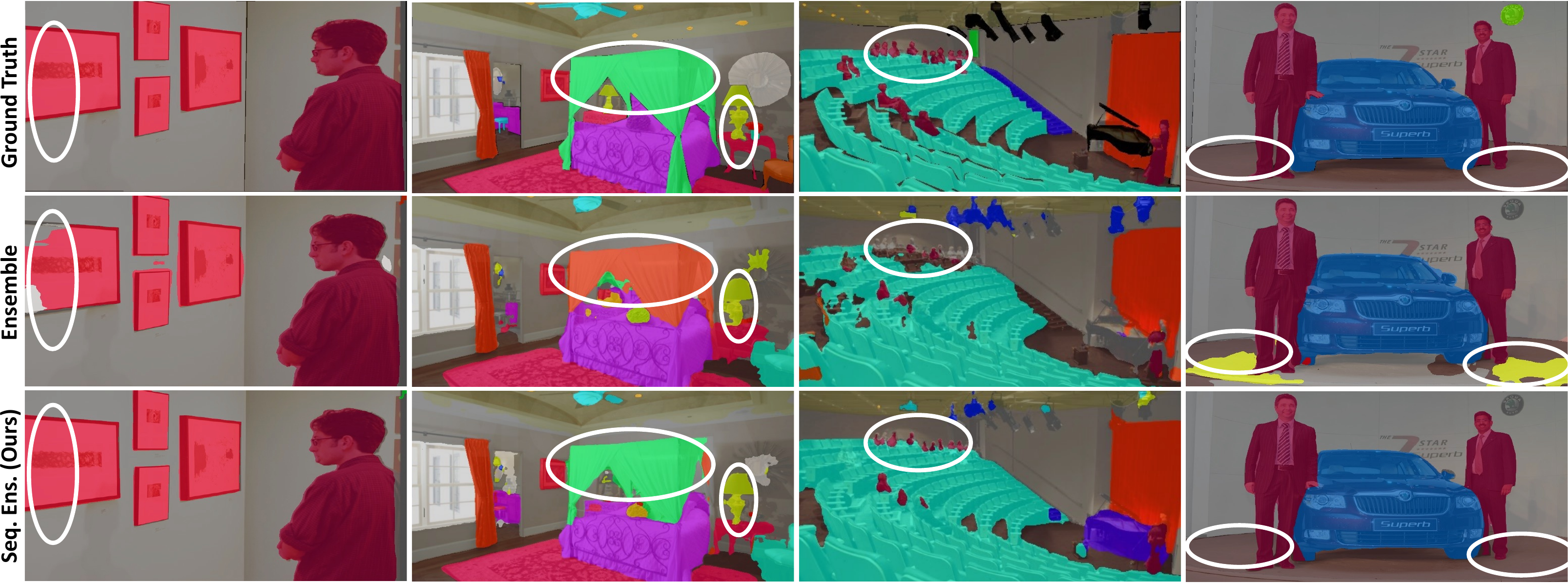}
\includegraphics[width=1\linewidth, height=0.4\linewidth]{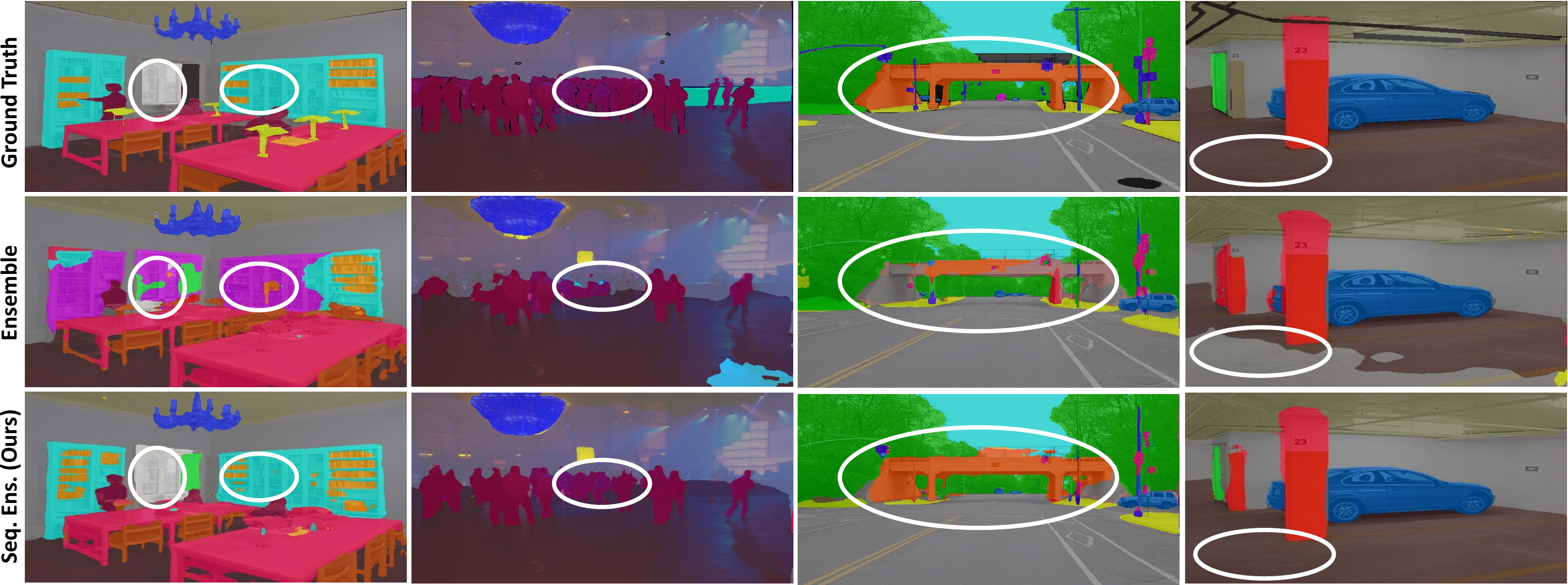}
\caption{Qualitative comparison of Sequential Ensembles and Simple Ensembles. Cityscapes (first three rows) and  ADE20K (last six rows) for HRNet-W48(N=2). The white eclipses highlight the fine-grained details that our approach captures in comparison to the baseline.}
\label{fig:qualitative}
\end{figure*}

On Cityscapes, SEQ-ENS shows consistent gains across all backbones over SIM-ENS. Gains from SEQ-ENS tend to be more exaggerated for scenarios susceptible to underfitting compared to single models or SIM-ENS. For example, mIoU improves most significantly for rare categories, such as \textit{train} and \textit{motorcycle}, and categories with fine structures, such as \textit{pole} and \textit{fence}. This shows SEQ-ENS distributes model capacity intelligently to focus on underrepresented classes and finer details. Our improvements are especially pronounced when ensembling light models like MobileNetv3 or evaluating on complex datasets with more classes. For example, for H-18s, SEQ-ENS outperforms SIM-ENS by +1.6 mIOU for Cityscapes ($19$ categories) and +1.6 for PASCAL-Context ($59$ categories), but by +3.1 for ADE-20K ($150$ categories) and +2.7 for COCO-Stuff ($171$ categories). We show qualitative results in \cref{fig:qualitative}.

% -------------------------
% --------------the main table!--------
\begin{table}[t]
   \centering
\captionsetup{font=small}
\setlength{\tabcolsep}{1.5pt} %% controls col space, bigger => more space between cols
  \vspace*{-0.3in}
    \resizebox{3.0in}{!}{
    \small
    \renewcommand{\arraystretch}{1.2} % Default value: 1
  \begin{tabular}{@{}c|c c | c c c c c@{}}
    \toprule[1.0pt]
  \multirow{2}{*}{\textbf{\# Models}} & \multicolumn{2}{c|}{\textbf{SIM-ENS}}  & \multicolumn{5}{c}{\textbf{SEQ-ENS}}  \\
  & mIoU & Params & mIoU & Params & GFLOPs & FPS & Train-Epochs \\
    \hline
   N = 1 & 19.7 & 9.8 & 19.7 & 9.8 & 39.6 &  65  & 127 \\
   N = 2 & 20.6 & 19.6 & 23.1 & 20.4 & 80.6 & 36 & 201 \\
   N = 4 & 22.4 & 39.2 & 27.6 & 41.6 & 163.2 & 20 & 311 \\
   N = 6 & 23.3 & 58.8 & 31.4 & 62.8 & 245.6 & 15 & 387 \\
   N = 8 & 23.9 & 78.4 & 33.1 & 84.0 & 328.0 & 11 & 440 \\
   N = 10 & 24.8 & 98.1 & 34.2 & 105.2 & 410.4 & 9 & 495 \\    
    \bottomrule[1.0pt]
    \end{tabular}
    }
    \vspace*{-0.1in}
    \caption{Metrics with varying $N$ for SIM-ENS and SEQ-ENS on ADE20K \texttt{val} set using FCN MobileNetv2-D8~\cite{sandler2018mobilenetv2}.}
    \label{table:dynamic_sim_ens_compare}
    \vspace*{-0.2in}
\end{table}
% -------------------------
Further, we report various metrics like mIoU, parameters (M), GFLOPs, test-time FPS and total training epochs with varying ensemble size (N) for SIM-ENS and SEQ-ENS in \cref{table:dynamic_sim_ens_compare} using MobileNetv2~\cite{sandler2018mobilenetv2}. As can be seen, later generations require fewer epochs for training in SEQ-ENS due to quicker convergence (495 vs. 127 epochs for $10^{th}$ vs. $1^{st}$ generation). Finally, our method allows for a dynamic trade-off between accuracy and latency by varying the ensemble size during inference ---a desirable property for edge devices.
\label{sec:experiments_sim-ens}

% ------------------------------------------
 \vspace*{-0.2in}
\subsection{Improving State-of-the-Art}

We improve the state-of-the-art among methods that do not use extra data on the Cityscapes and ADE20K \texttt{val} sets. We construct a sequential ensemble of size $3$ using Segformer MiT-B5~\cite{xie2021segformer} as $G_0$ followed by $2$ generations of HRNet-W48. Our models are trained at the same resolution and learning rate schedule as the Segformer and use test-time augmentation at $7$ scales with left-right flipping. SEQ-ENS improves the prior art by $0.8$ on Cityscapes and $2.2$ on ADE20K, as shown in \cref{table:sota_compare}. SEQ-ENS with fewer parameters outperforms Swin-Trans~\cite{liu2021swin}.

\begin{table}[hp]
% \captionsetup{font=scriptsize}
\captionsetup{font=small}
\vspace*{-0.1in}
    \centering
    \small
    \resizebox{3.4in}{!}{
        \setlength{\tabcolsep}{1.2pt}
        \renewcommand{\arraystretch}{1.2} % Default value: 1, controls row space
        \begin{tabular}{@{}l|c|c|c c c|c c c@{}}
        \toprule[1.0pt]
          \multirow{2}{*}{\textbf{Method}} & \multirow{2}{*}{\textbf{Arch}} & \multirow{2}{*}{\textbf{Params}} & \multicolumn{3}{c|}{\footnotesize{\textbf{Cityscapes}}}  &  \multicolumn{3}{c}{\footnotesize{\textbf{ADE-20K}}}\\
          & & & \footnotesize{FLOPs} $\downarrow$ & FPS $\uparrow$ & mIoU $\uparrow$ & \footnotesize{FLOPs} $\downarrow$ & FPS $\uparrow$ & mIoU $\uparrow$\\
          \hline

  FCN~\cite{long2015fully} & ResNet-101                             &  68.6M  &  2203G  &  1.2  &  76.6  &  276G  &  14.8  &  41.4  \\
  EncNet~\cite{zhang2018context} & ResNet-101                       &  55.1M  &  1748G  & 1.3  &  76.9  &  219G  & 14.9 & 44.7 \\
  PSPNet~\cite{zhao2017pyramid} & ResNet-101                        &  68.1M  &  2049G & 1.2  &  78.5  &    256G  & 15.3 & 44.4 \\
  CCNet~\cite{huang2019ccnet} & ResNet-101                          &  68.9M  &  2225G & 1.0  &  80.2  & 278G & 14.1 & 45.2 \\
  DeeplabV3+~\cite{chen2018encoder} & ResNet-101                    &  62.7M  &  2032G & 1.2  &  80.9  & 255G & 14.1 & 44.1 \\
  OCRNet~\cite{yuan2020object} & HRNet-W48                          &  70.5M  &  1297G & 4.2   &  81.1  & 165G & 17.0 & 45.6 \\
  GSCNN~\cite{takikawa2019gated} & \footnotesize{WideReNet38}       &  -      &  -   & - &  80.8  & - & - & - \\
  Ax-DeepLab~\cite{wang2020axial} & \footnotesize{AxialResNet-XL}   &  -      &  2447G  & -  &  81.1  & - & - & - \\
  \footnotesize{Dynamic Routing} & \footnotesize{Dyn-L33-PSP}       &  -      &  270G  &  -  &  80.7  & - & - & - \\
  Auto-Deeplab~\cite{li2020learning} & \footnotesize{NAS-F48-ASPP}  &  -      &  695G  &  -     &  80.3  & - & - & 44.0 \\
  SETR~\cite{zheng2021rethinking} & ViT-Large                       &  318.3M &  -  &  0.5     &  82.2  & - & 5.4 & 50.2 \\
  SegFormer~\cite{xie2021segformer} & MiT-B5                        &  84.7M  &  1448G &  2.5     &  84.0 & 183G & 9.8 & 51.8   \\
  Swin Trans.~\cite{liu2021swin} & Swin-L                           &  234.9M &  -      & -     & -      & 3230G & 6.2 & 53.5 \\
          \hline
          
  \textbf{SEQ-ENS} (Ours) & MiT-B5, H-18            & 94.3M     & 1596G &  2.2  &  84.3 & 278G & 8.6 &  52.5 \\
  \textbf{SEQ-ENS} (Ours) & MiT-B5, H-48            & 150.5M    & 1635G  & 2.0  &  84.6 & 323G  & 8.3 &53.8 \\
  \textbf{SEQ-ENS} (Ours) & MiT-B5, (H-48)x2    & 216.3M    & 1822G  & 1.7  &  \textbf{84.8} & 464G & 7.8 & \textbf{54.0} \\
        \bottomrule[1.0pt]
        \end{tabular}
    }
    \vspace{-0.1in} %% added manually for ECCV    
    \caption{Improving the state-of-the-art methods using sequential ensembling on Cityscapes and ADE-20K \texttt{val} set.  }
    \label{table:sota_compare}
    \vspace*{-0.2in}
\end{table}

\label{sec:experiments_sota}

% ------------------------------------------
 \vspace*{-0.05in}
\section{Analysis}

%%%%----------------------------------------------------------
We first show ablation studies with respect to ADON block placement and parameters and then discuss several interesting features of our framework.

\noindent
\textbf{ADON Block Placement:} We analyse the effect of adding ADON blocks at various depths in the HRNet-W18s backbone (having 4 layers) in Tab.~\ref{table:adon_placement}. Specifically, we investigate three strategies for feature modulation: Layer 1 (\textit{early}), Layer 2 (\textit{middle}) and Layer 3 (\textit{late}). We observe that on the Cityscapes dataset, $G_0$ model predictions lack low-level image details like object boundaries and inserting the ADON block in early layers performs best. On the other hand, on ADE20K dataset with $150$ categories, a majority of $G_0$ segmentation errors are due to class-confusion. In this case, inserting the ADON block into later layers of the backbone helps resolve class confusion.

% \begin{table}[H]
% % \captionsetup{font=small}
% \small

% \begin{center}
%     \setlength{\tabcolsep}{2.0pt}
%     \renewcommand{\arraystretch}{1.2}
%     \rowcolors{3}{}{lightgray}
%     \begin{tabular}{@{}l|c c|c c@{}}
%     \Xhline{3\arrayrulewidth}
%     \multirow{2}{*}{\textbf{Position}} & \multicolumn{2}{c|}{\textbf{Cityscapes}}   & \multicolumn{2}{c}{\textbf{ADE20K}}\\
%      & \scriptsize{\texttt{single-scale}} & \scriptsize{\texttt{multi-scale}} & \scriptsize{\texttt{single-scale}} & \scriptsize{\texttt{multi-scale}}\\
%     \hline
%     \textit{Stage-1} & \textbf{78.2}  &  \textbf{79.8}  & 34.0 & 35.1\\
%     \textit{Stage-2} &  77.8  & 79.1 & 35.4 & 36.5\\
%     \textit{Stage-3} &  77.1 & 78.4 & \textbf{36.7} & \textbf{38.3}\\
%     All Stages &  77.1 & 78.4 & \textbf{36.7} & \textbf{38.3}\\
%     \Xhline{3\arrayrulewidth}
%     \end{tabular}

% \caption{We analyse the effect of inserting ADON blocks at three depths (\textit{stage: 1, 2, 3}) in the HRNet-W18s~\cite{sun2019high} backbone. ADE20K with predominantly class-confusion errors benefit from \textit{late} block insertion whereas Cityscapes benefit from \textit{early} insertion capturing high-resolution segmentation details.}
% \label{table:adon_placement}
% \end{center}
% \end{table}

\begin{table}[H]
\captionsetup{font=small}
\small

\begin{center}

\resizebox{1.6in}{!}{
    \setlength{\tabcolsep}{2.0pt}
    \renewcommand{\arraystretch}{1.2}
    \rowcolors{3}{}{lightgray}
    \begin{tabular}{@{}l|c|c@{}}
    \Xhline{3\arrayrulewidth}
    \textbf{Ablation} & \textbf{Cityscapes} & \textbf{ADE20K}\\
    \hline
    $G_0$ & 76.2  & 33.1 \\
    \textit{Layer-1} & 78.2 & 34.0 \\
    \textit{Layer-2} &  77.8 & 35.4 \\
    \textit{Layer-3} &  77.1 & 36.7 \\
    \textit{All Layers ($G_1$)} &  \textbf{78.9} & \textbf{37.6} \\
    \Xhline{3\arrayrulewidth}
    \end{tabular}
    }
\resizebox{1.6in}{!}{
    \thinspace
    \setlength{\tabcolsep}{2.0pt}
    \renewcommand{\arraystretch}{1.2}
    \rowcolors{3}{}{lightgray}
    \begin{tabular}{@{}l|c|c@{}}
    \Xhline{3\arrayrulewidth}
    \textbf{Method} & \textbf{Cityscapes} & \textbf{ADE20K}\\
    \hline
    $G_0$ & 78.2  & 34.6 \\
    SIM-ENS & 78.6 & 35.1 \\
    SEQ-ENS &  \textbf{80.6} & \textbf{39.0} \\
    \Xhline{3\arrayrulewidth}
    \end{tabular}
    }

\caption{(Left) Effect of inserting ADON blocks at various layers in the HRNet-W18s~\cite{sun2019high} backbone with single scale testing. (Right) Test-time augmentation using six-scales and left-right flipping for $N=2$. Note that our $G_1$ model with single scale testing (two inferences) outperforms $G_0$ with multi-scale testing (12 inferences).}
\label{table:adon_placement}
\vspace*{-0.25in}
\end{center}
\end{table}

%%%%----------------------------------------------------------
% \noindent
% \textbf{ADON Block Architecture:} We vary the dimension $K$ of the latent space of the ADON block used for modulation on the Cityscapes dataset using various backbones as shown in Tab.~\ref{table:adon_architecture}. We observe that increasing $K$ improves performance across backbones, with $K=128$ achieving the optimal balance between the parameter overhead and segmentation performance.

% \input{tables/adon_architecture}
%%%%----------------------------------------------------------

% -------------------------------------------
\begin{figure}
\centering
    \includegraphics[width=0.9\linewidth]{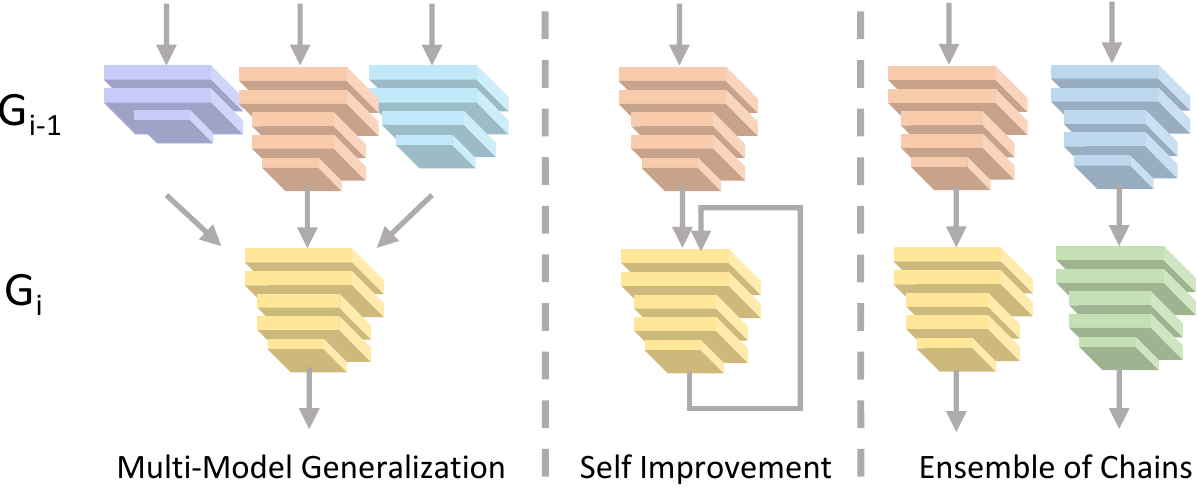}
\caption{Sequential ensembles allows segmentation prediction conditioning between generations in various ways.}
\label{fig:schema}
\vspace*{-0.2in}
\end{figure}

% %%%%----------------------------------------------------------
% \noindent
% \textbf{Effect of Model Capacity:} SEQ-ENS is most effective with light-weight $G_0$ models as shown in \cref{table:cityscapes_backbone_trend}. We also report increase in number of model parameters.
% \input{tables/cityscapes_backbone_trend}

%%%%----------------------------------------------------------
\noindent
\textbf{Multi-Model Generalization:} Our SEQ-ENS method can be \textit{generalized} to multiple architectures, where $G_i$ is trained to randomly condition on probability inputs $P_{i-1}$ from \textit{different} $G_0$ segmentation models (see \cref{fig:schema} (\textit{left})). By training a single $G_i$, during inference we see improved performance for all the different $G_0$s. In \cref{table:generalization} we show that using HRNet-W18s~\cite{sun2019high} backbone, $G_1$ improves the performance of all three backbones in the HRNet family on Cityscapes and ADE20K datasets. The \textit{generalized} sequential ensembles also outperforms simple ensemble baseline and have comparable performance to the default scenario of individual sequential ensembles using one-third number of parameters.

\begin{table}[h]
\small
\captionsetup{font=small}
\begin{center}
    \setlength{\tabcolsep}{4.0pt}
    \renewcommand{\arraystretch}{1.2}
    \rowcolors{3}{}{lightgray}
    \begin{tabular}{@{}l|c c l|c c c@{}}
    \Xhline{3\arrayrulewidth}
    \multirow{2}{*}{\textbf{Arch}} & \multicolumn{3}{c|}{\textbf{Cityscapes}}   & \multicolumn{3}{c}{\textbf{ADE20K}}\\
         & $G_0$ & $G_1$  & $G_1$  & $G_0$ & $G_1$  & $G_1$ \\    
         &  & \tiny Default & \tiny Generalized &  & \tiny Default & \tiny Generalized\\

    \hline
    \textbf{H-18s} & 76.2  & 78.9  & 78.1 & 33.1 & 37.6 & 37.0  \\
    \textbf{H-18} &  78.7  & 79.8 & 79.4  & 36.8 & 39.1  & 38.5 \\
    \textbf{H-48} &  80.5  & 80.7 & 80.5 & 42.0 & 42.6 & 42.3  \\
    
    \Xhline{3\arrayrulewidth}
    \end{tabular}

\caption{Comparison of \textit{generalized} vs default sequential ensembles using HRNet-W18s~\cite{sun2019high} backbone as $G_1$. Our approach allows improving performance by training $G_1$ models for each of the three architectures separately (default scenario). A \textit{single} generalized $G_1$ model can give similar performance with $33\%$ parameters.}
\label{table:generalization}
\vspace*{-0.1in}
\end{center}
\end{table}

\begin{figure}
\centering
    \includegraphics[width=1\linewidth]{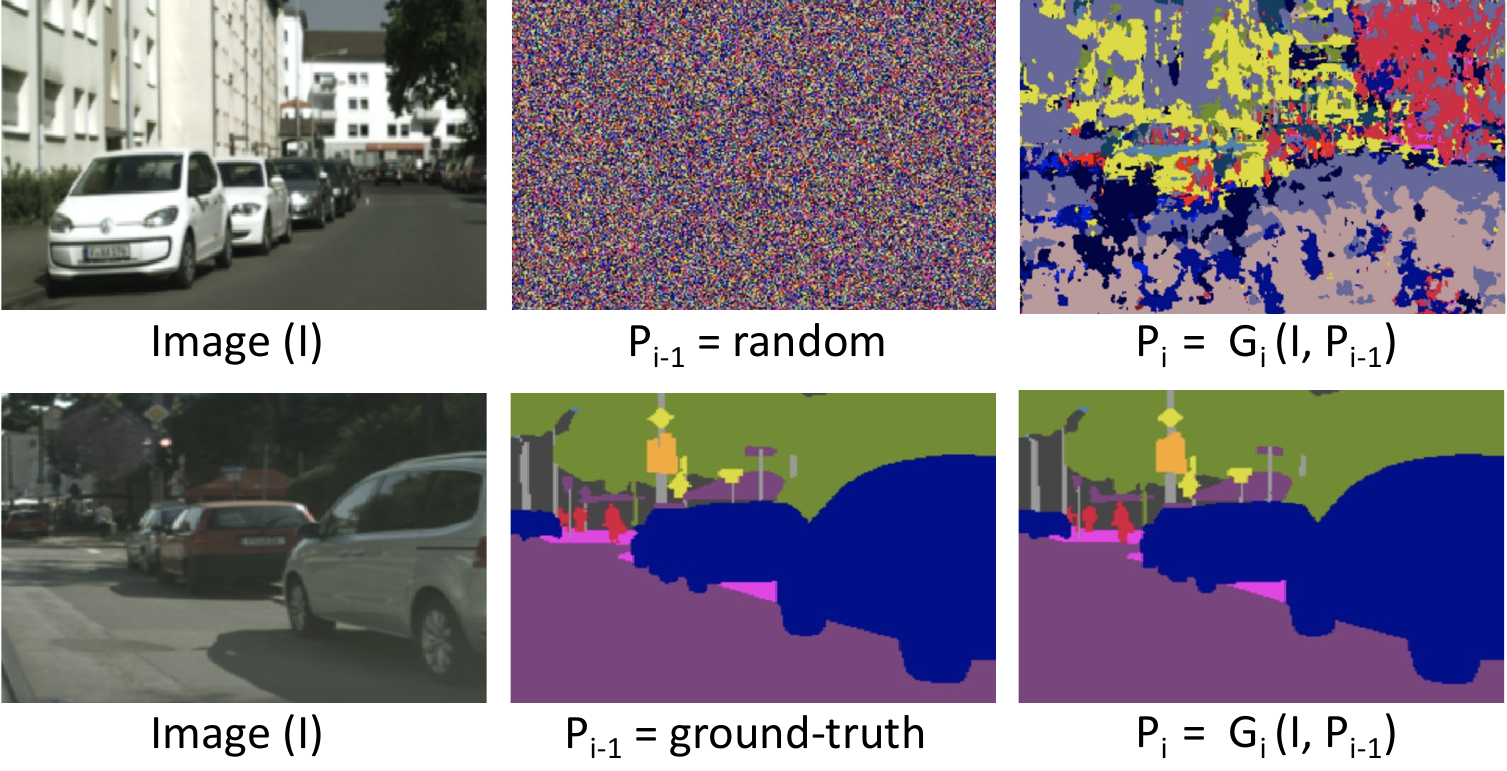}
\caption{$\mathbf{G}_i$ refines/utilizes conditional probabilities $\mathbf{P}_{i-1}$. Top row shows an extreme example where $G_i$ refines random input probabilities. If ground-truth probabilities are provided as conditioning, $G_i$ utilizes them. We use this observation for self-improvement and achieve a gain of $0.6\%$ when using $3$ self-loops with HRNet-W18s backbone as shown in~\cref{table:self_improvement}.}
\label{fig:self_improvement}
\end{figure}

%%%%----------------------------------------------------------
\vspace*{0.1in}
\noindent
\textbf{Self-Improvement:}~\cref{fig:self_improvement} (top row) shows that $G_i$ attempts to fix segmentation errors in the input probability map $\mathbf{P}_{i-1}$. For this example, networks improves pixel accuracy to $18.9\%$ when conditioned on a random probability map (accuracy: $1.8\%$). Interestingly, when conditioned on the ground-truth, $G_i$ attempts to utilize it. Inspired by this, we evaluate self-refinement as described in Fig.~\ref{fig:schema} (\textit{middle}), where we feed the predicted probabilities $P_i$ of $G_i$ as its own input during inference.~\cref{table:self_improvement} shows non-negligible improvement in performance using self-refinement.

\begin{table}[b]
\centering
\captionsetup{font=small}
\small
    \renewcommand{\arraystretch}{1.2}
    \rowcolors{1}{}{lightgray}
    \setlength{\tabcolsep}{3pt}
    \begin{tabular}{@{}l| c c | c c c c @{}}
        \Xhline{3\arrayrulewidth}
         \footnotesize{\textbf{Arch}}   & \footnotesize{$\mathbf{G}_0$} & \footnotesize{\textbf{SIM-ENS}} & \multicolumn{4}{c} {\footnotesize{\textbf{SEQ-ENS: Self-Improvement}}}\\        
           &  &  & $T=0$  & $T=1$ & $T=2$ & $T=3$\\
        \hline
        \textbf{H-18s}  &   76.2    & 77.1  & 78.9    & 79.3  & 79.5 & \textbf{79.5}   \\
        \textbf{H-18}   &   78.7    & 79.0  & 79.8    & 80.2  & 80.4 & \textbf{80.6}  \\
        \textbf{H-48}   &   80.5    & 80.6  & 81.3    & 81.6  & 81.4 & \textbf{81.7}  \\
        \Xhline{3\arrayrulewidth}
    \end{tabular}
\caption{Self-Improvement results on Cityscapes \texttt{val} set using single-scale inference. $T$ denotes the number of self-loops using $G_1$. $T=0$ represents the default sequential ensemble. Self-looping during inference provides non-negligible gains.}
\label{table:self_improvement}
\end{table}

%%%%----------------------------------------------------------
\vspace*{0.1in}
\noindent
\textbf{Ensemble of Chains:} Our method allows creation of general graphs where a \textit{node} is a segmentation network and the \textit{edges} define the conditional dependence using the predicted probabilities. We showed that adding models/nodes in a sequential manner in a chain reduces dataset error. We can further reduce the error by adding multiple \textit{chains} and averaging their predictions (Fig.\ref{fig:schema} (\textit{right})). Tab.~\ref{table:ensemble_of_chains} shows results on ensembling $5$ chains with varying length using MobileNetv2-D8~\cite{sandler2018mobilenetv2} backbone on the Cityscapes dataset.

\setlength{\belowcaptionskip}{-15pt}
\begin{table}
\captionsetup{font=small}
    \centering
    \small
    \renewcommand{\arraystretch}{1.2}
    \rowcolors{3}{}{lightgray}
    \setlength{\tabcolsep}{3pt}
    \begin{tabular}{@{}l c  c  c c c@{}}
        \Xhline{3\arrayrulewidth}
        & \multicolumn{5}{c}{\textbf{Number of Chains (C)}} \\
            \cmidrule(r){2-6}
           & $C=1$   & $C=2$   & $C=3$ & $C=4$ & $C=5$  \\
        \hline
        $N = 1$     &   61.3    &   62.2    &   62.4    &   62.7    &   62.7  \\
        $N = 2$     &   64.2    &   65.7    &   66.3    &   65.8    &   66.5   \\
        $N = 3$     &   66.1    &   67.2    &   67.4    &   67.8    &   \textbf{68.1}    \\
        \Xhline{3\arrayrulewidth}
    \end{tabular}
    \caption{Multiple sequential-ensemble chains can be used to improve performance. We show performance by varying the number of chains ($C$) and the number of model generations ($N$) using MobileNetv2-D8~\cite{sandler2018mobilenetv2} backbone on the Cityscapes \texttt{val} set.}
    \label{table:ensemble_of_chains}
\end{table}

% %%%%----------------------------------------------------------
% \textbf{Capacity/Trade off between inference time and accuracy.}
% dynamic offloading

%%%%----------------------------------------------------------
% \textbf{Monotonic Gains - Case Analysis}

%%%%----------------------------------------------------------
\vspace*{0.1in}
\noindent
\textbf{Limitations and Future Work:} Sequential ensembling trains one generation at a time. This improves accuracy compared to simple ensembles, but at the cost of increased training time. Though, empirically, we observe quicker convergence for later generation when training. This is a one-time, offline cost, with potential ways to speed-up, \eg several generations could be trained in parallel using latest prior generation predictions, possibly further reducing training time. Additional improvements could come by injecting feature maps from prior generations as inputs to future generations along with the probability map. This can further decrease the numbers of parameters required in the sequential ensemble to improve performance. 

An open question for future work is whether large models give superior performance in comparison to ensembles with same number of parameters. \cite{wu2016high} shows that ResNet-101 outperforms ResNet-152 for semantic segmentation, indicating that na\"{i}vely increasing parameters may lead to overfitting. Our sequential strategy of building complexity gradually allows training using fewer resources and enables us to modulate the accuracy vs complexity tradeoff at test time.

\label{sec:experiments_analysis}

%%%------------------------------------------------------------------------
%%%%%%%%% CONCLUSION
\vspace*{-0.1in}
\section{Conclusion}
\label{sec:conclusion}

% Existing semantic segmentation approaches focus on training a single model and often employ inference-time multi-scale data augmentation to boost performance. In this paper, we explore deep ensembles to improve the performance of segmentation models. Inspired from boosting, we propose \textit{sequential ensembling} where each generation is trained conditioned on the predicted probabilities of the previous generation. We introduced a novel ADON block to utilize previous generation probabilities as the conditioning input. The ADON block spatially modulates the intermediate features via affine transformation conditioned on previous generation probabilities. 

\noindent
In this work, we explore deep ensembles to improve the performance of segmentation models. We benchmark ensembles of state-of-the-art deep models for multiple datasets. Inspired by boosting approaches, we propose \textit{sequential ensembling} -- a strategy to gradually increase model complexity to improve performance. This an alternative to the standard practice of training large models and compressing them for on-device deployment, and is suitable for problems with dynamic resource constraints. Our proposed ADON block utilizes feature modulation to efficiently connect multiple generations in the sequential ensemble. Sequential ensembles demonstrate state-of-the-art results on challenging datasets. We hope that our work will inspire future research in ensembling for semantic segmentation as well as other dense prediction tasks such as depth and pose estimation. 

\label{sec:experiments_qualitative}

% ------------------------------------------
\begin{figure*}
\centering
\includegraphics[width=1\linewidth, height=0.4\linewidth]{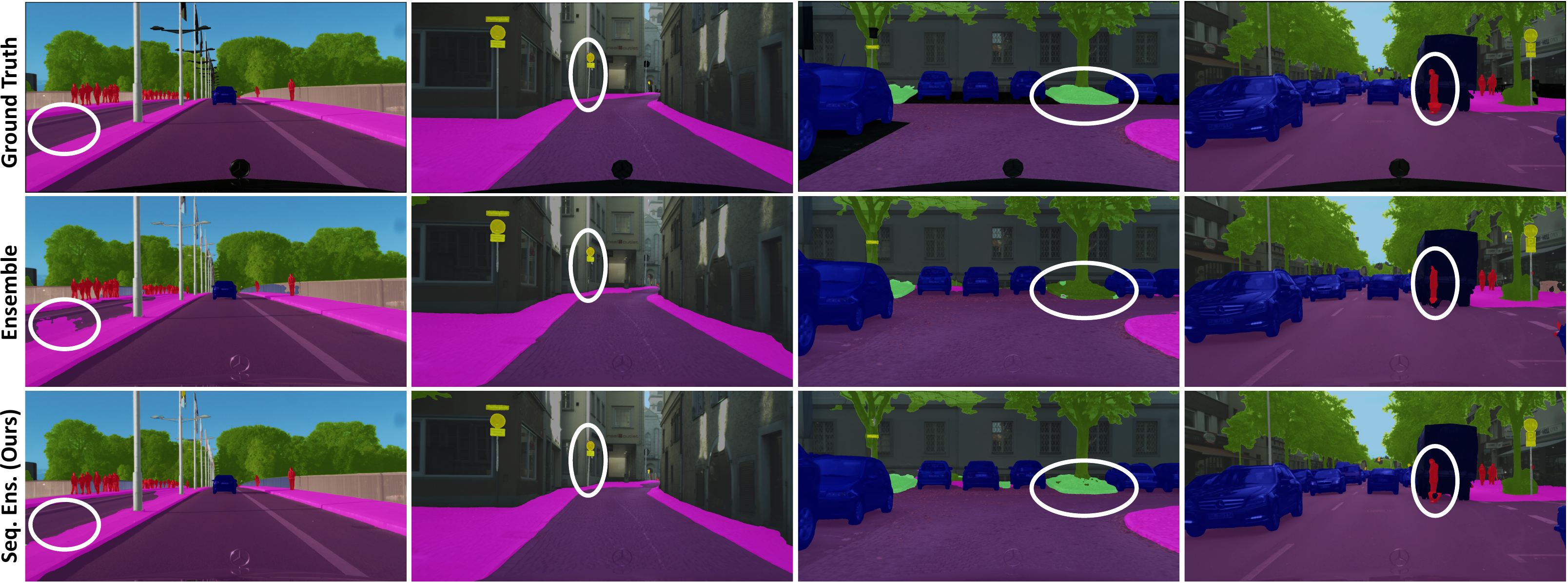} \\
\vspace{0.3cm}
\includegraphics[width=1\linewidth, height=0.4\linewidth]{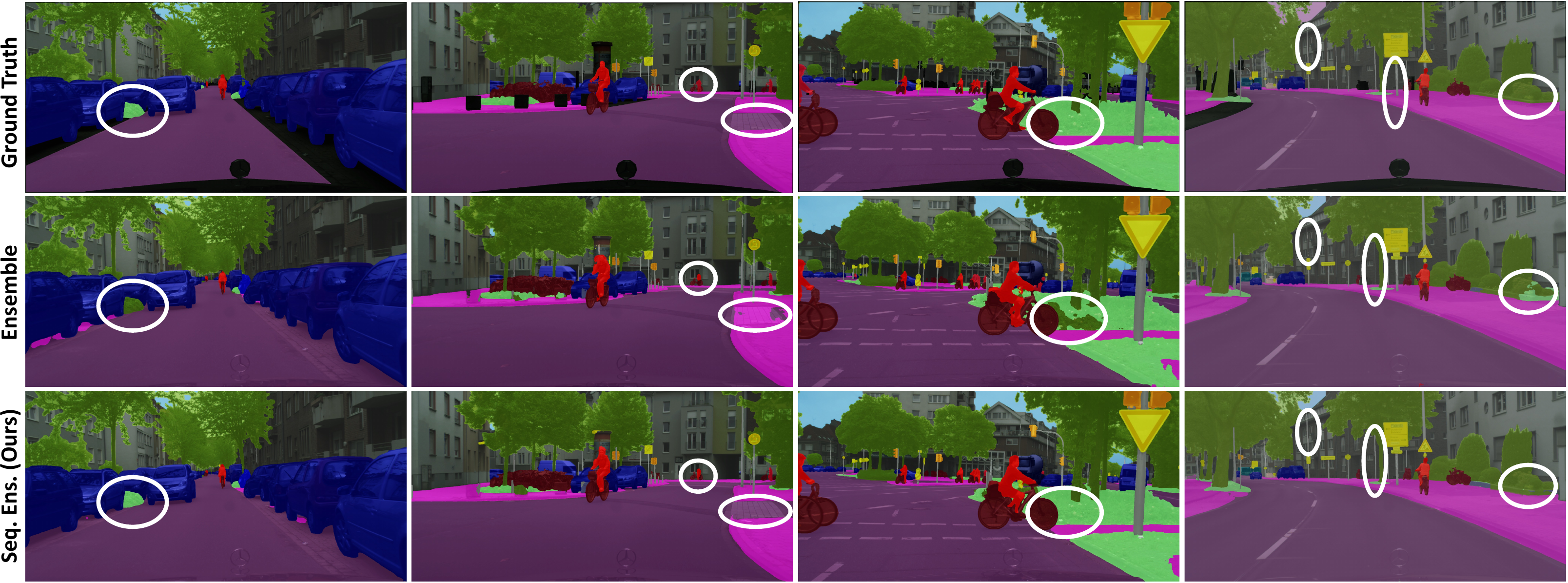}\\
\vspace{0.3cm}
\includegraphics[width=1\linewidth, height=0.4\linewidth]{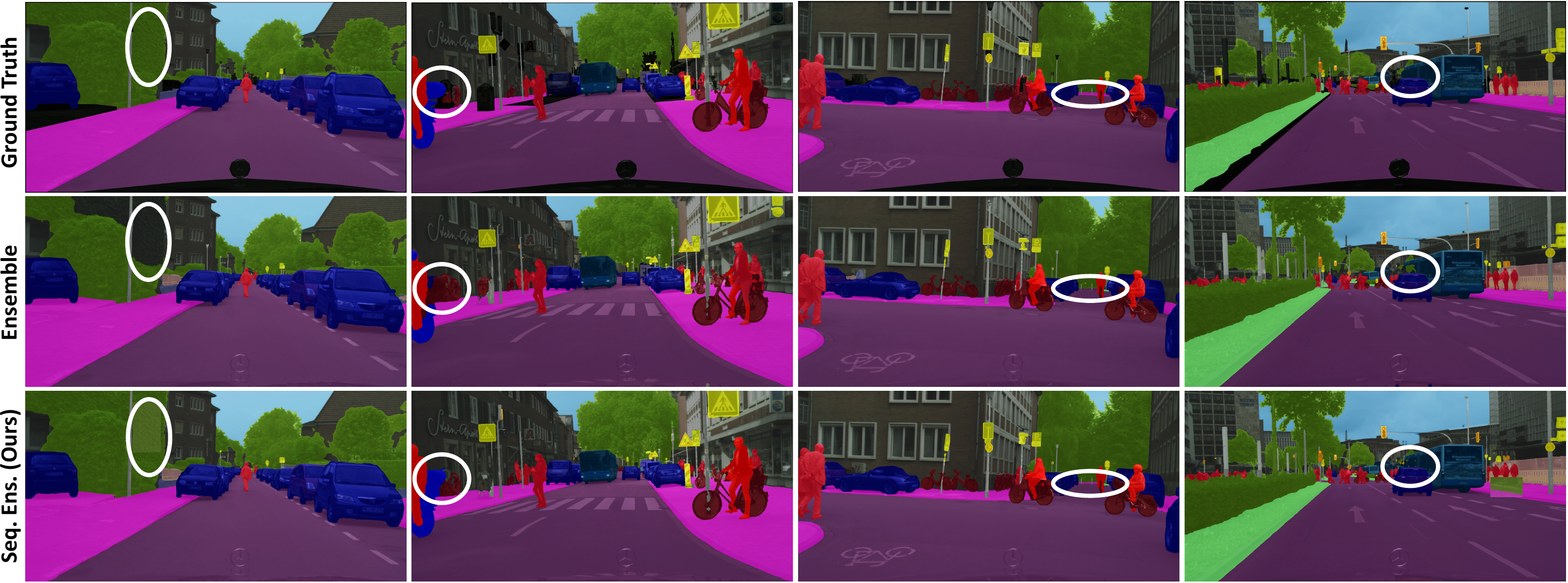} 
\caption{Qualitative comparison of Sequential Ensembles (SEQ-ENS) with the Simple Ensembles (SIM-ENS) on Cityscapes \texttt{val} set using HRNet-W48 backbone (N=2). The white eclipses highlight the fine-grained details that our approach captures in comparison to the baseline. Zoom in for details.}
\label{fig:supplementary:qual_cityscapes}
\end{figure*}

\newpage

\begin{figure*}
\centering
\includegraphics[width=1\linewidth, height=0.4\linewidth]{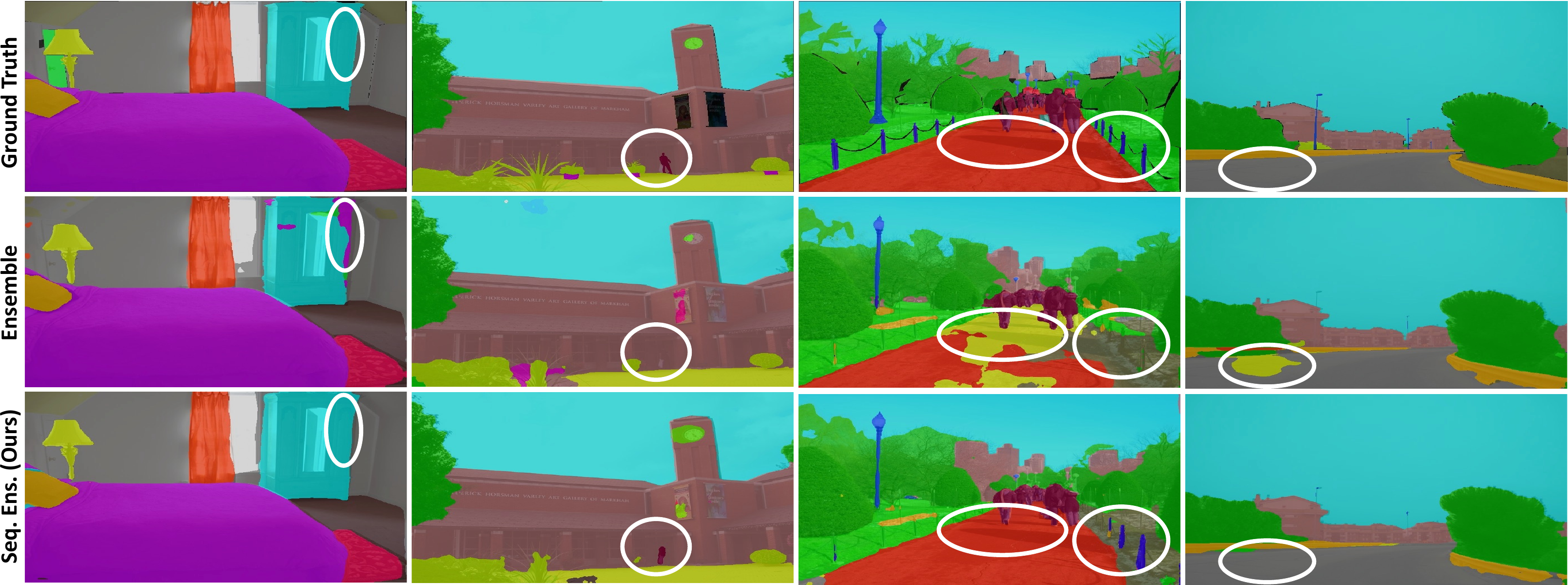} \\
\vspace{0.3cm}
\includegraphics[width=1\linewidth, height=0.4\linewidth]{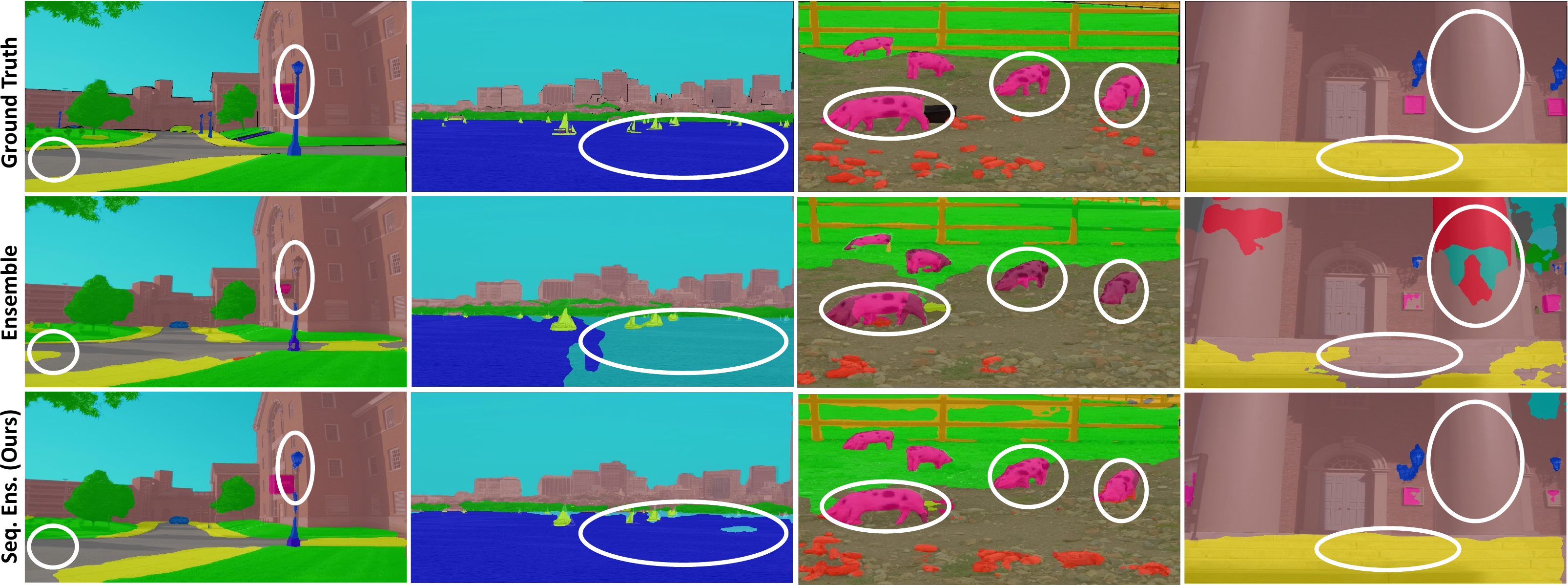}\\
\vspace{0.3cm}
\includegraphics[width=1\linewidth, height=0.4\linewidth]{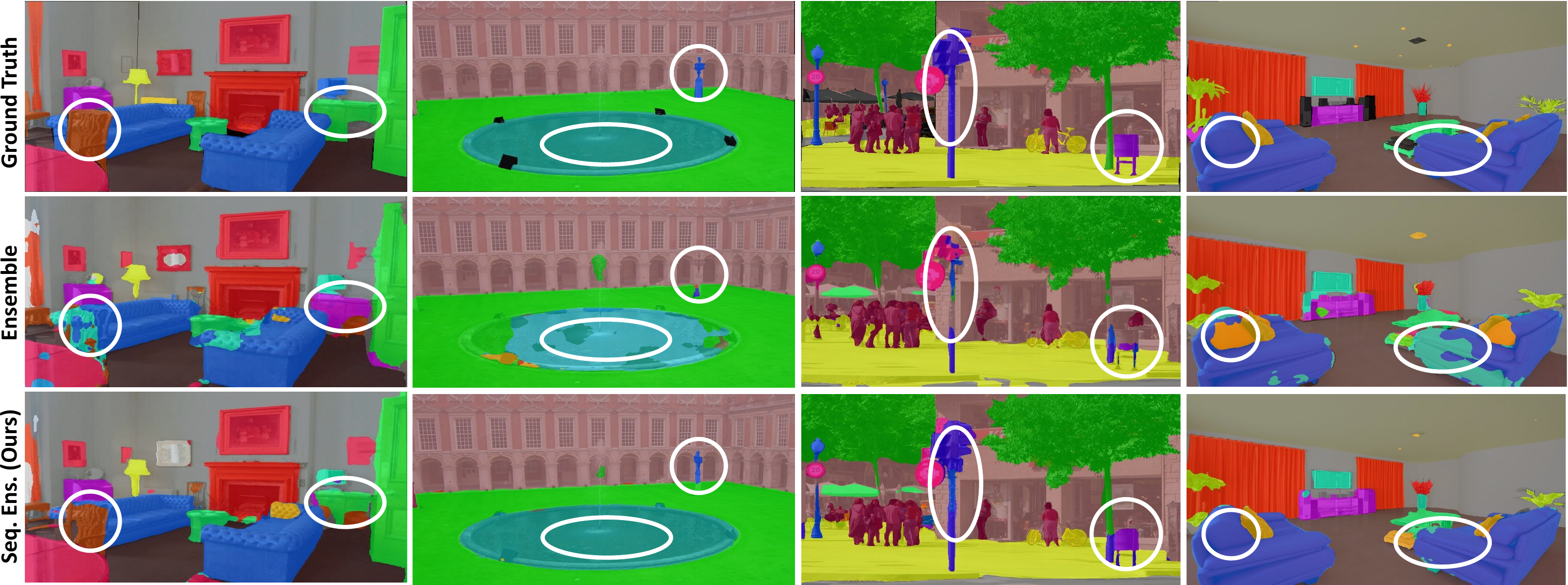}
\caption{Qualitative comparison of Sequential Ensembles (SEQ-ENS) with the Simple Ensembles (SIM-ENS) on ADE20K \texttt{val} set using HRNet-W48 backbone (N=2). SEQ-ENS is especially effective on a complex dataset like ADE-20K with $150$ categories. We set a new state-of-art on ADE20K \texttt{val} set. Zoom in for details.}
\label{fig:supplementary:qual_ade20k}
\end{figure*}

\clearpage

%%%----------------------------------------------------------------------
%%%%%%%%% REFERENCES

{\small
\bibliographystyle{ieee_fullname}
\bibliography{references}
}

\end{document}

%%%%%%%%% TITLE
\title{Sequential Ensembling for Semantic Segmentation}

\author{{Rawal Khirodkar}{\tt\small rkhirodk@cs.cmu.edu}}
\maketitle

%%%------------------------------------------------------------------------
\section{Adaptive Modulation (ADON) Block Code}
In this section, we describe the code of ADON Block in \texttt{PyTorch}. The code in Listing.~\ref{code:adon} outlines the details of functions $\mathbf{F}_{\text{shared}}$, $\mathbf{F}_{\text{scale}}$ and $\mathbf{F}_{\text{bias}}$. ADON blocks can be incorporated in any existing feature extraction backbone.

\begin{lstlisting}[language=Python, caption=Code for ADON block., label=code:adon]
class ADON(nn.Module):
  
    def __init__(self, num_classes, num_channels, hidden_channels=128):
        """
            ADON Block for Sequential Ensembles
            num_classes: number of classes in the dataset
            num_channels: number of channels in the intermediate feature X
            hidden_channels: K, dimensionality of the latent space of the ADON block
        """
        super(ADON, self).__init__()
        
        self.F_shared = nn.Sequential(
                nn.Conv2d(num_classes, hidden_channels, kernel_size=3, padding=1),
                nn.ReLU()
            )
        self.F_scale = nn.Conv2d(hidden_channels, num_channels, kernel_size=3, padding=1)
        self.F_bias = nn.Conv2d(hidden_channels, num_channels, kernel_size=3, padding=1)
        return

    def forward(self, X, pred_prob):
        """
            X: intemediate feature of the segmentation backbone
            pred_prob: class probabilities from the previous generation
        """
        pred_prob = F.interpolate(pred_prob, size=X.size()[2:], mode='bilinear', align_corners=False)
        
        lambda_ = self.F_shared(pred_prob)
        scale = self.F_scale(lambda_)
        bias = self.F_bias(lambda_)
        
        X_prime = X * scale + bias
        
        return X_prime
\end{lstlisting}

%%%------------------------------------------------------------------------
\section{Insertion of ADON Blocks}
\begin{figure*}
\centering
\includegraphics[width=\linewidth]{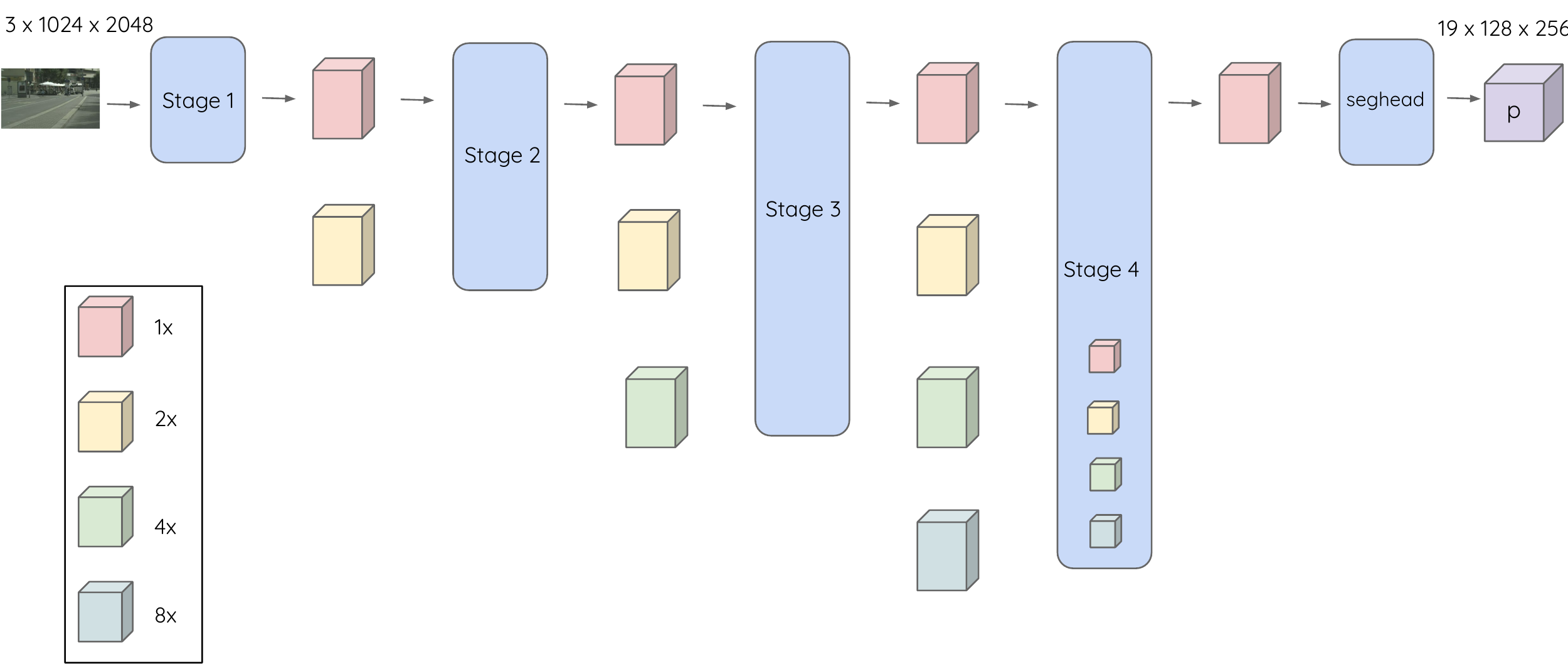}
\caption{Illustration of HRNet-W48 backbone for Cityscapes dataset at input resolution $1024 \times 2048$. The blue blocks depict the four stages in the HRNet architecture.}
\label{fig:supplementary:hrnet}
\end{figure*}

\begin{figure*}
\centering
\includegraphics[width=\linewidth]{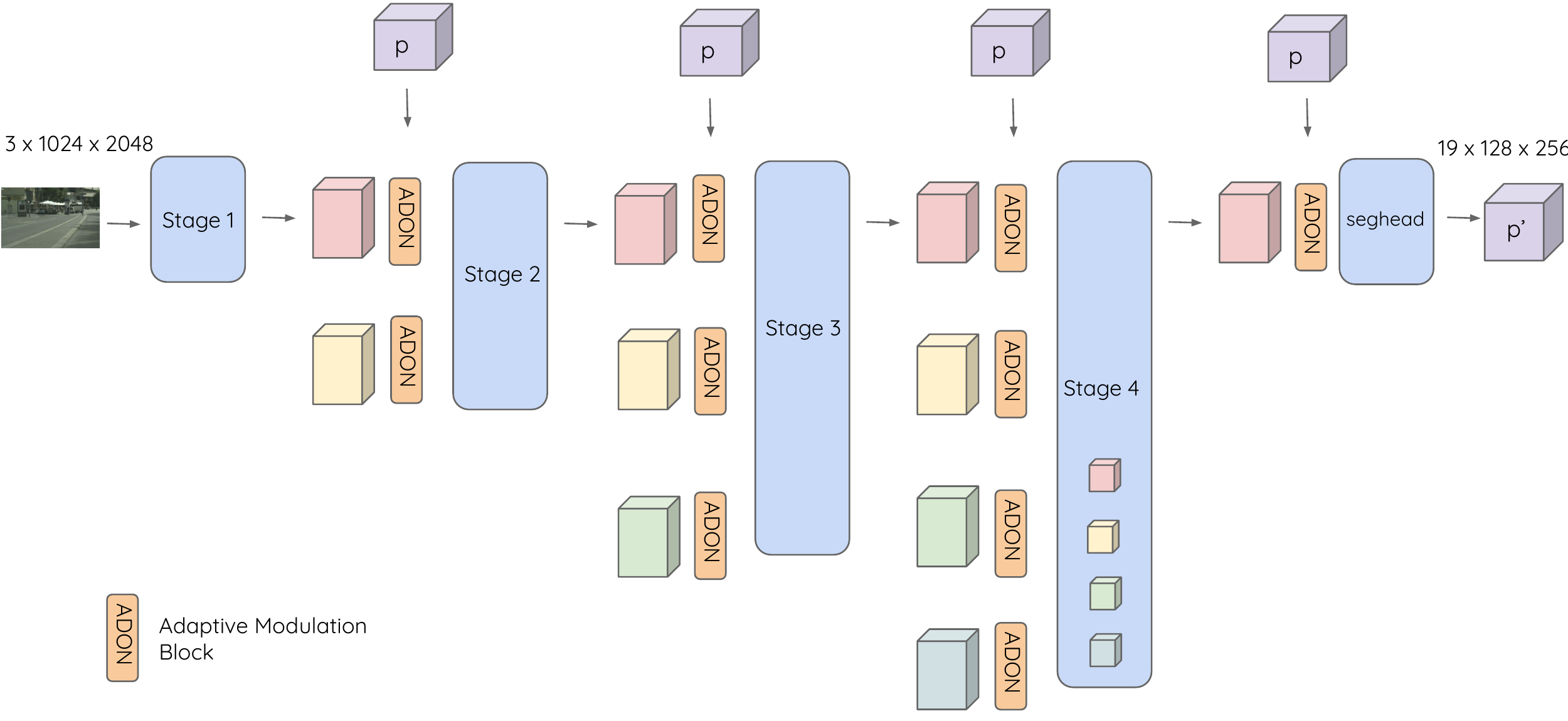}
\caption{Illustration of HRNet-W48 backbone with ADON blocks for Cityscapes dataset at input resolution $1024 \times 2048$. We insert 10 ADON blocks in the baseline architecture.}
\label{fig:supplementary:hrnet_adon}
\end{figure*}

\begin{figure*}
\centering
\includegraphics[width=\linewidth]{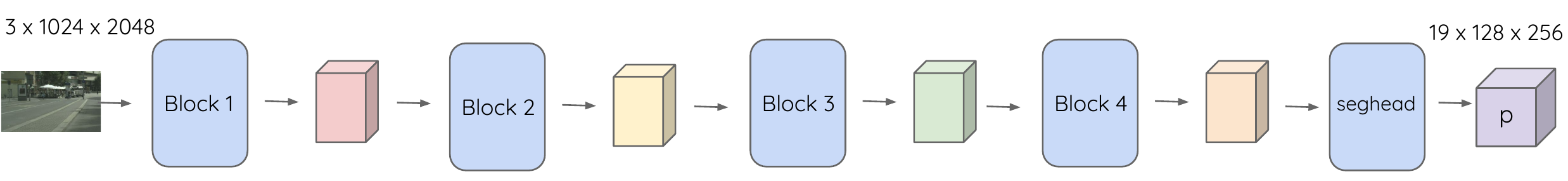}
\caption{Illustration of ResNet101 backbone for Cityscapes dataset at input resolution $1024 \times 2048$. The blue blocks depict the ResNet blocks in the architecture.}
\label{fig:supplementary:resnet}
\end{figure*}

\begin{figure*}
\centering
\includegraphics[width=\linewidth]{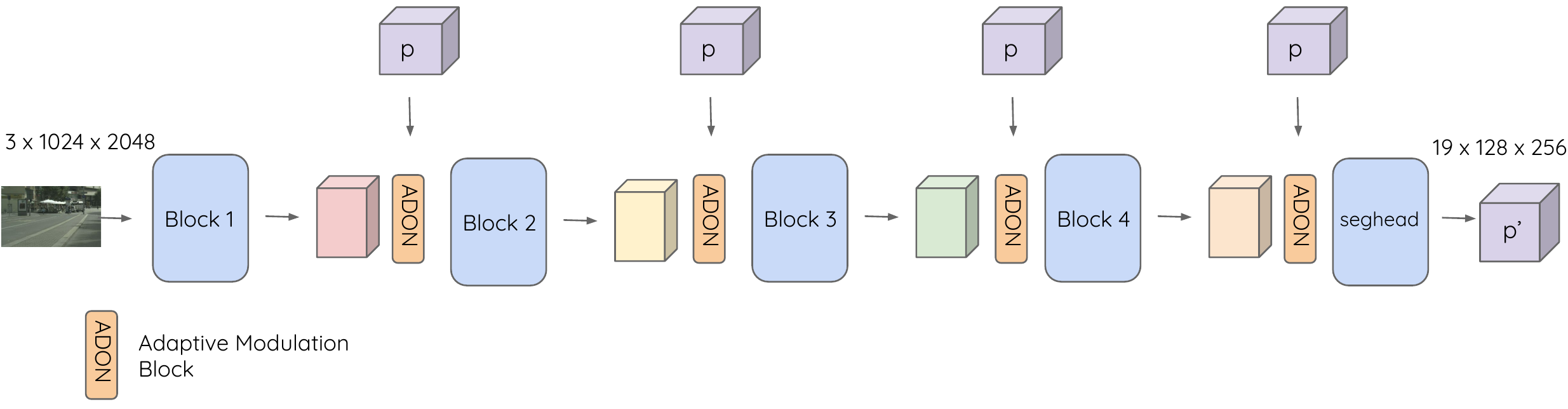}
\caption{Illustration of ResNet101 backbone with ADON blocks for Cityscapes dataset at input resolution $1024 \times 2048$. We insert 4 ADON blocks in the baseline architecture.}
\label{fig:supplementary:resnet_adon}
\end{figure*}

\begin{figure*}
\centering
\includegraphics[width=\linewidth]{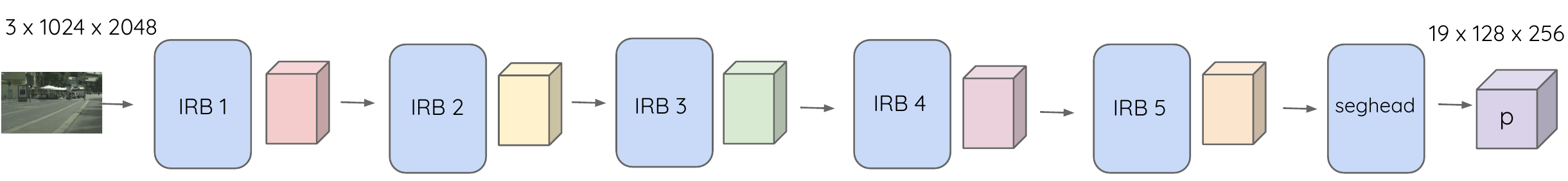}
\caption{Illustration of MobileNetv3-Small backbone for Cityscapes dataset at input resolution $1024 \times 2048$. The blue blocks depict the Inverted Residual Blocks (IRB) in the architecture.}
\label{fig:supplementary:mobilenetv3}
\end{figure*}

\begin{figure*}
\centering
\includegraphics[width=\linewidth]{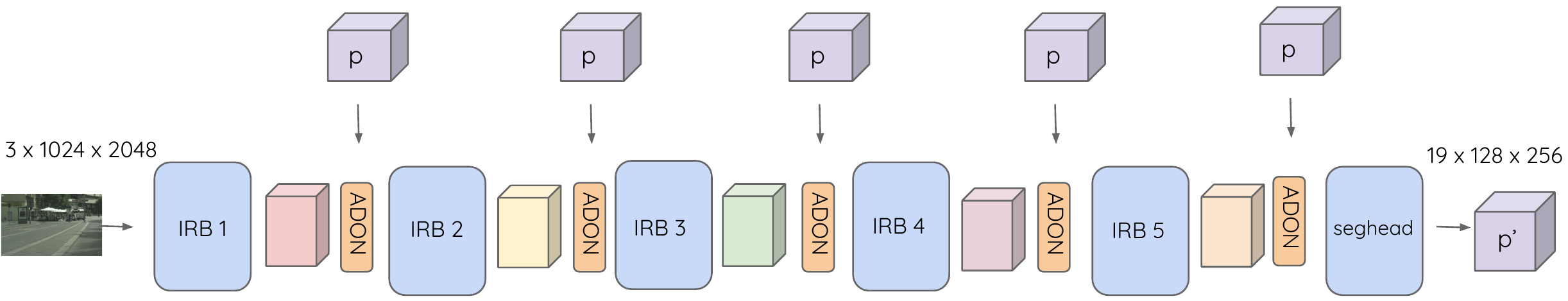}
\caption{Illustration of MobileNetv3-Small backbone with ADON blocks for Cityscapes dataset at input resolution $1024 \times 2048$. We insert 5 ADON blocks in the baseline architecture.}
\label{fig:supplementary:mobilenetv3_adon}
\end{figure*}

We show the insertion of multiple ADON blocks in HRNet~\cite{sun2019high}, ResNet~\cite{he2016deep} and MobileNet-V3~\cite{howard2019searching} architectures. We depict HRNet as $G_0$ in \cref{fig:supplementary:hrnet} and $G_1$ in \cref{fig:supplementary:hrnet_adon}. Similarly, ResNet as $G_0$ in \cref{fig:supplementary:resnet}, $G_1$ in \cref{fig:supplementary:resnet_adon} and MobileNet-V3 as $G_0$ in \cref{fig:supplementary:mobilenetv3}, $G_1$ in \cref{fig:supplementary:mobilenetv3_adon}.

% ----------hrnet--------------

\clearpage
\newpage

% ----------resnet--------------

% ----------mobilenetv3--------------

%%%------------------------------------------------------------------------
\newpage
\clearpage
\section{Comparison to other Ensembling Baselines}
We explore various ensembling strategies for semantic-segmentation and compare them with sequential ensembles. In addition to ensembling baselines such as \textit{Voting} and \textit{W-Avg} (Tab.~1, main paper) which achieved similar performance as SIM-ENS, we compare SEQ-ENS with two additional baselines in \cref{table:supplementary:baseline}.

\vspace*{0.2in}
Specifically, we train a separate weight predictor network based on ResNet-50 backbone to predict combination weights $w_i$ from the input image used for ensembling. The weight predictor is trained after independently training all models in the ensemble. We consider two cases when (a) $w_i \in \mathbb{R}$ and (b) $w_i \in \mathbb{R}^{H \times W}$. As shown in the table, we observe that SEQ-ENS with $N=2$ outperforms all ensembling baselines on all datasets. For example, SEQ-ENS with $N=2$ outperforms W-Avg ENS $w_i \in \mathbb{R}^{H \times W}$ with $N=2$ by $2.1$ on the challenging ADE20K dataset. Due to GPU memory limitations, we are not able to report numbers using W-Avg ENS $w_i \in \mathbb{R}^{H \times W}$ with $N=15$.

\vspace*{0.4in}
\begin{table}
% \captionsetup{font=scriptsize}
\begin{center}
\vspace*{-0.2in}
\resizebox{4.6in}{!}{
    \small
    \setlength{\tabcolsep}{8pt}
    \renewcommand{\arraystretch}{1.5}
    \begin{tabular}{@{}l|l|l|l @{}}
    \toprule[1.0pt]
         \small{\textbf{Method}} &\small{\textbf{Cityscapes}}   & \small{\textbf{ADE20K}} & \small{\textbf{COCO-Stuff}}\\
    \hline
    \small{Single Model} & 76.2  &  33.1 & 32.4 \\
    \hline
    \small{SIM-ENS} \scriptsize{($N=2$)} &  77.1 (+0.9) & 34.5 (+1.4) & 33.5 (+1.1)\\
    \small{SIM-ENS} \scriptsize{($N=5$)} &  77.5 (+1.3) & 34.9 (+1.8) & 33.8 (+1.4) \\
    \small{SIM-ENS} \scriptsize{($N=15$)} &  77.8 (+1.6) & 35.8 (+2.7) & 34.9 (+2.5) \\
    \hline
    \small{W-Avg ENS}, $w_i \in \mathbb{R}$ \scriptsize{($N=2$)} &  77.4 (+1.2) & 34.9 (+1.8) & 32.9 (+0.5) \\
    \small{W-Avg ENS}, $w_i \in \mathbb{R}^{H \times W}$ \scriptsize{($N=2$)} &  77.8 (+1.6)  & 35.5 (+2.4) & 34.1 (+1.7) \\
    \hline
    \small{Voting Ens.} \scriptsize{($N=15$)} &  76.9 (+0.7) & 35.3 (+2.2) & 35.1 (+2.7) \\
    \small{W-Avg ENS}, $w_i \in \mathbb{R}$ \scriptsize{($N=15$)} &  78.0 (+1.8) & 35.0 (+1.9) & 34.6 (+2.2) \\
    
    \hline
    \small{SEQ-ENS} \scriptsize{($N=2$)} & \textbf{78.9 (+2.7)} & \textbf{37.6 (+4.5)} & \textbf{36.3 (+3.9)} \\
    \bottomrule[1.0pt]
    \end{tabular}
}
\vspace*{0.1in}
\caption{Comparison of SEQ-ENS with other ensembling baselines for $N=2$ on various datasets using HRNetW18s-v2 backbone.}
\label{table:supplementary:baseline}
\end{center}
\end{table}

%%%%----------------------------------------------------------
% \noindent
\section{ADON Block Architecture:} We vary the dimension $K$ of the latent space of the ADON block used for modulation on the Cityscapes dataset using various backbones as shown in Tab.~\ref{table:adon_architecture}. We observe that increasing $K$ improves performance across backbones, with $K=128$ achieving the optimal balance between the parameter overhead and segmentation performance.

\begin{table}[h]
\captionsetup{font=small}
\centering
\small
    \renewcommand{\arraystretch}{1.2}
    \rowcolors{1}{}{lightgray}
    \setlength{\tabcolsep}{3pt}
    \begin{tabular}{@{}l|c c c c c c @{}}
        \Xhline{3\arrayrulewidth}
        \textbf{Arch} & $G_0$  & $K=16$  & $K=32$  & $K=64$ & $K=128$ & $K=256$\\
        \hline
        \textbf{H-18s} & 76.2  &   76.3    & 76.9  & 77.6    & \textbf{78.9}  & 78.7   \\
        \textbf{H-18}  & 78.7  &   78.7    & 79.0  & 79.2    & \textbf{79.8}  & 79.7  \\
        \textbf{H-48}  & 80.5  &   80.6    & 80.7  & 81.2    & 81.3  & \textbf{81.4}  \\
        \Xhline{3\arrayrulewidth}
    \end{tabular}
\caption{Variation of ADON block's latent space dimension $K$ on the Cityscapes \texttt{val} set using single-scale inference.}
\label{table:adon_architecture}
\end{table}

%%%------------------------------------------------------------------------
\newpage
\clearpage
\section{Simple Ensembles}
We implemented Simple-Ensembles (SIM-ENS) as an equal weighted average of prediction probabilities from each model. Each model is independently trained with a different random seed. We compare two initialization strategies for SIM-ENS, \textit{Random} Vs \textit{ImageNet}. \textit{Random:} All the model parameters in the ensemble are randomly initialized.  \textit{ImageNet:} Only the segmentation head parameters in the ensemble are randomly initialized.  For the model backbone, we use the ImageNet pretrained parameters.

\subsection{Ensemble Segmentation Accuracy}
\begin{figure}[H]
\centering
\includegraphics[width=0.8\linewidth]{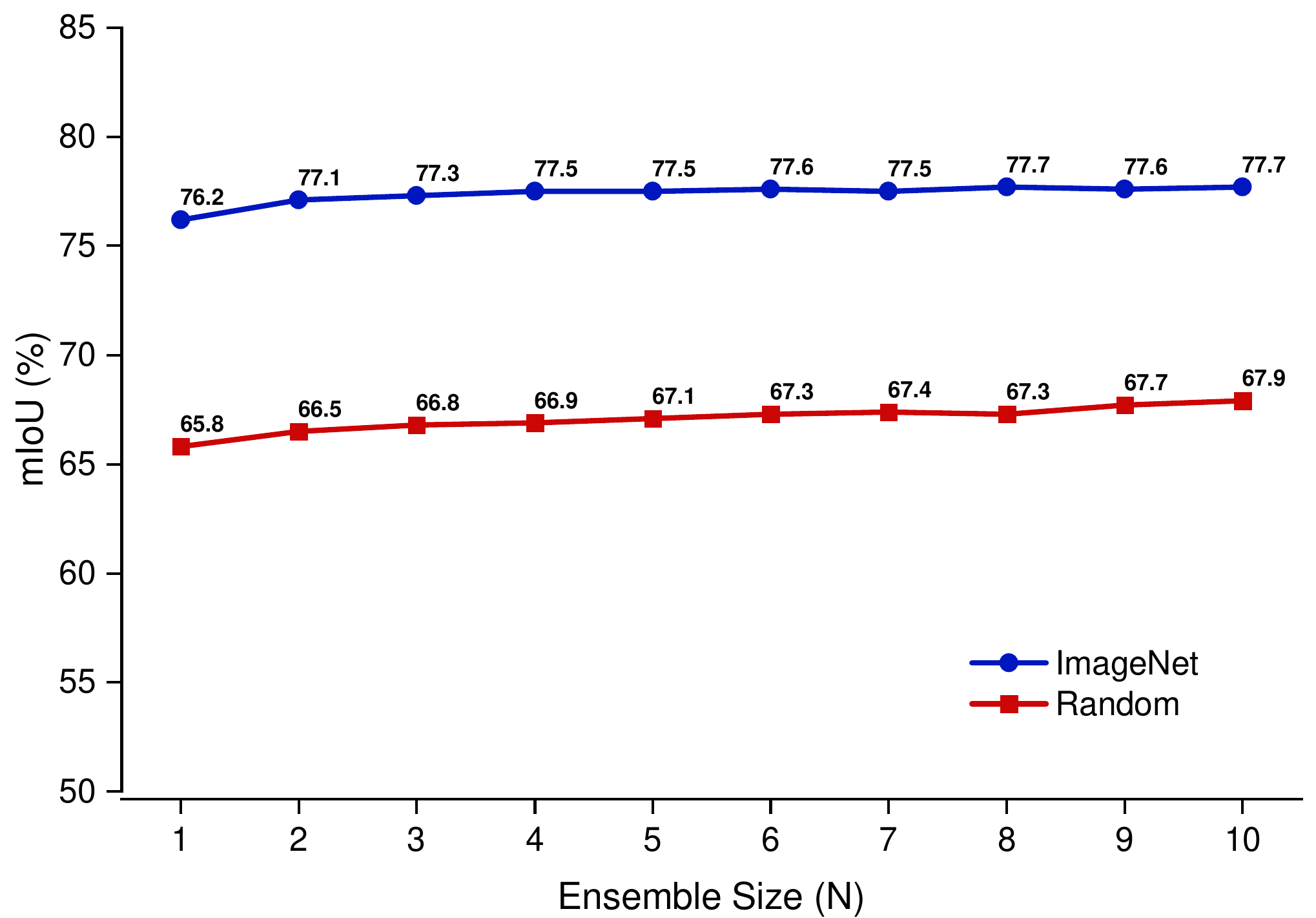}
\caption{Comparison of \textit{Random} Vs \textit{ImageNet} initialization in Simple Ensembles on Cityscapes dataset using HRNet-W18s architecture trained at $512 \times 1024$ image resolution.}
\label{fig:supplementary:random_vs_imagenet}
\end{figure}

We show that simple-ensembles with ImageNet pretraining significantly outperform randomly initialized simple-ensembles in \cref{fig:supplementary:random_vs_imagenet}. This finding is consistent with the observations in modern transfer learning~\cite{he2019rethinking, pan2009survey}. Further, with the increasing ensemble size ($N$) we observe slightly better gains for \textit{Random} in comparison to \textit{ImageNet} initialization. 

We use \textit{ImageNet} initialization due to its better performance for both SIM-ENS and SEQ-ENS in all our experiments.

\vspace{2cm}

\newpage

\subsection{Ensemble Diversity} 
\begin{figure}[H]
\centering
\includegraphics[width=1\linewidth]{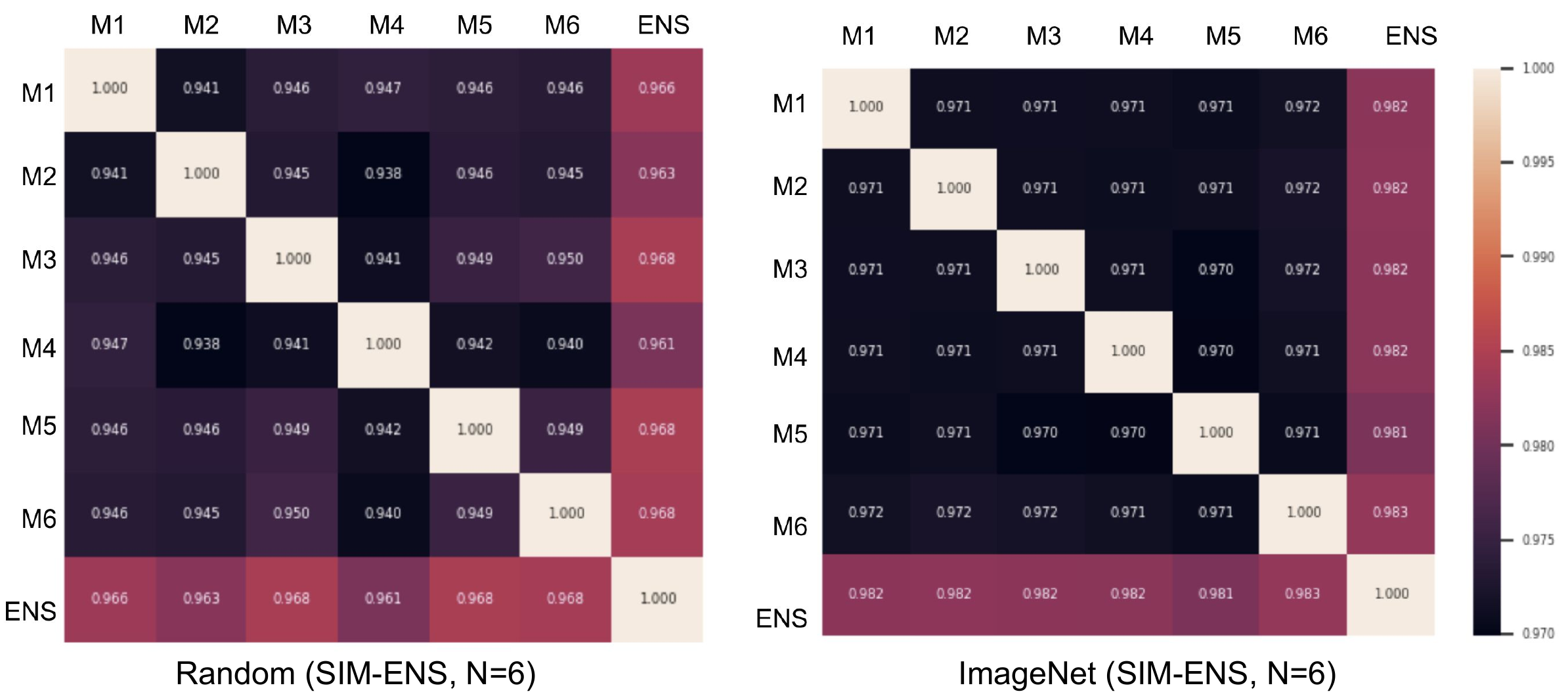}
\caption{Pair-wise cosine similarity (1 implies perfect correlation, 0 implies orthogonality) between segmentation probabilities for \textit{Random} Vs \textit{ImageNet} for simple ensemble (N=6) on Cityscapes. Individual predictions in \textit{Random} SIM-ENS are less correlated than \textit{ImageNet} SIM-ENS.}
\label{fig:supplementary:random_vs_imagenet_prediction_distance}
\end{figure}

\begin{figure}[H]
\centering
\includegraphics[width=1\linewidth]{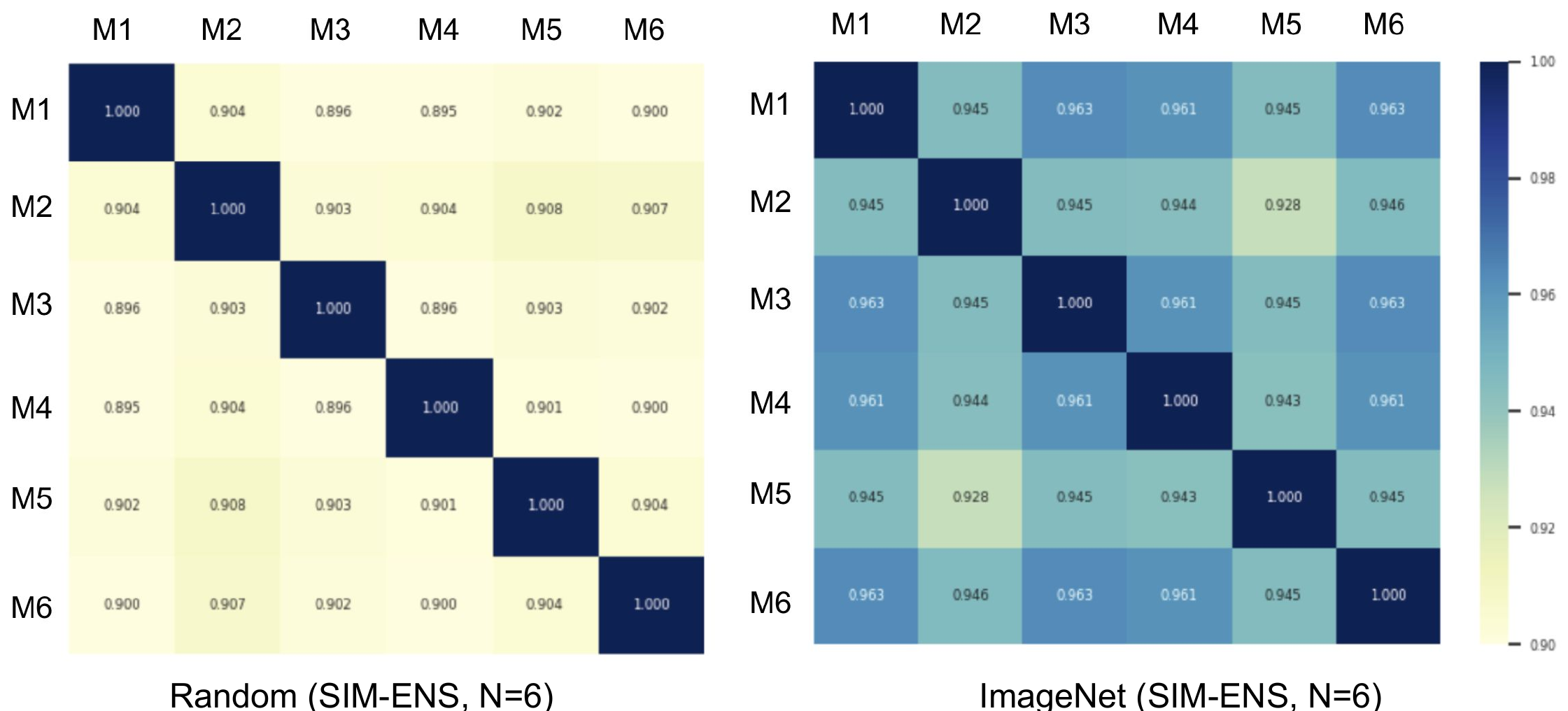}
\caption{Pair-wise cosine similarity (1 implies perfect correlation, 0 implies orthogonality) between model parameters after training for \textit{Random} Vs \textit{ImageNet} for simple ensemble (N=6) on Cityscapes. Model weights in \textit{Random} SIM-ENS are less correlated than \textit{ImageNet} SIM-ENS.}
\label{fig:supplementary:random_vs_imagenet_model_distance}
\end{figure}

We measure the diversity of the ensemble in the prediction space and the model parameter space. We measure pair-wise cosine similarity between segmentation probabilities \cref{fig:supplementary:random_vs_imagenet_prediction_distance} and model weights after training \cref{fig:supplementary:random_vs_imagenet_model_distance}. Each model in the \textit{Random} SIM-ENS is less correlated in both predictions space and parameter space in comparison to models in \textit{ImageNet} SIM-ENS.

\newpage
\subsection{Ensemble Confidence Calibration} 
\begin{figure}[H]
\centering
\includegraphics[width=1\linewidth]{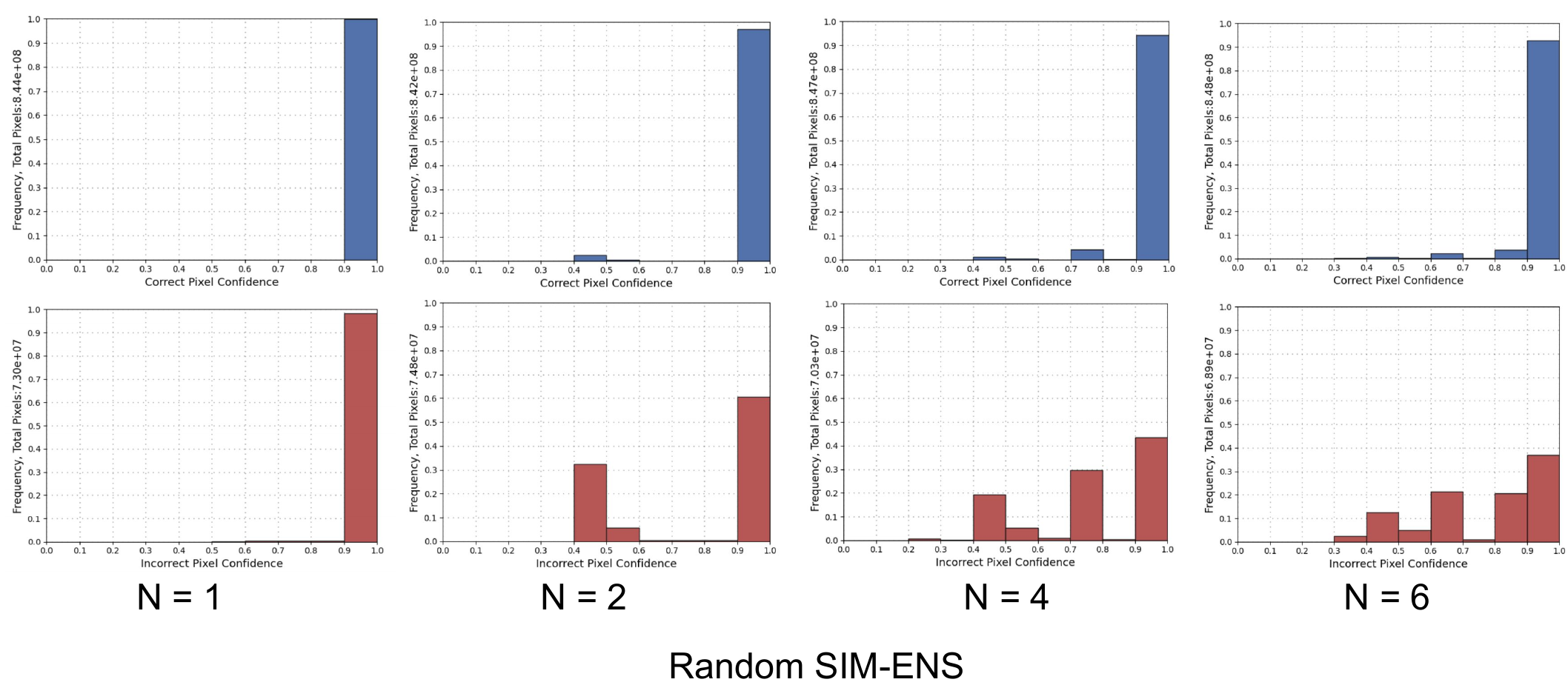}
\caption{Per pixel confidence histogram for correct in \textcolor{blue}{blue} and incorrect pixels \textcolor{red}{red} for \textit{Random} SIM-ENS with an ensemble size of 6 on Cityscapes. We observe improved confidence calibration with increasing ensemble size.}
\label{fig:supplementary:random_calibration}
\end{figure}

\begin{figure}[H]
\centering
\includegraphics[width=1\linewidth]{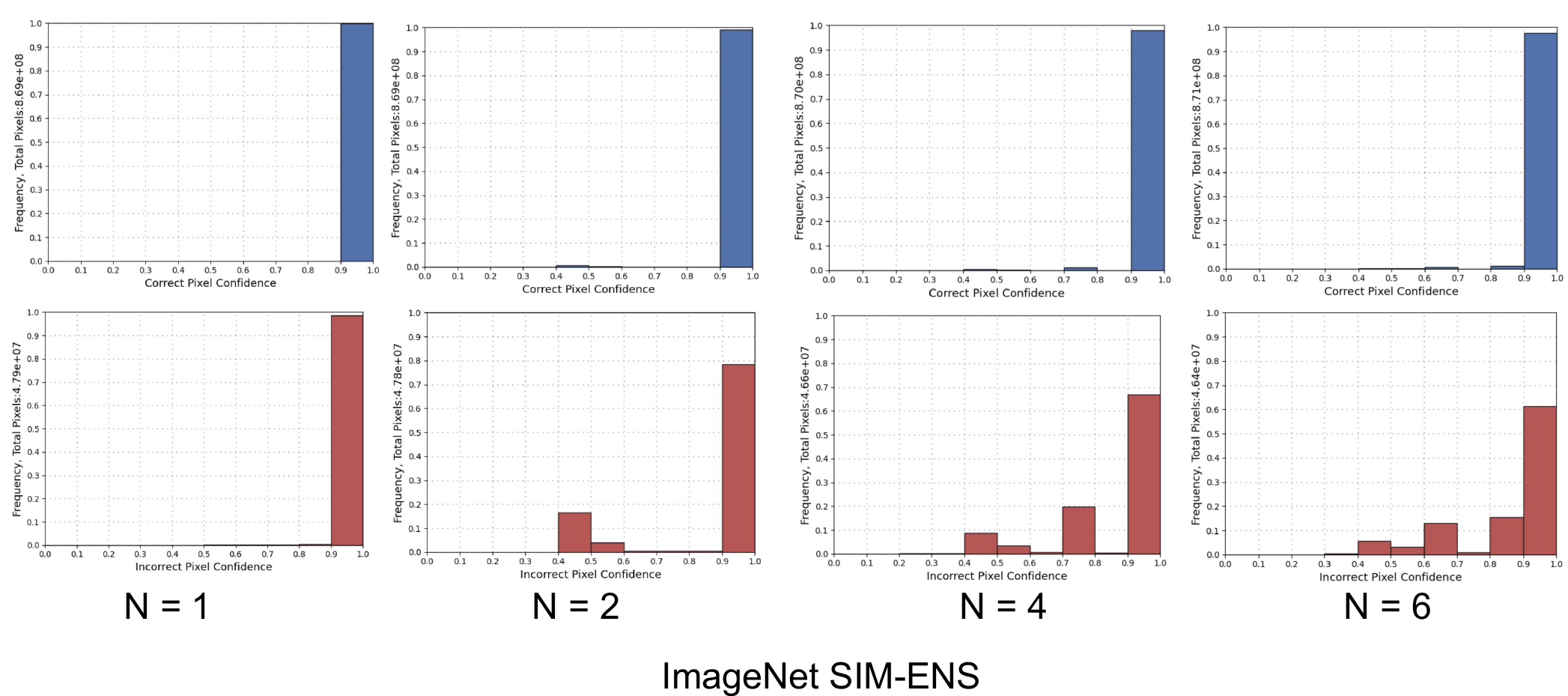}
\caption{Per pixel confidence histogram for correct in \textcolor{blue}{blue} and incorrect pixels \textcolor{red}{red} for \textit{ImageNet} SIM-ENS with an ensemble size of 6 on Cityscapes. }
\label{fig:supplementary:imagenet_calibration}
\end{figure}

In semantic segmentation, we often want probabilistic predictions to reflect
confidence or uncertainty. We empirically evaluate the confidence calibration by plotting the confidence histogram of correct and incorrect pixel predictions for \textit{Random} and \textit{ImageNet} SIM-ENS in \cref{fig:supplementary:random_calibration} and \cref{fig:supplementary:imagenet_calibration}. We show that truely independently trained models in an ensemble (\textit{Random} SIM-ENS) showcase better confidence calibration.

\begin{figure}[H]
\centering
\includegraphics[width=0.65\linewidth]{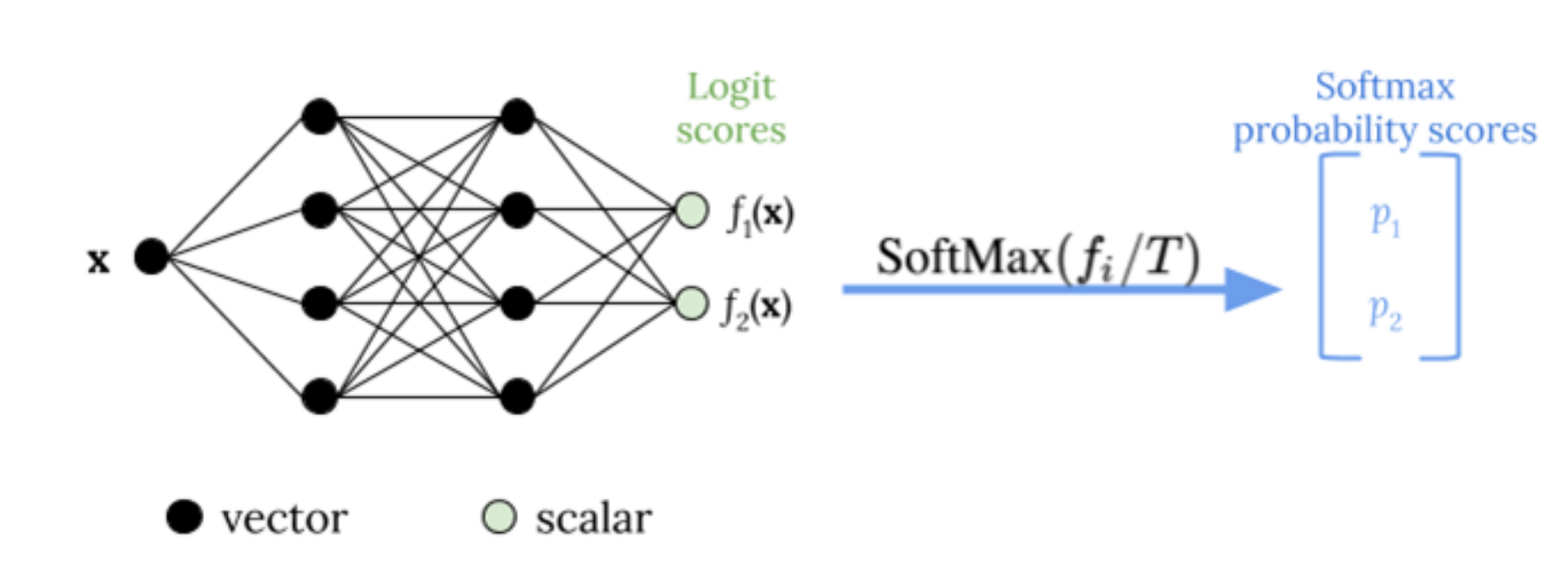}
\caption{The logits from the segmentation network are divided by the temperature $T$ before softmax to calibrate the certainty estimates.}
\label{fig:supplementary:temperature_scaling}
\end{figure}

\newpage
\textbf{Softmax Temperature Scaling.} We further investigate the effect of confidence calibrating individual models in the \textit{ImageNet} SIM-ENS using Softmax Temperature Scaling~\cite{guo2017calibration} as shown in \cref{fig:supplementary:temperature_scaling}. 

\begin{figure}[H]
\centering
\includegraphics[width=0.9\linewidth]{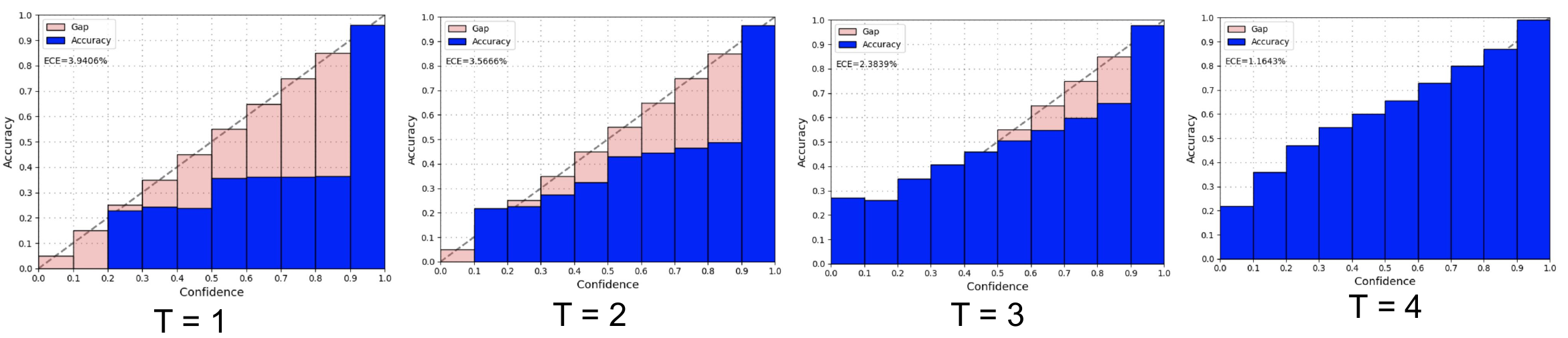}
\caption{We measure Expected Calibration Error (ECE)~\cite{guo2017calibration} by varying temperature $T$ for $N=1$ in \textit{ImageNet} SIM-ENS. We observe best ECE at $T=4$. Note, temperature scaling for calibration does not affect segmentation accuracy.}
\label{fig:supplementary:softmax_temperature}
\end{figure}

We vary the temperature $T$ and ensembling the predictions using the scaled probabilities as weights. \cref{fig:supplementary:softmax_temperature} shows that $T=4$ obtains minimum expected calibration error (ECE). 

\begin{figure}[H]
\centering
\includegraphics[width=0.85\linewidth]{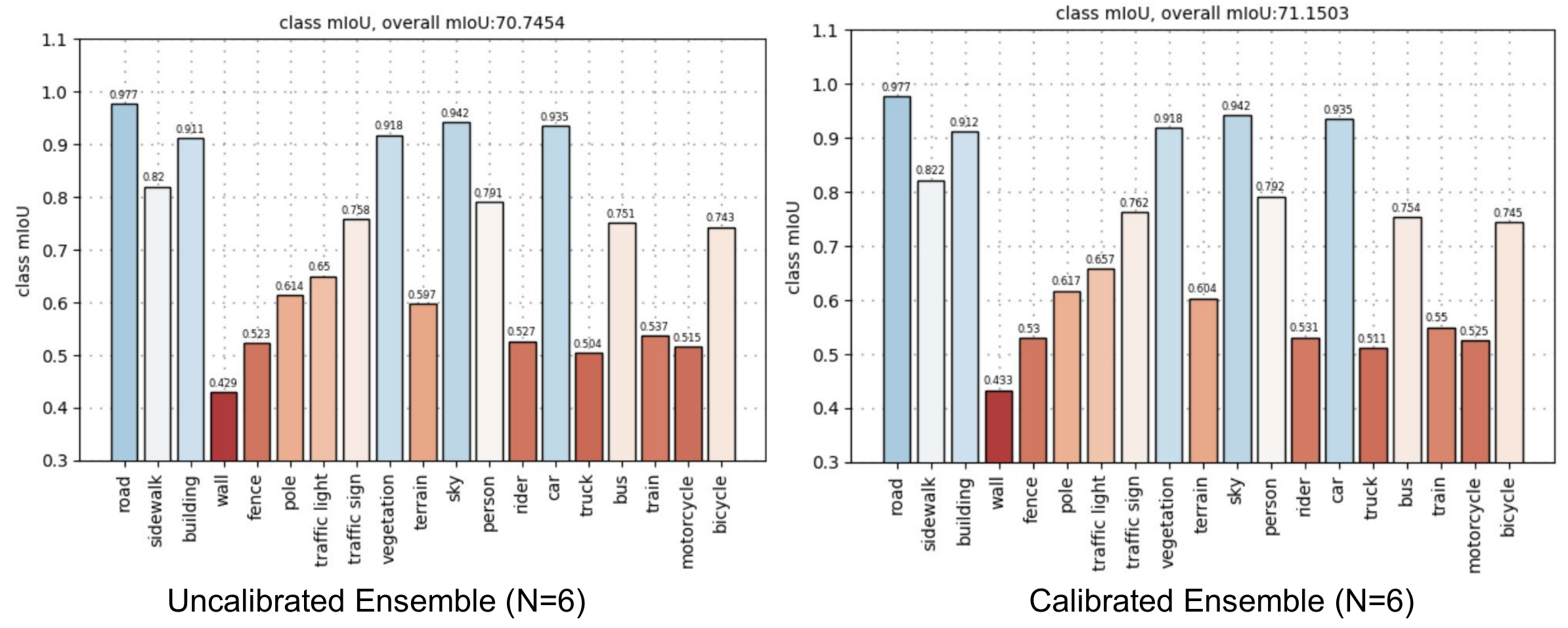}
\caption{Uncalibrated Vs Calibrated \textit{ImageNet} SIM-ENS (N=6) classwise mIoU on Cityscapes \texttt{val} set using HRNet-W18s backbone. Note, each model is trained using only $25$\% of the training data for experiment ease. mIoU at $N=1$ is at 68.92. Each model is calibrated using temperature $T=4$, we observe that ensembling calibrated models performs slightly better than uncalibrated models.}
\label{fig:supplementary:uncalibrated_vs_calibrated}
\end{figure}

Further, we use $T=4$ to create an calibrated ensemble which outperforms uncalibrated ensemble by $0.4$ mIoU on Cityscapes as shown in \cref{fig:supplementary:uncalibrated_vs_calibrated}.

% \textbf{Learning Confidence Explicitly:} Following \cite{devries2018learning}, we train each segmentation network to output prediction uncertainty along with the segmentation probabilities. The segmentation head is modified to output $C+1$ values per pixel instead of $C$ values, where $C$ is the number of classes in the dataset. The segmentation network is trained with the weighted cross entropy loss and the confidence loss~\cite{devries2018learning}.

%%%------------------------------------------------------------------------
\section{Sequential Ensembles}

\subsection{Implementation Details}
We use the \textit{mmsegmentation}\footnote{\url{https://github.com/open-mmlab/mmsegmentation}} model zoo for our implementation. We report segmentation mean intersection over union (mIoU) using the official evaluation code released by the datasets. For training speedup, we use distributed training using PyTorch~\cite{paszke2019pytorch} using $96$ GeForce RTX 2080 GPUs. Our biggest model with its hyperparameters in SEQ-ENS can be trained using $8$ GPUs at a time. We use FCN~\cite{long2015fully} method for HRNet models, LRASPP~\cite{howard2019searching} for MobileNetv3 models in our experiments. In Table 2 (main paper), the train crop size is set to $512 \times 1024$ for Cityscapes, $512 \times 512$ for ADE20K, $512 \times 512$ for COCO-Stuff, $480 \times 480$ for Pascal-Context. All models are trained for 160k iterations using SGD~\cite{bottou2012stochastic} optimizer. When constructing SIM-ENS, we only add independently trained models to the ensemble if it improves performance. For Table 3 (main paper), the train crop size is set to $1024 \times 1024$ for Cityscapes and $640 \times 640$ for ADE20K. We train HRNet-W48 for 160K iterations with the same learning rate as Segformer~\cite{xie2021segformer} using the ADAM~\cite{kingma2014adam} optimizer.

\subsection{Comparisons with Simple Ensembles}

\noindent
\textbf{All Categories.} We compare SIM-ENS and SEQ-ENS on Cityscapes for all $19$ categories in \cref{table:supplementary:cityscapes}. We observe consistent gains in comparison to SIM-ENS across backbones especially for rare categories.

\begin{table*}[t]
\centering
\small
% \setlength{\tabcolsep}{2.7pt} %% controls col space, bigger => more space between cols
% \rowcolors{1}{}{lightgray} %%% alternating row grey color, comment to remove

\resizebox{7.0in}{!}{

    \renewcommand{\arraystretch}{1.2} % Default value: 1, controls row space
    \begin{tabular}{@{}l|l| c c c c c c c c c c c c c c c c c c c| l@{}}
    \Xhline{3\arrayrulewidth}

    \textbf{Method} & \textbf{Arch} & \textbf{road} & \textbf{swlk} & \textbf{bld.} & \textbf{wall} & \textbf{fnce} & \textbf{pole} & \textbf{tlgt} & \textbf{tsgn} & \textbf{veg} & \textbf{terain} & \textbf{sky} & \textbf{prsn} & \textbf{ridr} & \textbf{car} & \textbf{trck} & \textbf{bus} & \textbf{train} & \textbf{mcyl} & \textbf{bcyl} & \textbf{mIoU} \\
    
     \textcolor{gray}{Class (\%) $\rightarrow$} & & \textcolor{gray}{32.9} & \textcolor{gray}{4.7} & \textcolor{gray}{19.2} & \textcolor{gray}{0.6} & \textcolor{gray}{0.7} & \textcolor{gray}{1.2} & \textcolor{gray}{0.2} & \textcolor{gray}{0.6} & \textcolor{gray}{15.2} & \textcolor{gray}{0.7} & \textcolor{gray}{2.9} & \textcolor{gray}{1.1} & \textcolor{gray}{0.2} & \textcolor{gray}{5.7} & \textcolor{gray}{0.3} & \textcolor{gray}{0.3} & \textcolor{gray}{0.1} & \textcolor{gray}{0.1} & \textcolor{gray}{0.6} & \\
    
    \hline
    \footnotesize{MobileNetv3}~\cite{howard2019searching} & D8s & 95.9	& 77.6	& 87.7	& 39.4	& 47	& 52.5	& 45.1	& 63.1	& 89.8	& 54.3	& 90.2	& 71.1	& 45.1	& 90.6	& 45.7	& 64.7	& 51.5	& 39.3	& 67	& 64.1\\
    SIM-ENS & D8s & 96.1	&77.2	&88.1	&41.1	&49.1	&54.1	&48.1	&65.4	&90.2	&55.7	&91.8	&73.4	&46.4	&90.2	&47.9	&68.2	&51.5	&44.1	&68.3	&65.4 (+1.3) \\
    SEQ-ENS & D8s & \textbf{96.9}	&\textbf{79.1}	&\textbf{89.2}	&\textbf{42.6}	&\textbf{52.4}	&\textbf{56.8}	&\textbf{52.7}	&\textbf{68.3}	&\textbf{90.6}	&\textbf{58.4}	&\textbf{92.3}	&\textbf{74.5}	&\textbf{48.8}	&\textbf{91.4}	&\textbf{54.6}	&\textbf{70.8}	&\textbf{51.8}	&\textbf{47.9}	&\textbf{69.5}	&\textbf{68.7 (+4.6)}\\
    
    \hline
    \footnotesize{MobileNetv3}~\cite{howard2019searching} & D8 & 97.4	&80.9	&90.2	&48.3	&51.4	&57.1	&57.7	&69.6	&91.3	&60.1	&93.3	&75.5	&50.1	&92.7	&58.4	&73.5	&51.6	&50.3	&70.6	&69.5\\
    SIM-ENS & D8 &97.8	&81.6	&90.8	&49.0	&53.5	&58.3	&60.4	&71.6	&91.4	&61.3	&93.8	&76.4	&53.7	&93.1	&62.1	&74.5	&53.7	&54.8	&73.1	&71.1 (+1.6)\\
    SEQ-ENS & D8 &\textbf{98.1}	&\textbf{83.7}	&\textbf{91.1}	&\textbf{50.3}	&\textbf{56.8}	&\textbf{59.8}	&\textbf{65.0}	&\textbf{78.9}	&\textbf{92.0}	&\textbf{62.8}	&\textbf{94.1}	&\textbf{78.1}	&\textbf{57.8}	&\textbf{94.0}	&\textbf{68.9}	&\textbf{75.1}	&\textbf{57.4}	&\textbf{59.4}	&\textbf{75.4}	&\textbf{73.6 (+4.1)} \\

    \hline\hline
    HRNet~\cite{sun2019high} & H-18s &98.2	&85.5	&92.4	&53.2	&61.5	&66.4	&70.3	&77.9	&92.4	&63.8	&94.4	&81.1	&59.2	&94.4	&66.4	&84	 &70.5	&59.7	&76.1	&76.2\\
    
    SIM-ENS & H-18s &98.2	&85.6	&92.7	&56.3	&61.1	&67.1	&71.8	&79.4	&92.8	&\textbf{65.4}	&94.8	&82.4	&61.6	&94.9	&69.4	&84.4	&67.5	&61.9	&77.0	&77.1 (+0.9)\\
    
    SEQ-ENS & H-18s &\textbf{98.4}	&\textbf{86.8}	&\textbf{93.1}	&\textbf{59}	&\textbf{63.9}	&\textbf{69.2}	&\textbf{73.6}	&\textbf{79.6}	&\textbf{92.9}	&64.7	&\textbf{94.9}	&\textbf{82.8}	&\textbf{62.2}	&\textbf{95.1}	&\textbf{75.5}	&\textbf{88}	&\textbf{78}	&\textbf{63.4}	&\textbf{78.2}	&\textbf{78.9 (+2.7)}\\
    
    \hline
    HRNet~\cite{sun2019high} & H-18 &98.3	&86.4	&93.1	&59.1	&63.7	&69.3	&73.2	&80.6	&92.9	&66.4	&95	&82.8	&61.8	&94.9	&75.9	&87.7	&75.5	&60.4	&77.3	&78.7\\
    
    SIM-ENS & H-18 &98.3	&86.7	&93.3	&59.2	&63.6	&69.4	&74.1	&81.1	&93.0	&66.4	&95.3	&83.1	&62.0	&94.5	&76.5	&87.8	&74.3	&60.7	&78.1	&79.0 (+0.3)	\\
    
    SEQ-ENS & H-18 & \textbf{98.4}	&\textbf{87.1}	&\textbf{93.3}	&\textbf{58.0}	&\textbf{64.6}	&\textbf{71.0}	&\textbf{75.0}	&\textbf{82.0}	&\textbf{93.1}	&\textbf{66.4}	&\textbf{95.4}	&\textbf{84.0}	&\textbf{63.2}	&\textbf{95.2}	&\textbf{79.1}	&\textbf{90.1}	&\textbf{78.6}	&\textbf{62.7}	&\textbf{79.1}	&\textbf{79.8(+1.1)}\\
    \hline
    
    HRNet~\cite{sun2019high} & H-48 &98.5	&87.3	&93.4	&56.4	&65.6	&71.2	&75.5	&81.9	&92.9	&\textbf{63.2}	&95.3	&84.1	&66.3	&95.4	&79.7	&89.8	&84.2	&68.8	&80.1	&80.5\\
    
    SIM-ENS & H-48 & 98.6	&87.1	&93.4	&56.6	&65.1	&71.6	&75.8	&82.1	&92.9	&63.1	&95.4	&84.2	&67.1	&95.6	&79.8	&89.9	&84.6	&68.3	&79.9	&80.6 (+0.1)\\
    
    SEQ-ENS & H-48 & \textbf{98.6}	&\textbf{88.0}	&\textbf{93.7}	&\textbf{58.9}	&\textbf{66.9}	&\textbf{72.6}	&\textbf{76.8}	&\textbf{83.2}	&\textbf{93.1}	&62.9	&\textbf{95.7}	&\textbf{84.8}	&\textbf{68.3}	&\textbf{95.7}	&\textbf{80.1}	&\textbf{90.6}	&\textbf{85.6}	&\textbf{68.8}	&\textbf{80.9}	&\textbf{81.3 (+0.8)}\\
    
    \hline\hline
    DeepLabv3+~\cite{chen2018encoder} & R-18 & 98.0	&84.4	&92.3	&51.4	&58.2	&66.1	&69.9	&77.6	&92.2	&63.6	&94.4	&81.3	&60.8	&94.7	&76.1	&85.4	&72.5	&63.1	&76.3	&76.8 \\
    SIM-ENS & R-18 & 98.1	&85.0	&92.6	&51.1	&58.3	&67.3	&71.5	&78.9	&92.4	&64.1	&94.7	&82.1	&61.0	&95.0	&77.8	&87.3	&74.5	&64.3	&76.9	&77.8 (+1.0)\\
    SEQ-ENS & R-18 & \textbf{98.3}	&\textbf{86.3}	&\textbf{93.1}	&\textbf{51.2}	&\textbf{59.4}	&\textbf{69.8}	&\textbf{74.6}	&\textbf{80.8}	&\textbf{92.7}	&\textbf{64.9}	&\textbf{95.0}	&\textbf{83.6}	&\textbf{62.8}	&\textbf{95.6}	&\textbf{79.8}	&\textbf{88.1}	&\textbf{77.0}	&\textbf{67.2}	&\textbf{78.3}	&\textbf{78.9 (+2.1)}\\
    
    \hline
    DeepLabv3+~\cite{chen2018encoder} & R-50 &98.4	&86.6	&93	&51.2	&62.5	&69.5	&74.3	&82.2	&92.7	&63.5	&94.9	&84.2	&65.8	&95.6	&81.1	&88.8	&84.7	&68.6	&79.5	&79.8\\
    SIM-ENS & R-50 &98.5	&86.8	&93.1	&51.8	&62.8	&70.1	&74.5	&82.8	&92.7	&63.7	&95.0	&84.6	&65.9	&95.4	&\textbf{81.4}	&89.8	&85.1	&69.2	&79.8	&80.1 (+0.3)\\
    SEQ-ENS & R-50 & \textbf{98.5}	&\textbf{87.3}	&\textbf{93.3}	&\textbf{53.4}	&\textbf{63.2}	&\textbf{71.2}	&\textbf{75.6}	&\textbf{83.5}	&\textbf{92.9}	&\textbf{65.4}	&\textbf{95.2}	&\textbf{85.0}	&\textbf{66.3}	&\textbf{95.8}	&79.3	&\textbf{91.9}	&\textbf{85.2}	&\textbf{70.0}	&\textbf{80.2}	&\textbf{80.7 (+0.9)} \\
    
    \hline
    DeepLabv3+~\cite{chen2018encoder} & R-101 &98.4	&86.4	&93.4	&54.9	&64.4	&70.2	&74.6	&81.9	&93.0	&62.4	&95.1	&84.6	&67.7	&95.8	&85.2	&91.8	&85.2	&71.3	&80.3	&80.9 \\
    SIM-ENS & R-101 & 98.4	&86.7	&93.4	&55.1	&64.6	&70.1	&74.7	&82.0	&92.8	&62.8	&95.4	&84.7	&67.8	&\textbf{96.3}	&85.0	&91.9	&85.8	&71.9	&80.3	&81.1 (+0.2) \\
    SEQ-ENS & R-101 &\textbf{98.5}	&\textbf{87.0}	&\textbf{93.5}	&\textbf{56.0}	&\textbf{64.6}	&\textbf{71.1}	&\textbf{75.2}	&\textbf{82.6}	&\textbf{93.1}	&\textbf{63.5}	&\textbf{95.4}	&\textbf{84.9}	&\textbf{68.5}	&96.0	&\textbf{85.4}	&\textbf{92.3}	&\textbf{86.5}	&\textbf{72.5}	&\textbf{80.5}	&\textbf{81.5 (+0.6)} \\

    \Xhline{3\arrayrulewidth}
    
    \end{tabular}
}
    
\caption{Comparison of Sequential Ensembles (Ours) with Simple Ensembles for $N=2$ on the Cityscapes \texttt{val} set. R-@ and H-@ stand for ResNet-@ and HRNet-W@ respectively. D8 is the output stride of DeepLabv3+. \texttt{s} denotes the \textit{small} version of the backbone. Without using any extra data, all models are trained at the image resolution of $512 \times 1024$ and tested at $1024 \times 2048$ using single-scale inference. }
\label{table:supplementary:cityscapes}

\end{table*}

\vspace{0.5cm}

\begin{figure}[H]
\centering
\includegraphics[width=0.4\linewidth]{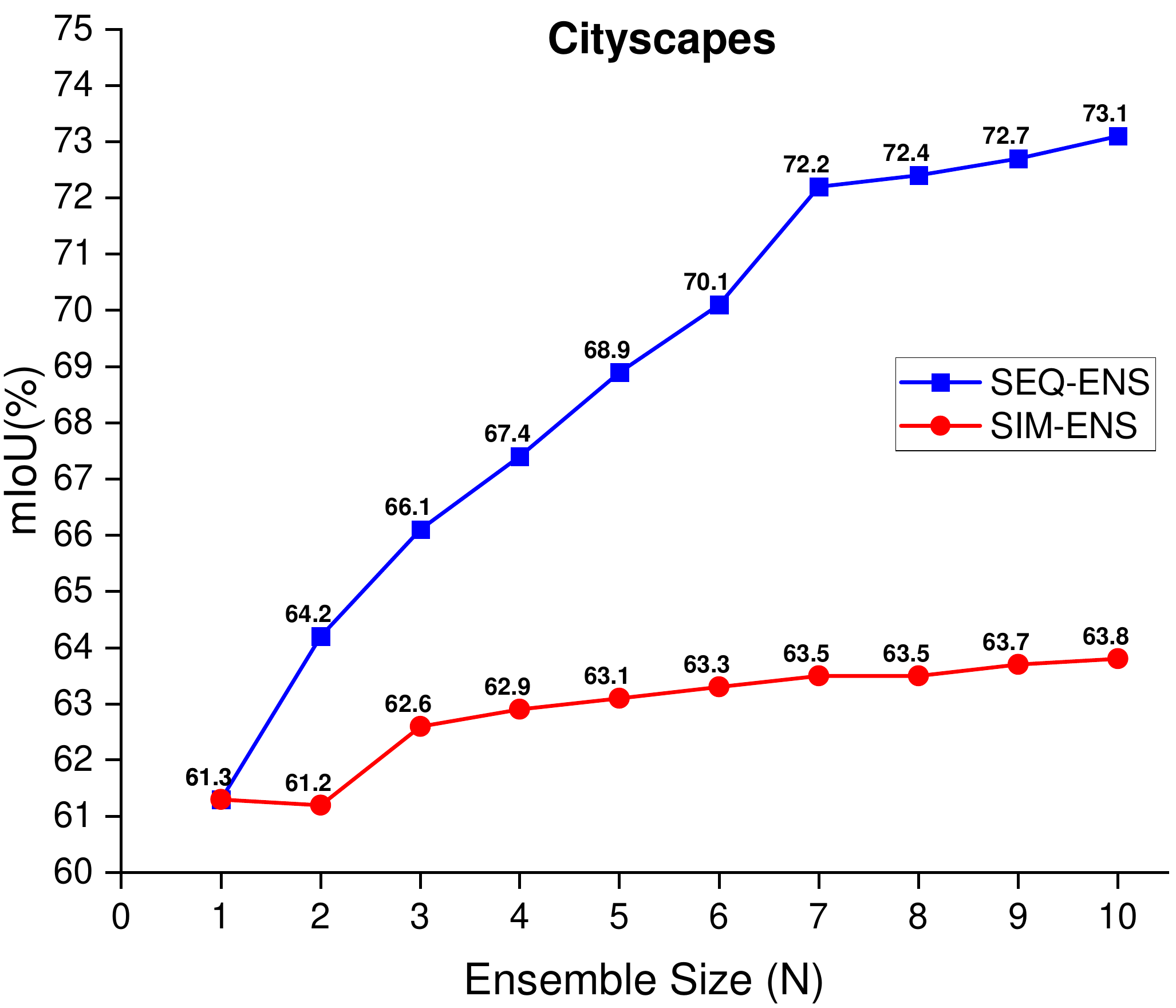}
\includegraphics[width=0.4\linewidth]{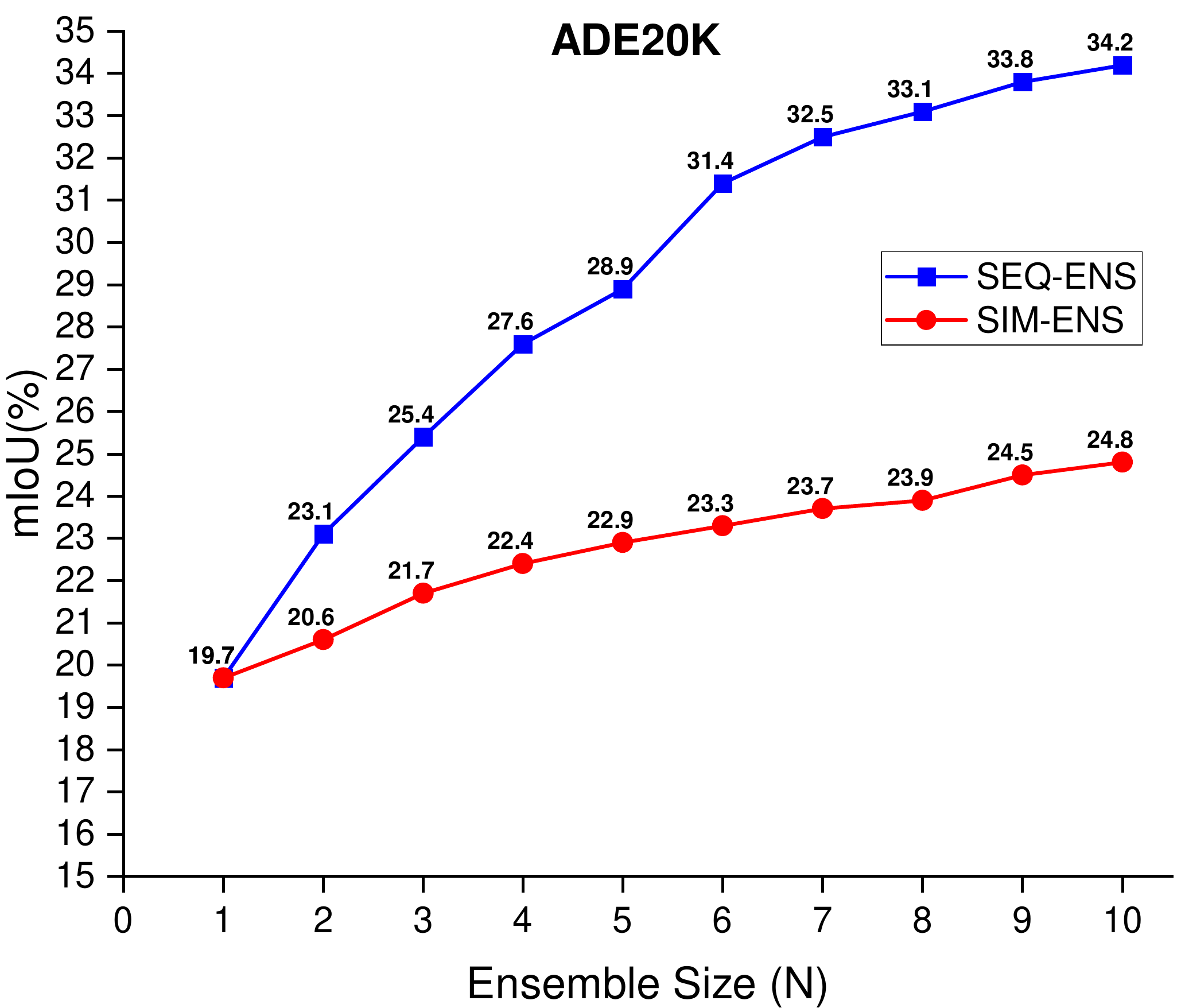}
\includegraphics[width=0.4\linewidth]{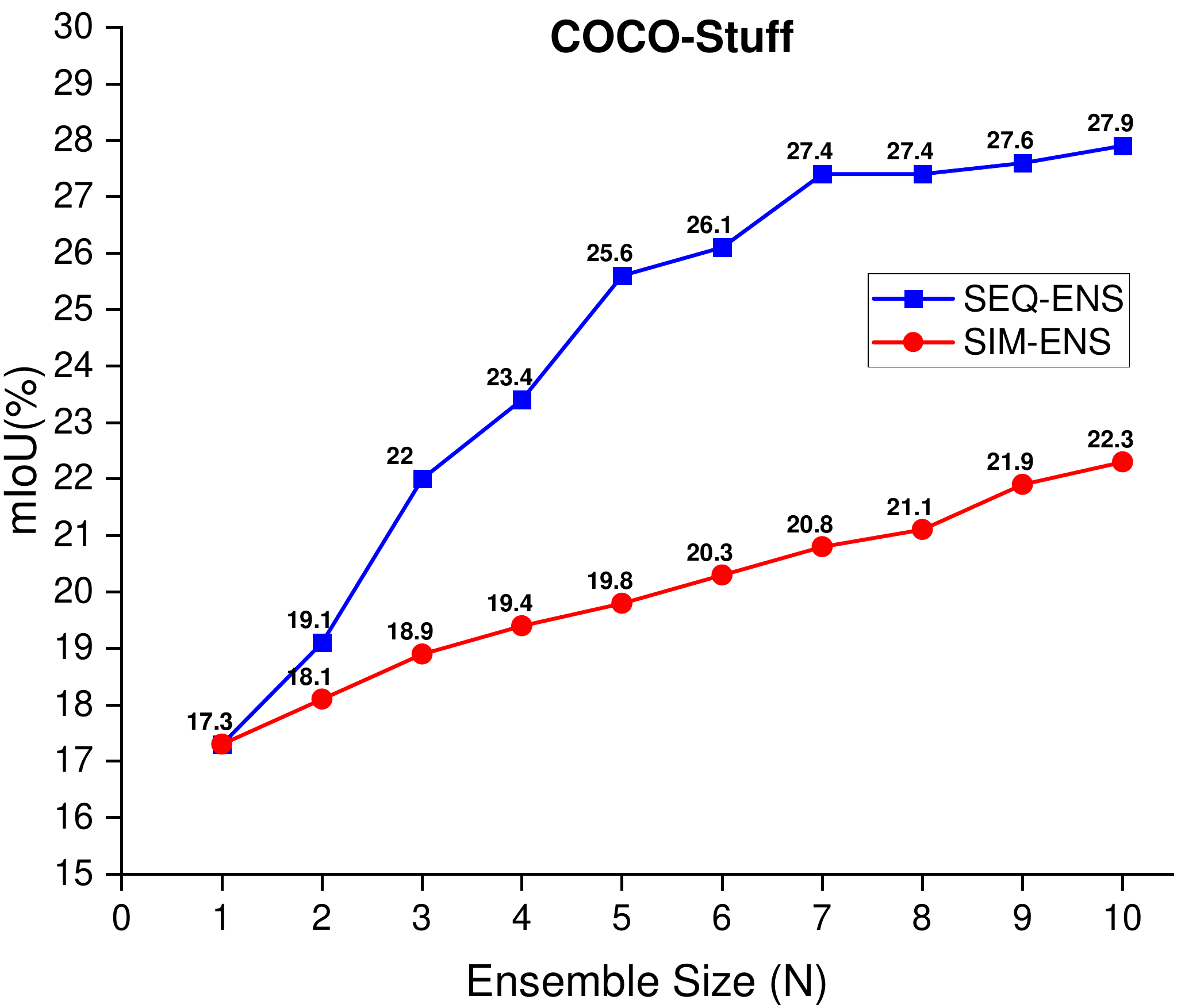}
\caption{Performance of Sequential Ensembling (SEQ-ENS) and Simple Ensembling (SIM-ENS) using FCN~\cite{long2015fully} MobileNetv2-D8~\cite{sandler2018mobilenetv2} backbone on the \texttt{val} set of the various datasets using single-scale testing. We see increase in performance as more generations are added, allowing run-time trade-off between accuracy and speed for edge devices.}
\label{fig:supplementary:all_seq_sim}
\end{figure}

\noindent
\textbf{Increasing N.} We also compare SIM-ENS and SEQ-ENS with increasing ensemble size $N$ on Cityscapes, ADE20K and COCO-Stuff dataset in \cref{fig:supplementary:all_seq_sim}. Across generations, SEQ-ENS consistently outperforms SIM-ENS for MobileNetv2-D8~\cite{sandler2018mobilenetv2} backbone.

\noindent
\textbf{Effect of number of parameters.} Further, to understand if the segmentation improvement obtained by SEQ-ENS is not simply due to increased number of parameters in comparison to SIM-ENS. We compare SEQ-ENS with SIM-ENS* in \cref{table:supplementary:fair_baseline}. SIM-ENS* has the same number of parameters as SEQ-ENS. We construct SIM-ENS* by training models which have the same architecture as $G_i$ in SEQ-ENS with the exception of $G_0$. However the ADON blocks take a fixed random embedding of size $H \times W \times C$ as input instead of the previous generation segmentation probabilities. We report numbers on the Cityscapes \texttt{val} set using HRNet-18-small v1 (from the official HRNet implementation) and v2 backbones~\cite{sun2019high}. SIM-ENS* and SIM-ENS achieve similar performance and SEQ-ENS outperforms SIM-ENS* with equal number of parameters. 

\begin{table*}[b]
\centering
\small
% \setlength{\tabcolsep}{2.7pt} %% controls col space, bigger => more space between cols
% \rowcolors{1}{}{lightgray} %%% alternating row grey color, comment to remove

\resizebox{7.0in}{!}{

    \renewcommand{\arraystretch}{1.2} % Default value: 1, controls row space
    \begin{tabular}{@{}l|l| c c c c c c c c c c c c c c c c c c c| l@{}}
    \Xhline{3\arrayrulewidth}

    \textbf{Method} & \textbf{Arch} & \textbf{road} & \textbf{swlk} & \textbf{bld.} & \textbf{wall} & \textbf{fnce} & \textbf{pole} & \textbf{tlgt} & \textbf{tsgn} & \textbf{veg} & \textbf{terain} & \textbf{sky} & \textbf{prsn} & \textbf{ridr} & \textbf{car} & \textbf{trck} & \textbf{bus} & \textbf{train} & \textbf{mcyl} & \textbf{bcyl} & \textbf{mIoU} \\
    
     \textcolor{gray}{Class (\%) $\rightarrow$} & & \textcolor{gray}{32.9} & \textcolor{gray}{4.7} & \textcolor{gray}{19.2} & \textcolor{gray}{0.6} & \textcolor{gray}{0.7} & \textcolor{gray}{1.2} & \textcolor{gray}{0.2} & \textcolor{gray}{0.6} & \textcolor{gray}{15.2} & \textcolor{gray}{0.7} & \textcolor{gray}{2.9} & \textcolor{gray}{1.1} & \textcolor{gray}{0.2} & \textcolor{gray}{5.7} & \textcolor{gray}{0.3} & \textcolor{gray}{0.3} & \textcolor{gray}{0.1} & \textcolor{gray}{0.1} & \textcolor{gray}{0.6} & \\
    
    \hline
  
    HRNet~\cite{sun2019high} & H-18s-v1 &97.6	&82.2	&90.9	&47.0	&53.0	&60.4	&63.8	&74.3	&91.8	&62.1	&93.7	&78.5	&53.9	&93.1	&50.5	&70.6	 &47.0	&49.8	&73.7	&70.3\\
    
    SIM-ENS & H-18s-v1 &97.5	&82.0	&90.8	&45.7	&52.0	&60.4	&64.2	&74.5	&91.9	&62.5	&93.8	&78.1	&54.0	&92.8	&46.8	&71.8	&54.8	&49.7	&73.1	&70.4 (+0.1)\\
    
    SIM-ENS* & H-18s-v1 &97.7	&82.8	&91.1	&47.7	&53.8	&61.3	&65.5	&75.4	&92.0	&62.9	&93.9	&79.1	&55.0	&93.3	&49.6	&73.7	&53.3	&52.5	&74.3	&71.3 (+1.0)\\
    
    SEQ-ENS & H-18s-v1 &\textbf{98.2}	&\textbf{85.2}	&\textbf{92.2}	&\textbf{55.0}	&\textbf{56.8}	&\textbf{66.1}	&\textbf{69.9}	&\textbf{77.7}	&\textbf{92.5}	&\textbf{65.0}	&\textbf{94.4}	&\textbf{81.2}	&\textbf{59.7}	&\textbf{94.1}	&\textbf{59.8}	&\textbf{80.7}	&\textbf{63.6}	&\textbf{57.3}	&\textbf{75.9}	&\textbf{75.1 (+4.8)}\\
    
    \hline
  
    HRNet~\cite{sun2019high} & H-18s &98.2	&85.5	&92.4	&53.2	&61.5	&66.4	&70.3	&77.9	&92.4	&63.8	&94.4	&81.1	&59.2	&94.4	&66.4	&84	 &70.5	&59.7	&76.1	&76.2\\
    
    SIM-ENS & H-18s &98.2	&85.6	&92.7	&56.3	&61.1	&67.1	&71.8	&79.4	&92.8	&\textbf{65.4}	&94.8	&82.4	&61.6	&94.9	&69.4	&84.4	&67.5	&61.9	&77.0	&77.1 (+0.9)\\
    
    SIM-ENS* & H-18s &98.2	&85.9	&92.7	&55.8	&60.5	&66.9	&71.3	&79.1	&92.8	&64.9	&94.8	&82.1	&60.8	&94.7	&69.3	&83.2	&68.7	&61.4	&77.0	&76.9 (+0.7)\\
    
    SEQ-ENS & H-18s &\textbf{98.4}	&\textbf{86.8}	&\textbf{93.1}	&\textbf{59}	&\textbf{63.9}	&\textbf{69.2}	&\textbf{73.6}	&\textbf{79.6}	&\textbf{92.9}	&64.7	&\textbf{94.9}	&\textbf{82.8}	&\textbf{62.2}	&\textbf{95.1}	&\textbf{75.5}	&\textbf{88}	&\textbf{78}	&\textbf{63.4}	&\textbf{78.2}	&\textbf{78.9 (+2.7)}\\

    \Xhline{3\arrayrulewidth}
    
    \end{tabular}
}
    
\caption{Comparison of SEQ-ENS with SIM-ENS* using HRNet-W18-small (v1 and v2) backbones for $N=2$. SEQ-ENS outperforms SIM-ENS* with equal number of parameters on Cityscapes dataset.}
\label{table:supplementary:fair_baseline}

\end{table*}

\subsection{Analysis}
We statistically analyse the number of segmentation mistakes of $G_0$ fixed by $G_1$ in \cref{table:supplementary:case_analysis}. Each pixel can be divided into 4 cases depending on whether $G_0$ and $G_1$ prediction is correct or incorrect. On the Cityscapes dataset for Mobile-Netv3-D8-small~\cite{howard2019searching} backbone, we observe that $G_1$ in SEQ-ENS fixes $G_0$'s mistake relatively 9 times in comparison to the case when $G_0$ is correct and $G_1$ is incorrect. We observe similar trend for Mobile-Netv3-D8 backbone.

\begin{table}[H]
    \centering
    \small
    \renewcommand{\arraystretch}{1.2}
    \rowcolors{3}{}{lightgray}
    \setlength{\tabcolsep}{3pt}
    \begin{tabular}{@{}c|c c|r  r|r r@{}}
        \Xhline{3\arrayrulewidth}
       \multirow{2}{*}{\textbf{Case No.}} & \multicolumn{2}{c|}{\textbf{Is Prediction Correct?}} & \multicolumn{2}{c|}{\textbf{Mobile-Netv3-D8s}} & \multicolumn{2}{c}{\textbf{Mobile-Netv3-D8}}\\
         \cmidrule(r){2-3}   \cmidrule(r){4-5}\cmidrule(r){6-7}
         & $\mathbf{G_0}$ & $\mathbf{G_1}$  & Abs. Frequency   & Rel. Frequency  & Abs. Frequency   & Rel. Frequency    \\
        \hline
    1 & \cmark & \cmark     &   90.5\%    &   150.8$x$  & 93.2\%  & 233$x$  \\
    2  & \cmark & \xmark   &   0.6\%    &   $x$   & 0.4\% & $x$ \\
    3    & \xmark & \cmark   &    5.4\%   &   9$x$  & 3.2\% & 8$x$ \\
    4    & \xmark & \xmark   &   3.5\%    &   5.8$x$  & 3.2\% & 8$x$\\

        \Xhline{3\arrayrulewidth}
    \end{tabular}
    \caption{Per pixel prediction analysis of $G_0$ and $G_1$ in SEQ-ENS. We divide all the pixels into four cases as shown in the table and compute the frequency of the case. We observe that $G_1$ predominantly fix mistakes made by $G_0$ (\textit{Case 3}) in comparison to \textit{Case 2}.}
    \label{table:supplementary:case_analysis}
\end{table}

\subsection{Comparison to Early and Late Fusion}
We compare different variants of SEQ-ENS in \cref{table:supplementary:early_late} with using ADON blocks on the Cityscapes dataset. \textit{Early-Fusion} concatenates the RGB image and the prediction probabilities channelwise. In \textit{Late-Fusion}, we downsample the prediction probabilities and concatenate with the intermediate feature after \textit{Stage-3} in the HRNet architecture. Injection of high resolution information at multiple-depths using ADON blocks outperforms both early and late fusion in SEQ-ENS.

\begin{table}[H]
\centering
\small
    \renewcommand{\arraystretch}{1.2}
    % \rowcolors{1}{}{lightgray}
    \setlength{\tabcolsep}{3pt}
    \begin{tabularx}{\textwidth}{ c *{6}{Y} }
        \Xhline{3\arrayrulewidth}
        \textbf{Arch} & $\mathbf{G_0}$  & \textbf{SIM-ENS}  & \textbf{SEQ-ENS} (\textit{Early-Fusion})  & \textbf{SEQ-ENS} (\textit{Late-Fusion}) & \textbf{SEQ-ENS} (\textit{ADON}) \\
        \hline
        \textbf{HRNet-W18-small} & 76.2  &   77.1    & 77.5  & 77.8    & \textbf{78.9}   \\
        \textbf{HRNet-W18}  & 78.7  &   79.0    & 79.2  & 79.1    & \textbf{79.8} \\
        \textbf{HRNet-48}  & 80.5  &   80.6   & 80.7  & 80.2    & \textbf{81.3}  \\
        \Xhline{3\arrayrulewidth}
    \end{tabularx}
\caption{Comparison of different variants of SEQ-ENS (N=2) using ADON blocks, early-fusion and late-fusion to utilize information from previous generations on the Cityscapes dataset.}
\label{table:supplementary:early_late}
\end{table}

%%%%%------------------------------------extra figures------------------------------------
\newpage

\begin{figure*}
\centering
\includegraphics[width=1\linewidth, height=0.4\linewidth]{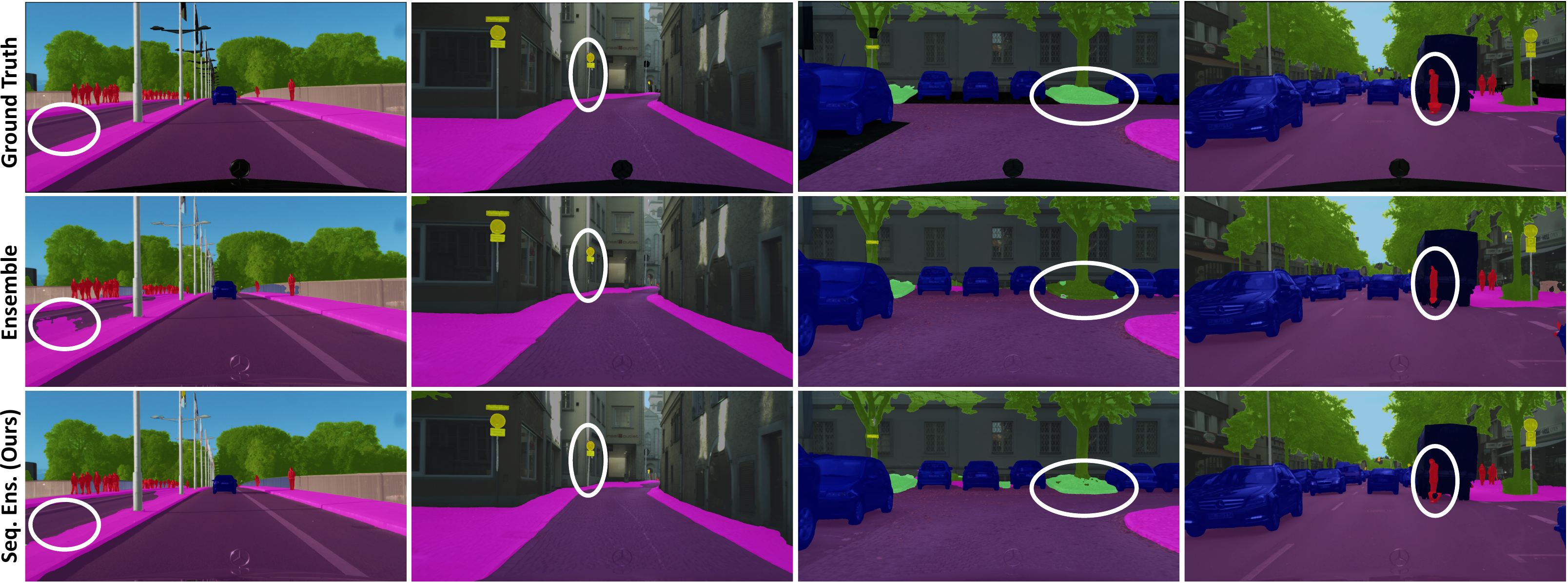} \\
\vspace{0.3cm}
\includegraphics[width=1\linewidth, height=0.4\linewidth]{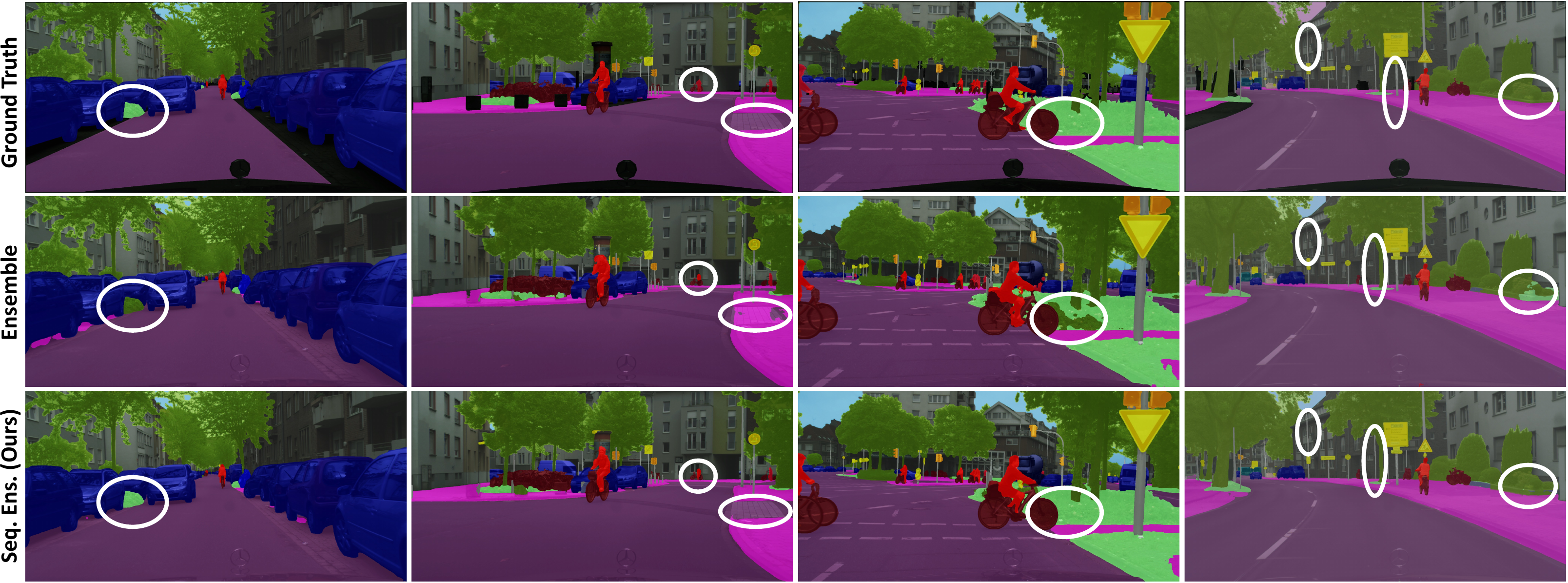}\\
\vspace{0.3cm}
\includegraphics[width=1\linewidth, height=0.4\linewidth]{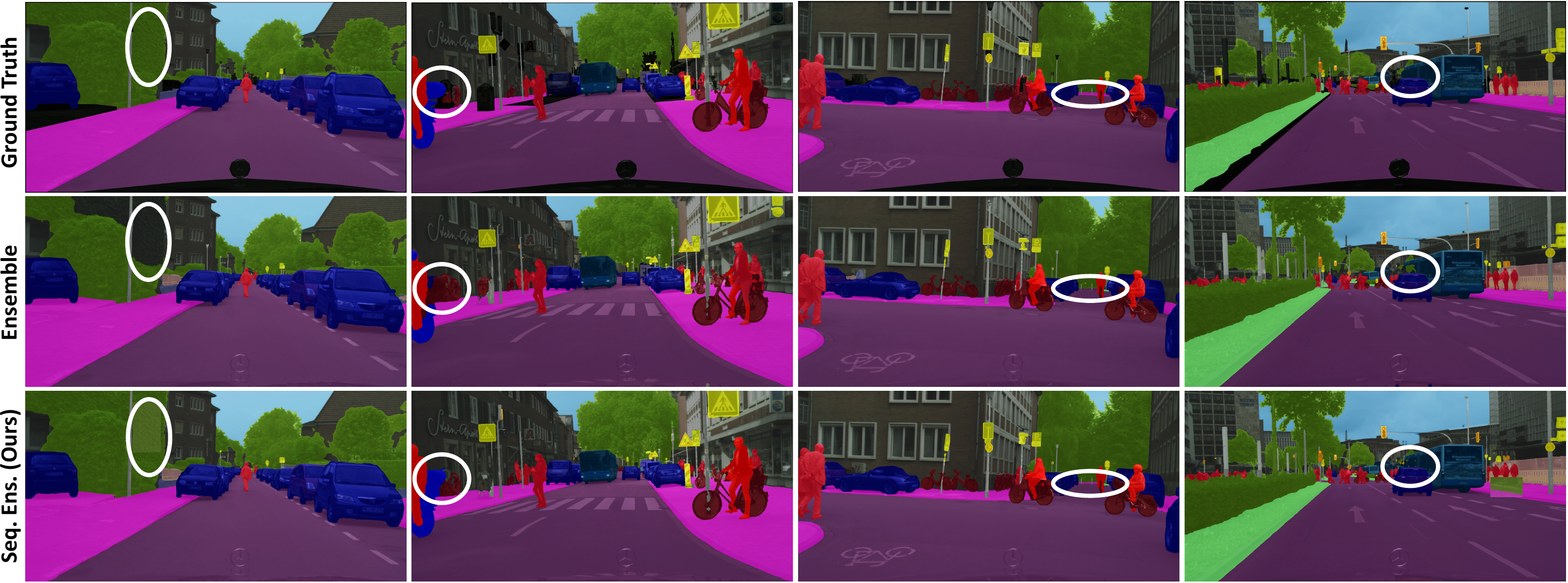} 
\caption{Qualitative comparison of Sequential Ensembles (SEQ-ENS) with the Simple Ensembles (SIM-ENS) on Cityscapes \texttt{val} set using HRNet-W48 backbone (N=2). The white eclipses highlight the fine-grained details that our approach captures in comparison to the baseline. Zoom in for details.}
\label{fig:supplementary:qual_cityscapes}
\end{figure*}

\newpage

\begin{figure*}
\centering
\includegraphics[width=1\linewidth, height=0.4\linewidth]{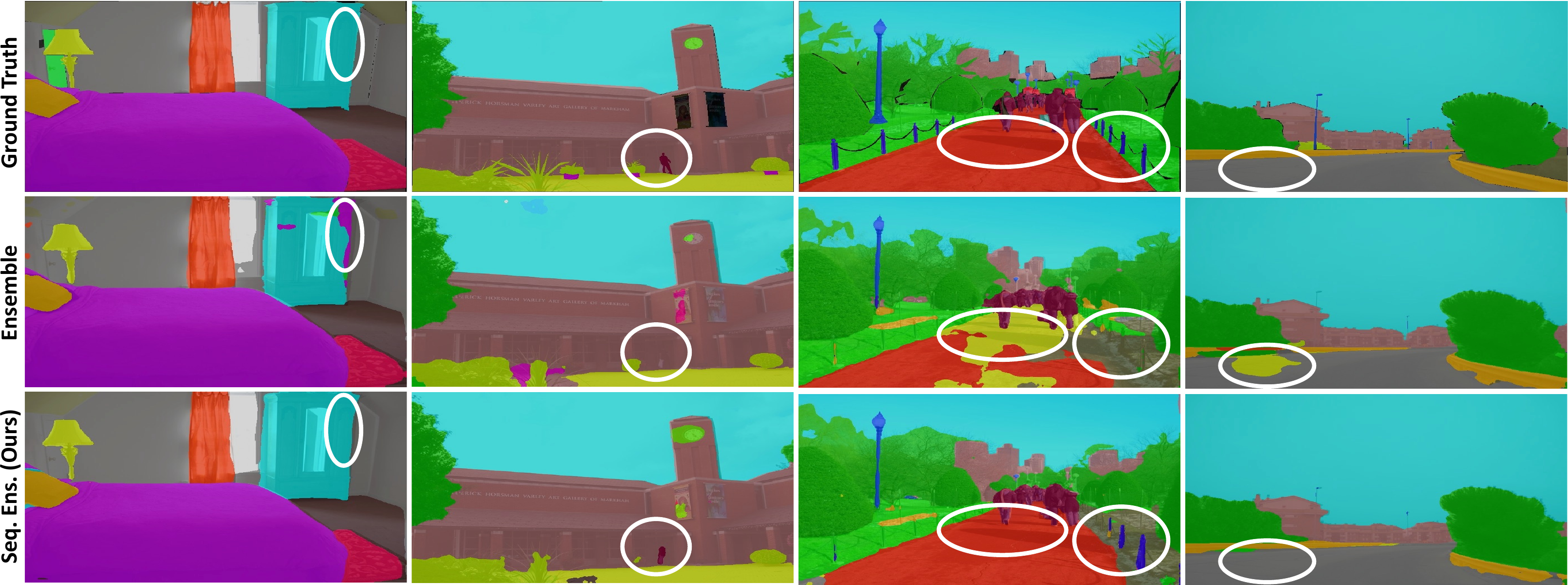} \\
\vspace{0.3cm}
\includegraphics[width=1\linewidth, height=0.4\linewidth]{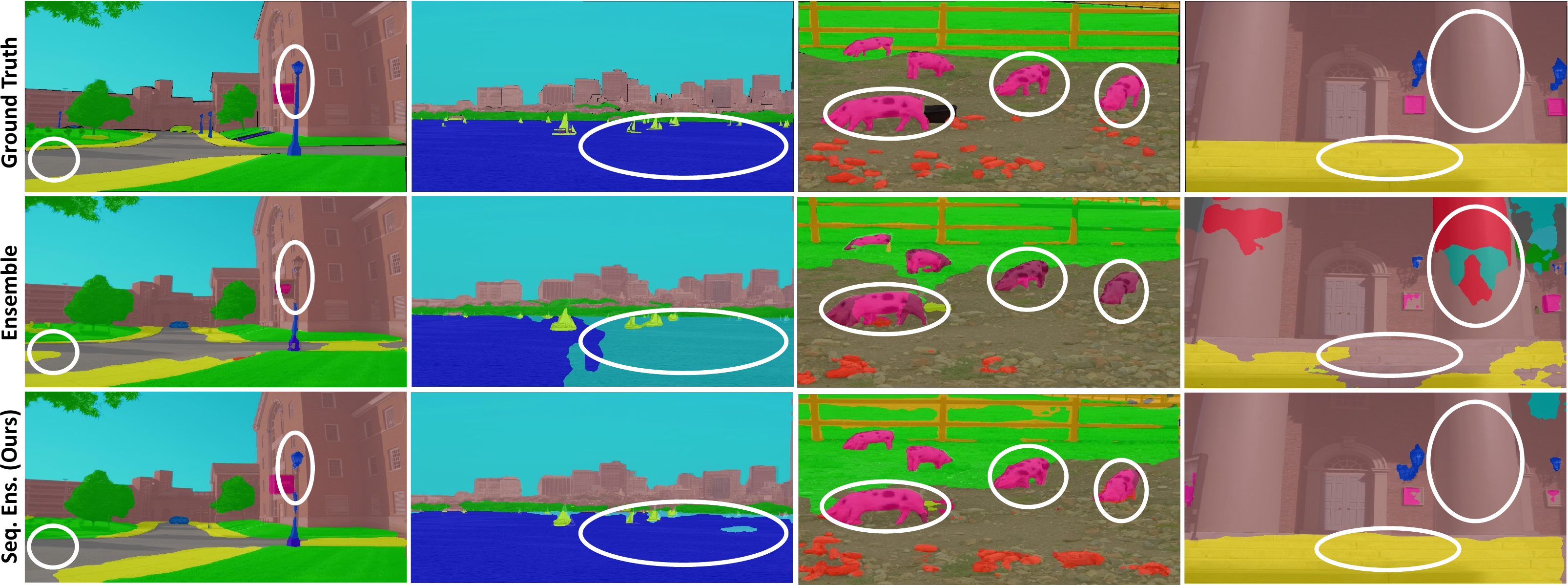}\\
\vspace{0.3cm}
\includegraphics[width=1\linewidth, height=0.4\linewidth]{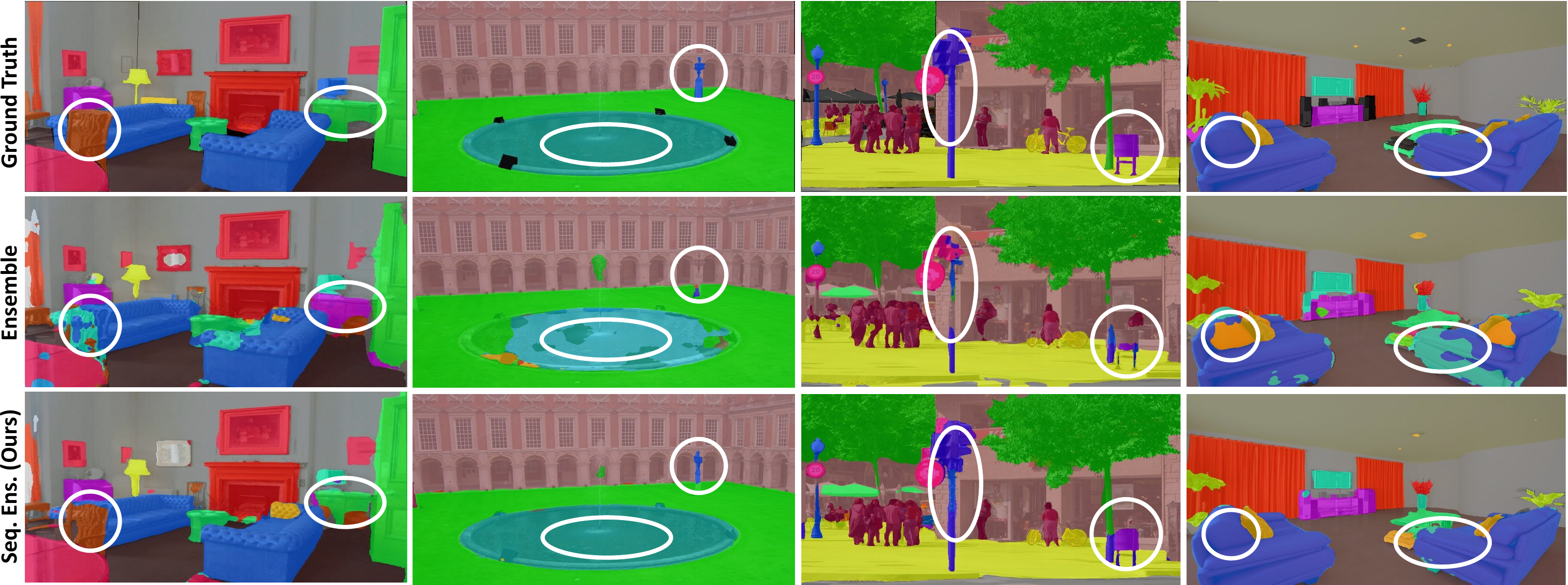}
\caption{Qualitative comparison of Sequential Ensembles (SEQ-ENS) with the Simple Ensembles (SIM-ENS) on ADE20K \texttt{val} set using HRNet-W48 backbone (N=2). SEQ-ENS is especially effective on a complex dataset like ADE-20K with $150$ categories. We set a new state-of-art on ADE20K \texttt{val} set. Zoom in for details.}
\label{fig:supplementary:qual_ade20k}
\end{figure*}

\clearpage
\newpage

{\small
\bibliographystyle{ieee_fullname}
\bibliography{references}
}